%% file: NEURCOMP-D-23-00020R1.tex
\documentclass[11pt]{article}
\usepackage{jmlr}

\input{NEURCOMP-D-23-00020R1-macros}

\begin{document}


\title{Towards Improving the Generation Quality of Autoregressive Slot VAEs}

\author{\name  Patrick Emami \email Patrick.Emami@nrel.gov \\
    \addr National Renewable Energy Lab
    \AND
    \name Pan He \email Pan.He@auburn.edu \\
    \addr Auburn University 
    \AND 
    \name Sanjay Ranka \email ranka@cise.ufl.edu \\
    \addr University of Florida
    \AND
    \name Anand Rangarajan \email anand@cise.ufl.edu \\
    \addr University of Florida
}

%

\editor{}

\maketitle

\begin{abstract}
    
 Unconditional scene inference and generation are challenging to learn jointly with a single compositional model. Despite encouraging progress on models that extract object-centric representations (``slots'') from images, unconditional generation of scenes from slots has received less attention. This is primarily because learning the multi-object relations necessary to imagine coherent scenes is difficult. We hypothesize that most existing slot-based models have a limited ability to learn object correlations.  We propose two improvements that strengthen object correlation learning. The first is to condition the slots on a global, scene-level variable that captures higher-order correlations between slots.  Second, we address the fundamental lack of a canonical order for objects in images by proposing to learn a consistent order to use for the autoregressive generation of scene objects. Specifically, we train an autoregressive slot prior to sequentially generate scene objects following a learned order. Ordered slot inference entails first estimating a randomly ordered set of slots using existing approaches for extracting slots from images, then aligning those slots to ordered slots generated autoregressively with the slot prior. Our experiments across three multi-object environments demonstrate clear gains in unconditional scene generation quality.  Detailed ablation studies are also provided that validate the two proposed improvements.

\end{abstract}


\section{Introduction}
Humans have the innate ability to understand scenes as a composition of multiple related objects and imagine new scenes containing those same objects~\citep{kahneman1992reviewing,baillargeon2004infants,spelke2007core}.
Developing agents capable of human-level compositional scene understanding would enable, for example, robots to reason about the effects of their actions on different objects~\citep{battaglia2013simulation,lake2017building,zhu2020dark}.
Variational autoencoders (VAEs)~\citep{kingma2013auto,Rezende2014} are a natural model family for compositional scene understanding since they are based on a principled unsupervised framework that unifies scene inference and generation~\citep{yuille2006vision}. 
Compositional scene inference entails \emph{segregating} raw perceptual inputs into object-centric entities, known as the \emph{binding problem}~\citep{greff2020binding}, whereas compositional scene generation entails imagining scenes by composing these representations.
However, learning these jointly in a single model has proven to be challenging.
Vanilla VAEs do not address the binding problem, making them easy to understand and train although their unstructured scene representations hinder their ability to systematically generalize to new scenes~\citep{pmlr-v97-greff19a,greff2020binding}.
An advanced class of VAEs instead represent scenes as mixture distributions where the mixture components are in correspondence with a set of latent slot variables.
These \emph{slot VAEs} achieve encouraging segmentation and disentanglement performance on static~\citep{burgess2019monet,pmlr-v97-greff19a,engelcke2019genesis,engelcke2021genesis,pmlr-v139-emami21a}, multi-view~\citep{nanbo2020learning,stelzner2021decomposing,yuan2021unsupervised,yu2021unsupervised,li2021object}, and dynamic visual scenes~\citep{watters2019cobra,veerapaneni2019entity,linimproving,zablotskaia2021provide,creswell2021unsupervised,kabra2021simone,zoran2021parts}.

However, slot VAEs have been limited in their ability to perform unconditional scene generation.
To simplify training and evaluation, oftentimes the generative aspects of the model are entirely removed (e.g., Slot Attention~\citep{locatello2020object}).
Other slot VAEs assume slots are independent and thereby uncorrelated, making them demonstrably incapable of generating coherent scenes~\citep{pmlr-v97-greff19a,pmlr-v139-emami21a}.
Certain slot VAEs that autoregressively generate slots~\citep{engelcke2019genesis,engelcke2021genesis} should in theory be capable of high-quality compositional generation, yet empirically have struggled in this regard.
We hypothesize that the poor sample quality of slot VAEs is explained in part by a limited ability to learn object correlations.
Learning how scene objects are correlated when training a slot VAE is essential for the model to learn to compositionally generate novel instances of such scenes.
This is necessary to avoid generating objects in physically implausible situations.
For example, if the model never sees a training scene where two cars occupy the same physical space, then it should not generate such scenes (unless explicitly guided to).

This paper introduces two strategies for improving the scene generation quality of slot VAEs.
Both strategies enable stronger learning of correlations between slots.
First, we propose conditioning autoregressive slots on a scene-level latent variable.
This creates a multi-level slot VAE where the top-level variable has the effect of increasing the correlation between the bottom-level slot variables.
Second, we reconcile the fact that \emph{objects have no canonical ordering} with \emph{sequential modeling of slots}.
In general, autoregressive models benefit from a canonical sequence order because sequences are seen during training in a \emph{consistent} order.
This provides a strong signal for learning how sequence elements are correlated. 
For example, consider the success of pixel-level autoregressive models based on raster scan order (a canonical ordering for pixels)~\citep{van2016pixel,salimans2017pixelcnn++,van2016conditional,parmar2018image,child2019generating,chen2020generative}.
These models always generate pixels in the top-left corner of an image first.
In images of outdoor scenes, for example, the top-left corner is often blue sky, which the model can exploit to ease the difficulty of learning a conditional distribution for these pixels.
This suggests that the lack of a canonical object order has likely hindered correlation learning in \revision{autoregressive} slot VAEs.

To address the lack of a canonical object order, we develop an approach for learning a consistent object inference and generation order.
Learning an order avoids the problem of inventing a fixed, unnatural object order to use for autoregressively modeling slots.
We introduce an inference algorithm that \emph{aligns} a randomly ordered set of slots, which we easily obtain from existing approaches, to ordered slots generated from an autoregressive prior rollout.
Crucially, the prior is trained to generate objects in a consistent order useful for high quality scene generation by an auxiliary loss that maximizes the visual quality of the scenes generated from the prior rollouts.

In our experiments, we observe promising improvements in generative quality due to the scene-level variable and inference with object order alignment across three multi-object environments.
We carefully ablate each proposed mechanism to assess their individual contributions to performance.
Comparisons are conducted with respect to multiple previous slot VAEs from the literature.

\section{Related Work}
\label{sec:relwork}

The objective of this work is to improve the generative quality of autoregressive slot VAEs.   
Slot VAEs are VAEs that have a set of $K$ interpretable latent variables associated with scene objects.
We categorize these VAEs by their choice of modeling distribution for the slots.
IODINE~\citep{pmlr-v97-greff19a} and EfficientMORL (EMORL)~\citep{pmlr-v139-emami21a} infer an independent (randomly ordered) slot distribution.
Neither model is capable of generating coherent scenes since both assume objects in scenes are uncorrelated.
Other simpler slot VAEs make similar independence assumptions that prevent them from scaling to non-trivial environments~\citep{anciukevicius2020object,von2020towards}.
GENESIS-v2 (GENv2)~\citep{engelcke2021genesis} also infers an independent slot distribution (via a  differentiable clustering algorithm capable of handling a variable number of slots) but uses an \emph{autoregressive} slot prior for scene generation.
We show in this work that inferring an independent slot distribution hurts the sample quality of generated scenes.
GENESIS (GEN) \citep{engelcke2019genesis}), which shares the same autoregressive slot prior, uses sequential attention to infer an autoregressive slot posterior.
However, this inference mechanism has been observed to work poorly on visual datasets with non-trivial textures, colors, and lighting~\citep{engelcke2021genesis}.
As there is no canonical object order to follow for autoregressive inference, GEN relies on learning a (potentially arbitrary) inference order.
By contrast, we introduce an auxiliary loss term to learn a \emph{consistent} object generation order for high quality scene generation, which we then impose during inference on a randomly ordered set of slots.
Generative Neurosymbolic Machine (GNM) \citep{jiang2020generative} is a hierarchical VAE with a set of highly structured symbolic latent variables augmented by a scene-level prior.
Inspired by GNM, we propose augmenting an autoregressive slot prior with a global prior for better modeling of higher-order correlations.
GNM's symbolic variables differ from the slots of the aforementioned VAEs in that they have explicit semantic meaning (e.g., as bounding box coordinates) and are modeled with a strong image prior~\citep{eslami2016attend,crawford2019spatially,lin2020space}.
This prior arranges the latent variables as cells in a raster-scan-ordered grid which aids scene generation yet may cause difficulty with discovering objects in environments where objects vary in size and occlude each other.
GSGN~\citep{deng2021generative} attempts to learn part-whole object hierarchies but uses the same type of symbolic latent variables as GNM.
\revision{SIMONe~\citep{kabra2021simone} uses spatiotemporal cues from dynamic scenes to infer both time-varying frame-level latents and time-invariant object latents.
Unlike the models we consider in this work, SIMONe is not a proper generative model and cannot sample novel videos unconditionally.
Additionally, it does not model correlations \emph{between} latent variables (i.e., it assumes all latents are independent).}

In this work, we draw a connection between slot inference and sequence modeling tasks where the input data has no canonical order (e.g., it is an arbitrarily ordered set) but the output is ordered~\citep{vinyals2016order}.
\citet{vinyals2016order} showed that ``order matters'', i.e., oftentimes an order can be learned end-to-end to achieve better performance than arbitrary or random orderings.
Popular set-to-sequence approaches include Pointer Networks~\citep{vinyals2015pointer}, continuous relaxations of permutations~\citep{adams2011ranking,mena2018learning,zhang2018learning,li2021discovering}, and differentiable sorting~\citep{,grover2019stochastic}.
See~\citet{jurewicz2021set} for a survey.
\revision{To learn a slot order end-to-end (e.g., via relaxations),  the model has to solve a difficult discrete inference problem. 
Therefore, we leave this direction for future work.}
Instead of directly learning the slot order, we \revision{explore learning} to generate objects in a consistent order with an autoregressive prior.
During inference, this learned order is extracted and imposed on a set of slots.
Differently, ``order-agnostic'' autoregressive models average predictions across many factorization orders; however, they assume the input sequences have (and remain in) a canonical order for consistency~\citep{germain2015made,uria2016neural,NEURIPS2019_dc6a7e65}.

\section{Generative Model}
\label{sec:model}
In this section, we describe our autoregressive slot VAE that, differently from previous such models, is conditioned on a scene-level latent variable.
The fundamental problem we are interested in is fitting the underlying data-generating process $p(\mathcal{D})$ for an unlabeled dataset $\mathcal{D}$ of i.i.d. scenes $\rvx \in \mathbb{R}^{N\times C}$ containing multiple objects.
We restrict our focus to collections of RGB images so that $N=H \times W$ and $C=3$.
We make a standard slot VAE assumption that the marginal likelihood of an image $p(\rvx)$ is augmented with $K$ slots $\rvz_k \in \mathbb{R}^{z}$.
Typically, $K$ is chosen to be larger than the number of objects in any given scene in $\mathcal{D}$.
\begin{wrapfigure}[12]{r}{0.4\textwidth}
\centering
\includegraphics[scale=0.95]{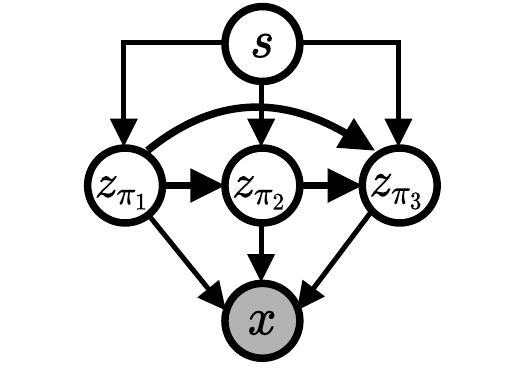}
\caption{The scene-slot generative model ($K=3$).\label{fig:a}}
\vspace{-8em}
\end{wrapfigure}
We introduce a latent variable $\rvs \in \mathbb{R}^{s}$ to serve as a hierarchical prior on the slots and that summarizes global statistics such as higher-order correlations between objects.
This gives the following joint distribution: $p(\rvx, \rvz_{\pi_{1:K}}, \rvs)$, where $\pi_{1:K} := \pi_1,\pi_2,\dots,\pi_K$ is some permutation of the integers $\{1,\dots,K\}$.
We use this notation to make slot order explicit. 
In what follows, $\theta$ are neural net parameters for prior distributions and $\psi,\phi$ are neural net parameters for variational posterior distributions.
Following common practice all latents are Gaussian with diagonal covariance.

To factorize the joint distribution over variables, we use a two-level hierarchy (Figure~\ref{fig:a}).
The image $\rvx$, when conditioned on the slots, is independent of the scene variable:
\begin{equation}
\label{eq:joint}
p_\theta(\rvx, \rvz_{\pi_{1:K}}, \rvs) = p(\rvs) p_\theta(\rvz_{\pi_{1:K}} \mid \rvs) p_\theta(\rvx \mid \rvz_{\pi_{1:K}}).
\end{equation}
We assume a standard Gaussian for $p(\rvs)$ throughout.
The slot prior is autoregressive with order $\pi_{1:K}$:
\begin{equation}
p_\theta(\rvz_{\pi_{1:K}} \mid \rvs) = p_\theta(\rvz_{\pi_1} \mid \rvs) \prod_{k=2}^K p_\theta(\rvz_{\pi_k} \mid \rvz_{\pi_{1:k-1}}, \vs).
\end{equation}
Unlike the autoregressive slot prior used by GEN and GENv2, the slot prior is conditioned on $\rvs$ which increases its expressiveness for modeling higher-order correlations.
The conditional image likelihood can be a Gaussian:
\begin{equation}
\label{eq:mixture-model}
p_\theta(\rvx \mid \rvz_{\pi_{1:K}}) := \prod_{i=1}^N \mathcal{N} \left( \sum_{k=1}^K m_{i,k} x_{i,k}, \sigma^2 \right),    
\end{equation}
or Mixture-of-Gaussians:
\begin{equation}
    \label{eq:MoG}
    p_\theta(\rvx \mid \rvz_{\pi_{1:K}}) := \prod_{i=1}^N \sum_{k=1}^K m_{i,k} \mathcal{N}  (  x_{i,k}, \sigma^2),
\end{equation}
where $x_{i,k} \in \mathbb{R}^{C}$ is an RGB pixel, $m_{i,k} \in [0,1]$ is a mask, and $\sigma^2$ is a variance shared across pixels and slots.
While a Gaussian likelihood is easier to optimize, the Mixture-of-Gaussians is more expressive and hence can achieve better segmentation and reconstruction quality~\citep{burgess2019monet,pmlr-v97-greff19a,engelcke2019genesis,engelcke2021genesis}.
We consider both in our experiments.
To facilitate comparisons with previous work, we use the same spatial broadcasting decoder (SBD)~\citep{watters2019spatial,pmlr-v139-emami21a,engelcke2021genesis} to map slots to mixture components, and leave the exploration of alternative autoregressive decoders for future work~\citep{singh2022illiterate}.
The autoregressive prior $p_\theta(\rvz_{\pi_{1:K}} \mid \rvs)$ is implemented with an LSTM~\citep{hochreiter1997long}.
To compute the input for the first slot prior $p_\theta(\rvz_{\pi_1} \mid \rvs)$, we project the scene variable to dimension $z$ via $\mW_s^{\intercal} \rvs$, $\mW_s \in \mathbb{R}^{s \times z}$ followed by a nonlinear activation function.
Two shared linear layers are used to map the $K$ LSTM outputs to Gaussian means and variances. 

To generate an image, we first sample from the scene prior and then sample sequentially from the autoregressive slot prior.
The $K$ sampled slots are passed to the SBD to predict masks and RGB images.
The $K$ masks are normalized with a softmax before the sum aggregation in Eq.~\ref{eq:mixture-model} or Eq.~\ref{eq:MoG}.

\begin{figure}[ht!]
    \centering
    \includegraphics[width=0.75\textwidth]{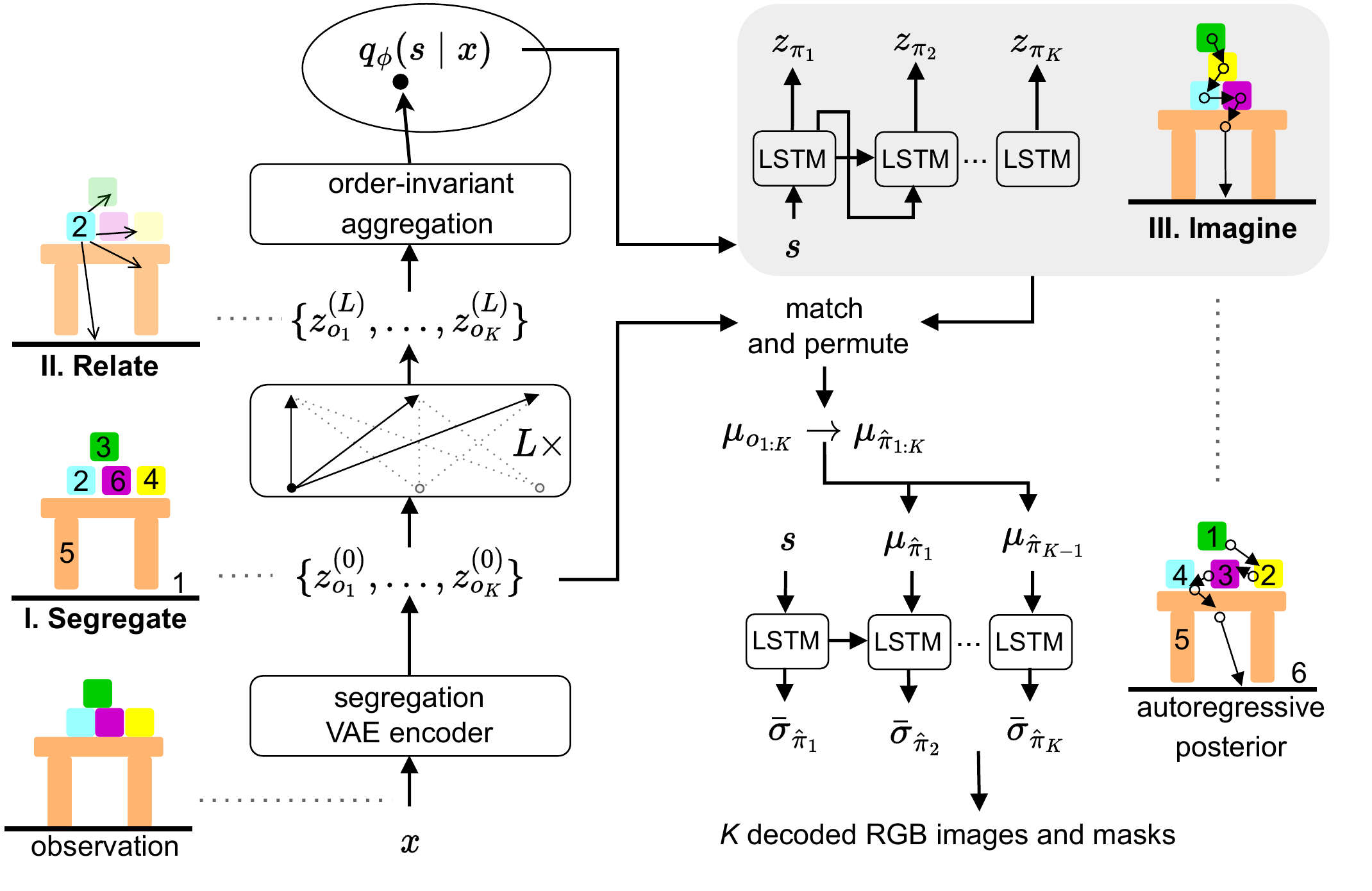}
    \caption{\textbf{Segregate, Relate, Imagine (SRI) Inference}. 
    To improve the ability of slot VAEs to learn correlations between objects, we propose learning a consistent object order for inference and generation.
    Inference begins by \emph{segregating} the observation $\rvx$ to obtain randomly ordered slots $\rvz_{o_{1:K}}$.
    Next, an order-invariant \emph{relational} embedding of this set is encoded into a global posterior $q_\phi(\rvs \mid \rvx)$.
    Finally, an \emph{imagination rollout} conditioned on a sample $\rvs$ gives us slots with consistent order $\pi_{1:K}$.
    We extract order $\hat{\pi}_{1:K}$ by matching the randomly ordered and imagined slots, then transform the segregation posterior into an autoregressive posterior with order $\hat{\pi}_{1:K}$ by predicting correlated variances $\bar{\sigma}_{\hat{\pi}_{1:K}}$.
    The autoregressive slots $\rvz_{\hat{\pi}_{1:K}}$ are decoded into RGB images and masks.}
    \label{fig:teaser}
\end{figure}

\section{Segregate, Relate, Imagine (SRI) Inference}
\label{sec:sri-inference}
\revision{In this section, we describe how to infer the latent variables in the hierarchical slot VAE introduced in Section~\ref{sec:model}.
Without a canonical object order, there is no \emph{a priori} order to use for slot inference, which makes inference difficult.
}
We now introduce our inference algorithm  \revision{which addresses this difficulty}. 
Inspired by the ``order matters'' principle~\citep{vinyals2016order}, we cast autoregressive slot inference as a set-to-sequence modeling problem\revision{, where} the slot inference order is \revision{aligned to match the order in which objects are generated}.

Our inference algorithm has three stages (Figure~\ref{fig:teaser}, Algorithm~\ref{alg:inference} in the appendix).
In the first stage (Segregate, Section~\ref{sec:sr}), we segregate the observation and obtain a randomly ordered slot posterior.
In the second stage (Relate, Section~\ref{sec:sr}), we compute order-invariant correlations between slots, which get summarized in a scene-level latent space.
The third stage (Imagine, Section~\ref{sec:imagine}) transforms the segregation slot posterior into an autoregressive posterior using a learned object order.

\revision{At this point, it is reasonable to ask whether a simpler alternative to our three-stage inference algorithm could suffice, such as one based on Slot Attention~\citep{locatello2020object} with a \emph{fixed left-to-right slot order}.
Slot Attention is a general mechanism for grouping image pixels into slots by applying recurrent key-value-attention-based updates to a set of slots.
However, in practice, imposing a fixed order on how objects are assigned to slots is surprisingly difficult.
We can try to achieve this by learning $K$ distinct slot initialization parameters end-to-end (somewhat similar to SloTTAr~\citep{gopalakrishnan2023unsupervised}).
However, it is unlikely that modifying Slot Attention to learn separate slot initialization parameters is sufficient for learning a fixed object inference and generation order.
This would likely also require imposing an artificial ordering on the segregation step itself (e.g., always assign the bottom-most block of the block tower to slot 1), which seems non-trivial to do and is exactly what our three-stage inference algorithm avoids.
As a sanity check, we compare our inference approach to a baseline that uses a modified Slot Attention encoder in Section~\ref{sec:fixed-experiments}.}

\subsection{Segregate, Relate}
\label{sec:sr}

\textbf{First Stage (Segregate): }The segregation slot posterior is $q_\psi(\rvz_{o_{1:K}} \mid \rvx) = \prod_{k=1}^K q_\psi(\rvz_{o_k} \mid \rvx)$ with random order $o_{1:K}$.
This posterior assumes slots are independent.
Any appropriate existing approach can be used here (e.g., GENv2 or EMORL---non-exhaustive list, details can be found in their respective papers).

\textbf{Second Stage (Relate): }We use methods from set representation learning~\citep{vaswani2017attention,zaheer2017deep} to compute correlations between slots $\rvz_{o_{1:K}}$ and to encode this set for the scene posterior $q_\phi(\rvs \mid \rvx)$.
This posterior summarizes order-invariant object correlations.
We factorize $q_\phi(\rvs \mid \rvx)$ as a product of two simpler distributions:
\begin{equation}
    \label{eq:scene}
    q_\phi(\rvs \mid \rvx) = \int q_\phi(\rvs \mid f_\phi(\rvz_{o_{1:K}}))q_\psi(\rvz_{o_{1:K}} \mid \rvx) d \rvz_{o_{1:K}}.
\end{equation}
Unlike GNM, which directly infers $q_\phi(\rvs \mid \rvx)$, we use the slots as a latent bottleneck to factorize $q_\phi(\rvs \mid \rvx)$.
Therefore, we interpret $\rvs$ as a \emph{summary} of the slots and their correlations.
We use a single sample $\rvz_{o_{1:K}}$ to approximate this integral.
To compute $q_\phi(\rvs \mid f_\phi(\rvz_{o_{1:K}}))$, we use a permutation-invariant neural network $f_\phi$ that estimates a relational embedding for $\rvz_{o_{1:K}}$.
From this embedding, we compute the posterior mean and variance with two linear layers.
There are many ways to implement $f_\phi$---we process the slots with $L$ set embedding layers followed by a permutation-invariant DeepSets~\citep{zaheer2017deep} MLP with sum pooling aggregation.
Each set embedding layer is implemented as residual set self-attention~\citep{vaswani2017attention}.
To preserve the order-invariance of the segregation posterior, we remove the positional encodings and share weights across slots.

\subsection{Imagining the Slot Order}
\label{sec:imagine}
\begin{figure}[t]
    \centering
    \includegraphics[width=0.9\textwidth]{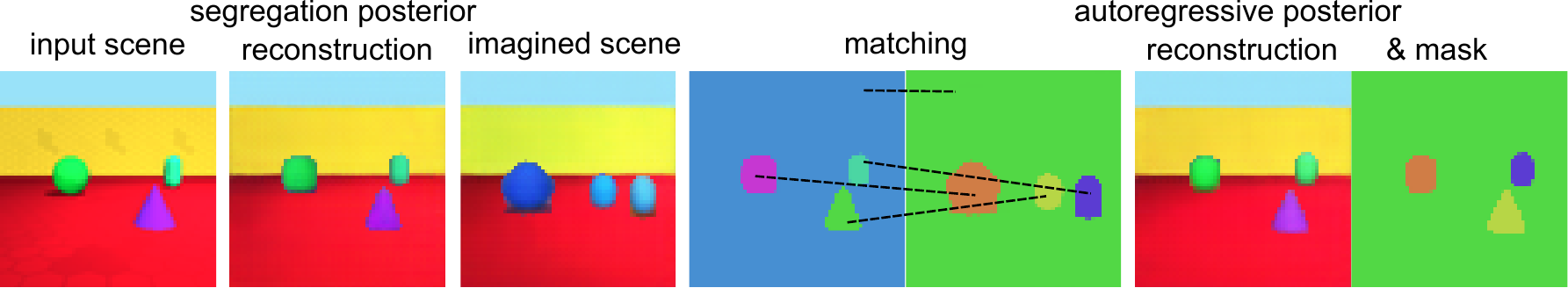}
    \caption{\textbf{Slot order alignment}. Imagined scenes are trained to closely resemble the input scene. For example, both scenes have three objects and similar camera poses. This allows a meaningful matching between the segregation slots and imagined slots to be computed. The matching extracts the slot order to use for autoregressive inference.}
    \label{fig:imagination_demo}
\end{figure}
So far, we have described the first two stages of SRI---first, for inferring a (randomly ordered) segregation posterior $q_\psi(\rvz_{o_{1:K}} \mid \rvx)$, and second, for inferring the scene posterior $q_{\phi}(\rvs \mid \rvx)$.
In the third stage, we must obtain an object order to transform the (randomly ordered) segregation slot posterior into an \emph{autoregressive} posterior.
We extract the learned order from a slot prior rollout.
Section~\ref{sec:training} describes a loss for training the slot prior to learn a \emph{consistent} object generation order.

\textbf{Third Stage (Imagine): }To obtain the slot order for inference, we first perform a scene-conditional imagination rollout to sample slots in order $\pi_{1:K}$:
\begin{equation}
    \label{eq:imagination}
    \rvs \sim q_\phi(\rvs \mid \rvx), \hspace{1em} \rvz_{\pi_1},\dots,\rvz_{\pi_K} \sim p_\theta(\rvz_{\pi_1} \mid \rvs),\dots,p_\theta(\rvz_{\pi_K} \mid \rvz_{\pi_{1:K-1}}, \rvs).
\end{equation}
We then extract the object order from $\rvz_{\pi_{1:K}}$ by solving for a matching between these slots and the randomly ordered segregation slots.
If the imagined scene is visually similar to the input, our intuition is that the imagined order will be a suitable order for inference on the input.
To encourage this similarity, we condition the rollout on the inferred scene variable (Figure~\ref{fig:imagination_demo}).

Extracting the order from the imagination rollout can be done efficiently with an $O(K)$ greedy matching (Appendix Algorithm~\ref{alg:matching}) that iteratively finds the best match for each slot in the rollout.
Alternatively, one could use the costlier Hungarian algorithm~\citep{munkres1957algorithms}.
Since a perfect alignment between the two sets is not guaranteed when using greedy matching, we use the notation $\hat{\pi}_{1:K}$ for the order obtained by permuting $o_{1:K}$ based on the matching.
We impose $\hat{\pi}_{1:K}$ on the randomly ordered posterior by permuting the posterior means, $\mu_{o_{1:K}} \rightarrow \mu_{\hat{\pi}_{1:K}}$, and predicting new correlated variances $\bar{\sigma}^2_{\hat{\pi}_{1:K}}$ based on the order.
The variances $   \bar{\sigma}^2_{\hat{\pi}_{1:K}}=\texttt{LSTM}([\mW_s^{\intercal}\rvs, \mu_{\hat{\pi}_1},\dots,\mu_{\hat{\pi}_{K-1}}])$ are a function of the (projected) global variable $\mW_s^{\intercal}\rvs$ and $\mu_{\hat{\pi}_{1:K-1}}$.
We do not update the permuted segregation posterior means $\mu_{\hat{\pi}_{1:K}}$ and only update the variances, which we find sufficient.
The scene-conditional autoregressive slot posterior is
\begin{align}
    q_\phi(\rvz_{\hat{\pi}_{1:K}} \mid \rvs, \rvx) = q_\phi(\rvz_{\hat{\pi}_1} \mid \rvs, \rvx) \prod_{k=2}^K q_\phi(\rvz_{\hat{\pi}_k} \mid \rvz_{\hat{\pi}_{1:k-1}}, \rvs, \rvx) := \prod_{k=1}^K \mathcal{N}( \mu_{\hat{\pi}_k}, \bar{\sigma}^2_{\hat{\pi}_k}).
\end{align}
To preserve order-invariance in the first two stages, we prevent gradients from flowing back through $\mu_{o_{1:K}}$ and $\rvz_{o_{1:K}}$.

\subsection{Training Losses} 
\label{sec:training}
We first introduce the auxiliary loss for training the slot prior to learn a consistent object generation order.
This loss is combined with a variational objective for training our hierarchical slot VAE.
The auxiliary loss is a reverse Kullback-Leibler (KL) divergence between the imagination rollout with order $\pi_{1:K}$ and autoregressive posterior with order $\hat{\pi}_{1:K}$.
Explicitly pushing the $K$-step rollout distribution $p_\theta(\rvz_{\pi_{1:K}} \mid \rvs)$ towards the transformed slot posterior trains the prior to rollout high quality scenes; consequentially, the prior learns to adopt a consistent generation order.
This loss also trains $f_\phi$ to extract correlations between slots $\rvz_{o_{1:K}}$.
Intuitively, if the imagined scene resembles the input scene, the scene variable $\rvs$ accurately captures correlations between objects in the input scene.
This ``cross order'' KL loss is
\begin{align}
     &\mathcal{L}_{\text{crossOrderKL}} = \mathbb{E}_{q_\phi(\rvs \mid \rvx)} \left[ \infdiv{q_\phi(\rvz_{\hat{\pi}_1} \mid \rvs, \rvx)}{p_\theta(\rvz_{\pi_1} \mid \rvs)} \right] \\
     &+ \mathbb{E}_{q_\phi(\rvs \mid \rvx)} \Biggl[ \sum_{k=2}^K   \mathbb{E}_{q_\phi(\rvz_{\hat{\pi}_{1:k-1}} \mid \rvs, \rvx)} \Bigl[\mathbb{E}_{{p_\theta(\rvz_{\pi_{1:k-1}} \mid \rvs)}} \bigl[ \infdiv{{\underbrace{q_\phi(\rvz_{\hat{\pi}_k} \mid \star)}_{\text{Order $\hat{\pi}_{1:K}$}}}}{ {\underbrace{p_\theta(\rvz_{\pi_k} \mid \star)}_{\text{Order $\pi_{1:K}$}}} } \bigr] \Bigr] \Biggr],\nonumber
\end{align}
where the symbol $\star$ is replaced as $q_\phi(\rvz_{\hat{\pi}_k} \mid \star) := q_\phi(\rvz_{\hat{\pi}_{k}} \mid \rvz_{\hat{\pi}_{1:k-1}}, \rvs, \rvx)$ and as {$p_\theta(\rvz_{\pi_k} \mid \star) := p_\theta(\rvz_{\pi_k} \mid \rvz_{\pi_{1:k-1}}, \rvs)$}. 

The negative ELBO objective for our hierarchical slot VAE is given by the sum of the following terms.
First, we have a negative log-likelihood (reconstruction error) term:
\begin{equation}
    \mathcal{L}_{\text{NLL}} = -\mathbb{E}_{q_\phi(\rvs \mid \rvx)}\left[ \mathbb{E}_{q_\phi(\rvz_{\hat{\pi}_{1:K}} \mid \rvs, \rvx)} \left[ \log p_\theta(\rvx \mid \rvz_{\hat{\pi}_{1:K}}) \right] \right].
\end{equation}
Next is a scene-level KL term:
\begin{equation}
    \label{eq:sceneKL}
    \mathcal{L}_{\text{sceneKL}} = \infdiv{q_\phi(\rvs \mid \rvx)}{p(\rvs)}.
\end{equation}
The slot-level reverse KL term $\mathcal{L}_{\text{slotKL}}$ resembles $\mathcal{L}_{\text{crossOrderKL}}$ except the inner expectation is taken with respect to the slot posterior with order $\hat{\pi}_{1:K}$.

See the appendix for the full derivation.
Overall, we minimize $\mathcal{L}$, the sum of the objectives for the segregation VAE ($\mathcal{L}_{\text{seg}}$) used in the Segregate stage and the SRI losses:
\begin{align}
    \label{eq:losses}
    \mathcal{L}_{\text{SRI}} &= \mathcal{L}_{\text{NLL}} + \mathcal{L}_{\text{sceneKL}} + \mathcal{L}_{\text{slotKL}} + \mathcal{L}_{\text{crossOrderKL}}\\
    \mathcal{L} &= \mathcal{L}_{\text{seg}} + \mathcal{L}_{\text{SRI}}.
\end{align}
We use the objective $\mathcal{L}_{\text{seg}}$ as defined by the segregation VAE.
Similar to previous models~\citep{pmlr-v139-emami21a,engelcke2021genesis} we dynamically balance NLL and KL with GECO~\citep{rezende2018taming}.
For simplicity, we use the same SBD decoder for $\mathcal{L}_{\text{seg}}$ and $\mathcal{L}_{\text{SRI}}$.

\section{Experiments}
\label{sec:experiments}
In this section we measure improvements in generation quality and correlation learning due to the scene-level variable and SRI inference.
\emph{Notably, we do not expect to see changes in segmentation quality}.
Section~\ref{sec:qual} qualitatively examines sample quality and the learned order, and Sections~\ref{sec:fixed-experiments} and \ref{sec:quant} quantitatively compares against key baselines. We conduct ablation studies in Section~\ref{sec:ablations}.
In the supplementary material we provide videos of random walks in the scene-level latent space to visualize higher-order object relationship learning.
Throughout, we use ``SRI'' to refer to the entire hierarchical slot VAE trained with SRI inference.
\footnote{Code and videos are available at \url{https://github.com/pemami4911/segregate-relate-imagine}}

\textbf{Datasets: }We use three synthetic multi-object datasets to evaluate SRI: \emph{Objects Room} and \emph{CLEVR6} from the standard Multi-Object Dataset~\citep{multiobjectdatasets19} and \emph{ShapeStacks}~\citep{groth2018shapestacks}. 
All three datasets consist of rendered images of 3D scenes containing 2-6 objects and have variable illumination, shadow, and camera perspectives.
Generating scenes from these datasets requires reasoning about non-trivial correlations between objects (e.g., occlusion in CLEVR6, objects resting on the floor in Objects Room, block towers in ShapeStacks).

\textbf{Setup: }On all datasets we train SRI with GENv2 as the segregation VAE using both Gaussian (SRI-G) and Mixture-of-Gaussians (SRI-MoG) likelihoods.
On CLEVR6 we also trained SRI with EMORL.
We fix $|\rvz_k| = 64$, $|\rvs| = 128$, and $L = 3$ across environments. 
Following standard practice, on CLEVR6 and Objects Room $K = 7$ and on ShapeStacks $K = 9$.
Remaining details for architectures, hyperparameters, and compute are in the appendix.

\textbf{Baselines: }We compare against state-of-the-art slot VAEs for scene generation GEN and GENv2, as well as an independent prior baseline EMORL.
Results for GEN and GENv2 (GENv2-MoG) are taken from~\citet{engelcke2021genesis} (both use the Mixture-of-Gaussians by default). 
GEN and GENv2-MoG results were missing for CLEVR6 so we trained these ourselves with official code releases.
We train GENv2 with the Gaussian likelihood (GENv2-G) on all three datasets to directly compare with SRI-G.
EMORL is trained using the official code release.

\subsection{Qualitative Evaluation}
\label{sec:qual}
\begin{figure}[t]
    \centering
    \begin{subfigure}{0.24\textwidth}
        \centering
        \includegraphics[trim=2.5cm 0 2.5cm 0,clip=true,scale=0.31]{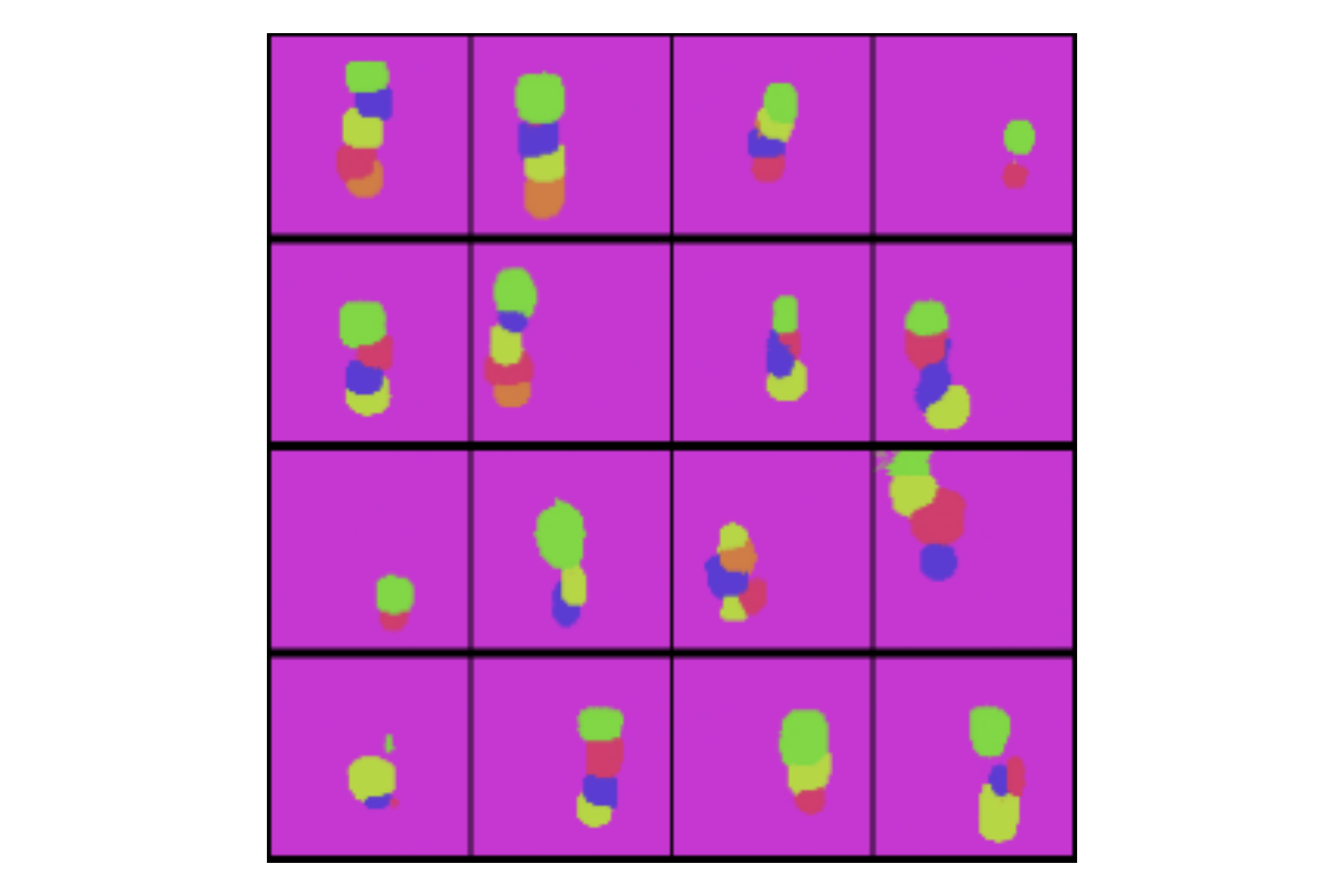}
        \caption{SRI-MoG\label{fig:slot_stability_sri_mog}}
    \end{subfigure}%
    \begin{subfigure}{0.24\textwidth}
        \centering
        \includegraphics[trim=2.5cm 0 2.5cm 0,clip=true,scale=0.31]{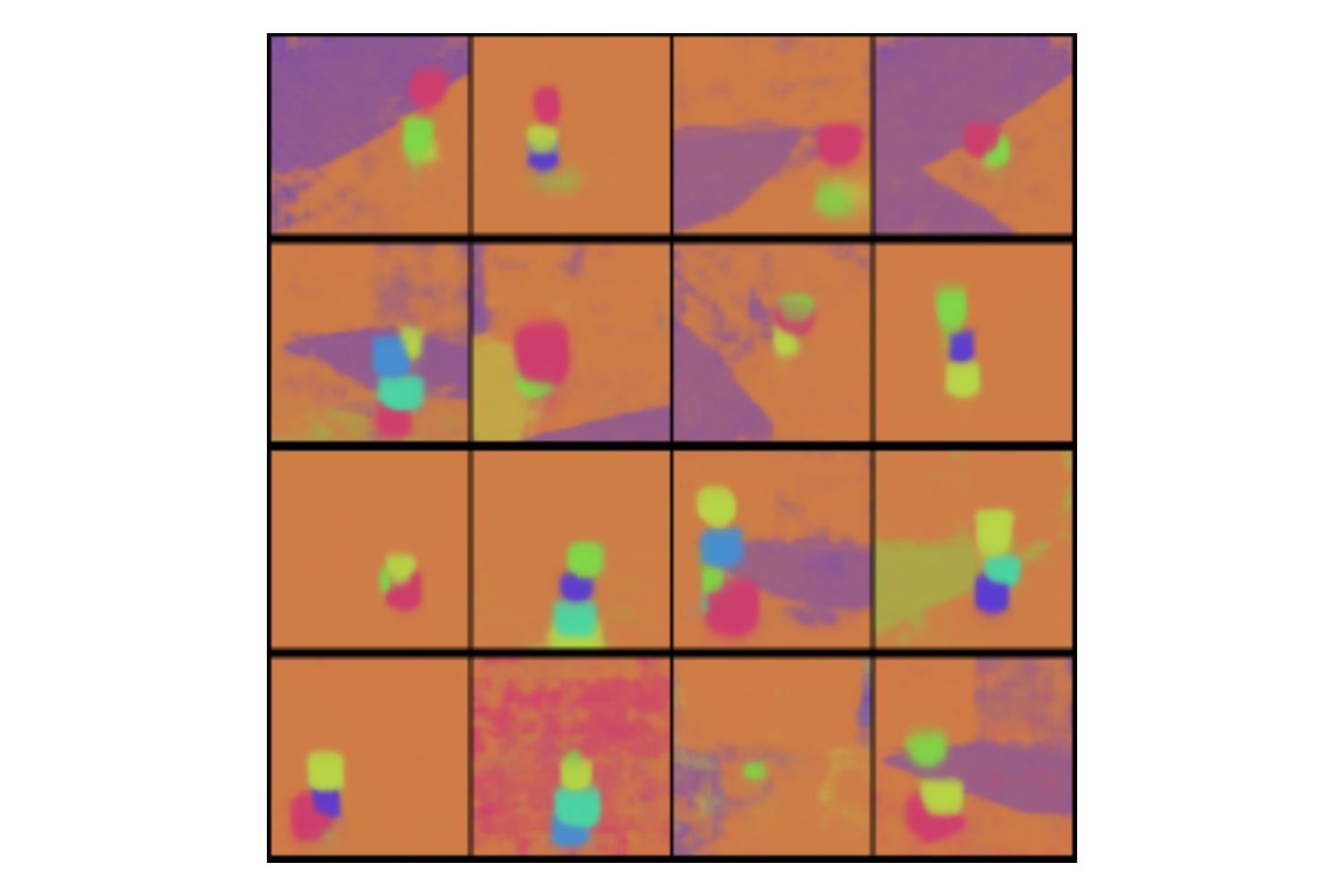}
        \caption{SRI-G\label{fig:slot_stability_sri_g}}
    \end{subfigure}%
    \begin{subfigure}{0.24\textwidth}
        \centering
        \includegraphics[trim=2.5cm 0 2.5cm 0,clip=true,scale=0.31]{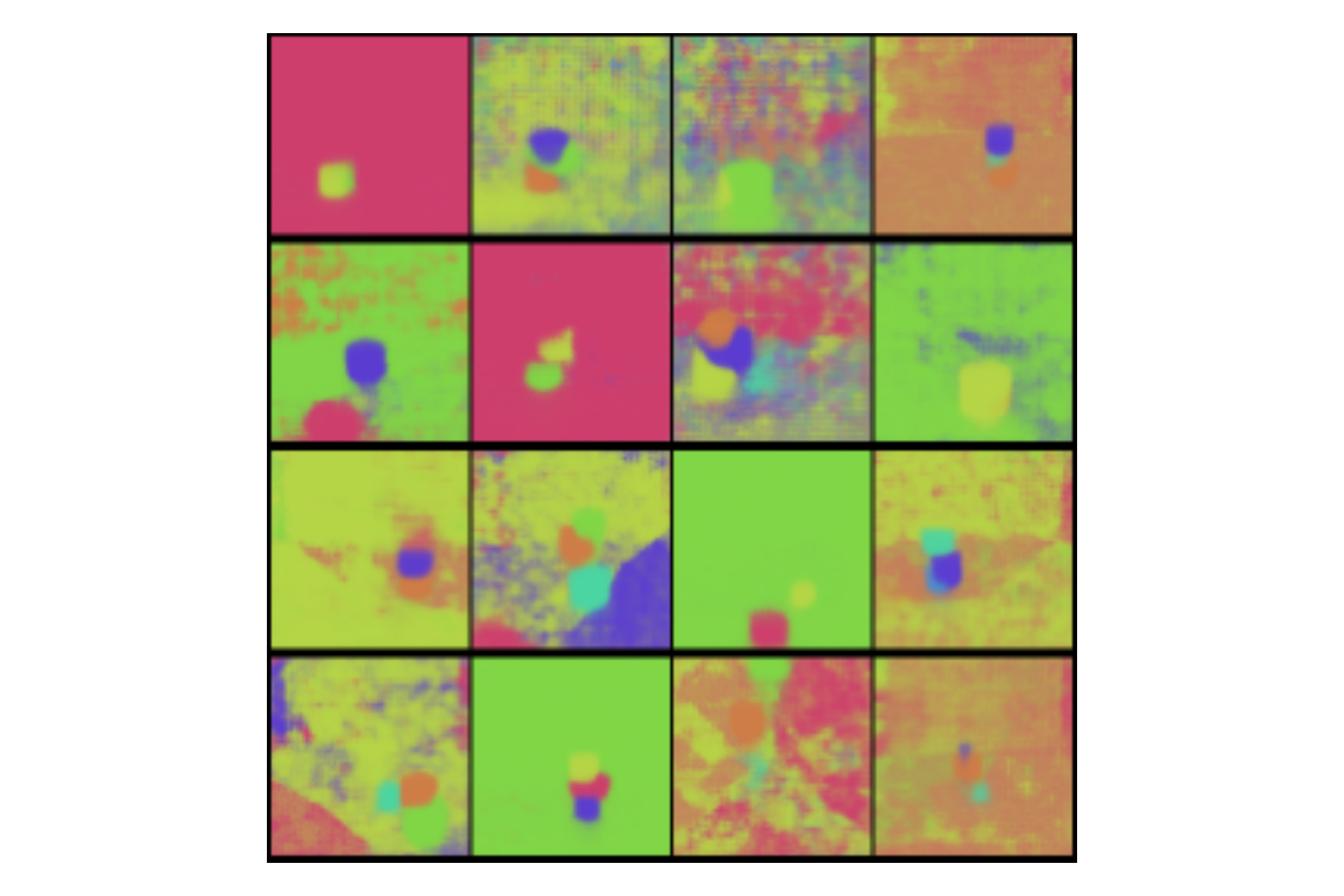}
        \caption{SRI-G*\label{fig:slot_stability_ablation}}
    \end{subfigure}%
    \begin{subfigure}{0.24\textwidth}
        \centering
        \includegraphics[trim=2.5cm 0 2.5cm 0,clip=true,scale=0.31]{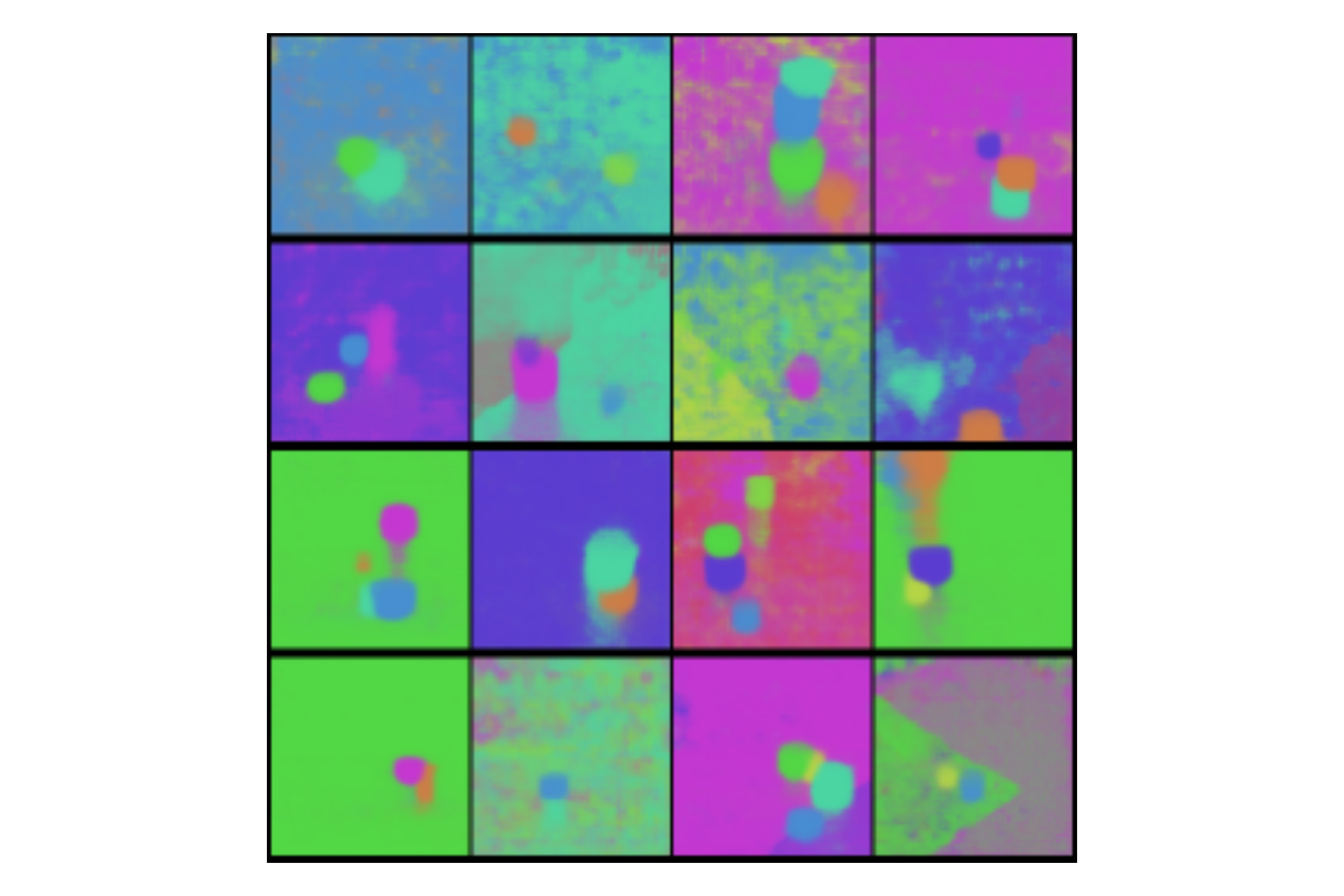}
        \caption{GENESISv2-G\label{fig:genv2_stability}}
    \end{subfigure}  
    \caption{\textbf{ShapeStacks object generation order}.  Slot number indicated by mask color (legend provided: Figure~\ref{fig:app:slot_color_palette}). Best viewed in color. a) SRI learns to generate objects in a consistent order. For this training run, the model uses slot 1 for the background and slots 4-9 for blocks (most towers have only 2-4 blocks). b) SRI-G uses slots 1 and 5 for background and 6-9 for blocks (occasionally 3-4). c) SRI-G* does not use the cross-order KL term $\mathcal{L}_{\text{crossOrderKL}}$. Objects are generated in random order with slots 5-9, which can be seen by the inconsistent use of mask colors. d) GENv2-G fails to learn a consistent order which hinders learning correlations. Viz. for Objects Room and CLEVR6 are in Figure~\ref{fig:app:SRI_MoG_qual}.}
    \label{fig:slot_order}
\end{figure}
\begin{figure}[t]
    \centering
    \includegraphics[width=0.85\textwidth]{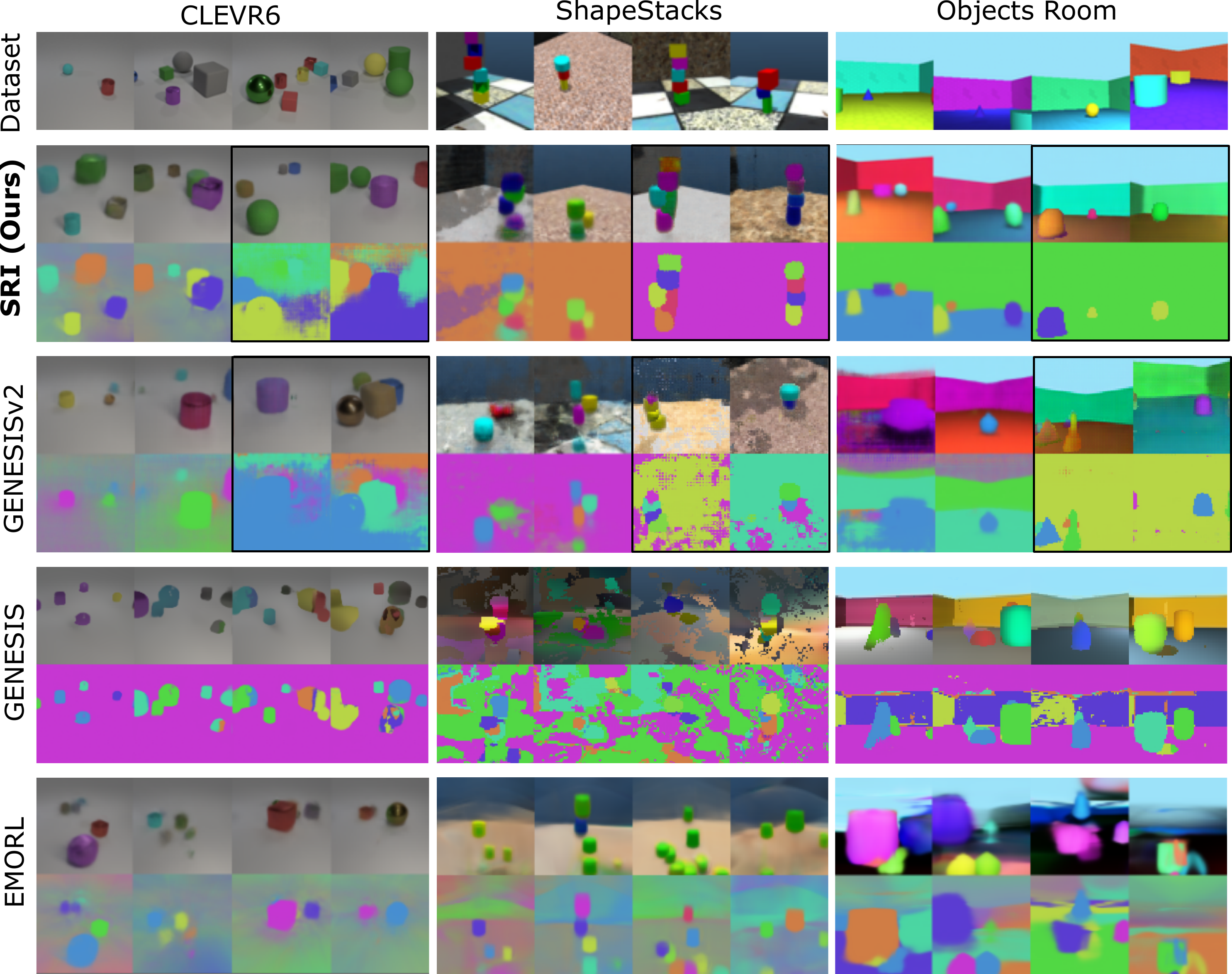}
    \caption{\textbf{Randomly generated scenes}. Samples from SRI-MoG and GENv2-MoG are outlined in black. Best viewed zoomed in and in color.\label{fig:qualitative}}
\end{figure}
We show that SRI learns a consistent generation order by visualizing segmentation masks for randomly generated ShapeStacks scenes  (Figure~\ref{fig:slot_order}).
By contrast, GENv2 fails to learn a consistent order (Figure~\ref{fig:genv2_stability}).
Random scenes sampled from each generative model are shown in Figure~\ref{fig:qualitative} with additional samples in the appendix (Figures~\ref{fig:app:SRI_MoG_qual},~\ref{fig:app:SRI_G_qual}).
Across environments, SRI generates higher quality scenes than GEN, GENv2, and EMORL (Figure~\ref{fig:qualitative}).
GENv2's samples contain artifacts and structural inaccuracies (e.g., extra walls floors in Objects Room and floating blocks in ShapeStacks). 
In Objects Room, GENv2-G segments the walls, floor, and sky across multiple slots, yet demonstrates a limited ability to learn the spatial correlation between these parts of the scene.
GENv2-MoG fares better on Objects Room because it segments the background into a single slot.
GEN's sequential attention occasionally settles on a semi-consistent order (e.g., placing the background in the first slot for CLEVR6 and Objects Room) but in general the learned order is arbitrary and its attention mechanism fails on challenging scenes (e.g., by trying to segment ShapeStacks scenes based on color).
Due to its independent slot prior, EMORL generates incoherent scenes.
See Appendix~\ref{sec:app:additional_results} for extra visualizations of random samples, test time generation with different numbers of slots, imagination rollouts, scene reconstructions, and scene segmentations.

\subsection{\revision{Comparison with Learned Slot Initialization}}
\label{sec:fixed-experiments}
\input{tables/NEURCOMP-D-23-00020R1-fixedorder}
\revision{In this section, we introduce SRI-SA*, a hierarchical autoregressive slot VAE with a modified Slot-Attention-based encoder (loosely inspired by SloTTAr~\citep{gopalakrishnan2023unsupervised} and compare it with SRI.
SRI-SA* tries to learn a left-to-right ordered segregation posterior $q_\psi(\rvz_{\pi_{1:K}} \mid \rvx)$ implemented by a modified Slot Attention module. 
To encourage a left-to-right ordering on the slots during inference, we replace the standard Normal slot initialization in Slot Attention with learnable mean and variance parameters per slot $(\mu_{1:K}, \sigma_{1:K}^2)$.
This pushes each slot to learn to specialize to a certain object in each image.
\emph{While this is likely insufficient to guarantee that a stable assignment of objects to slots will be learned, we implement this approach as a simple baseline for comparison.}
We map each slot to a mean and variance with a linear layer during inference. 
All else about the SRI-SA* model is the same as SRI-MoG and we train SRI-SA* using $\mathcal{L}_{SRI}$ (Equation~\ref{eq:losses}) without $\mathcal{L}_{\text{crossOrderKL}}$.
}

\begin{figure}[t]
    \centering
    \includegraphics[width=0.7\textwidth]{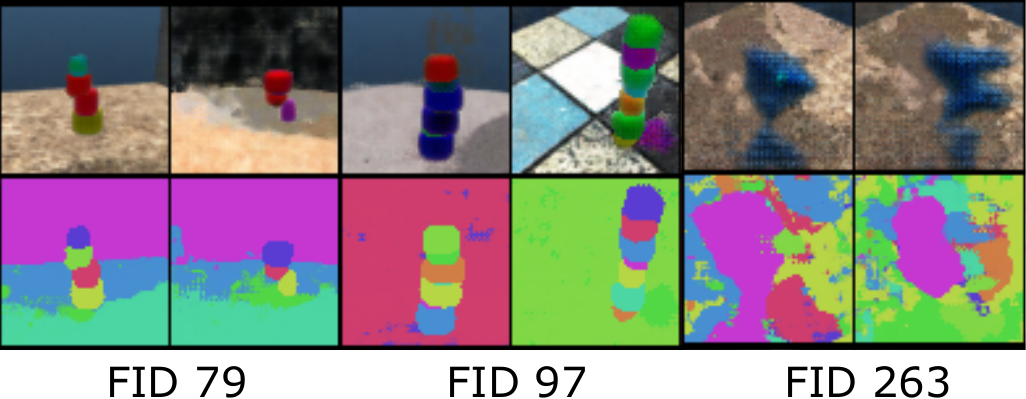}
    \caption{\revision{\textbf{SRI-SA* qualitative results.} We show two randomly generated scenes for each of the three models (each trained with a different random seed) and annotate them with the model's FID score. The best model (left) learns a consistent object ordering as indicated by the consistent use of colors for the background and foreground objects unlike the second best model (middle). We observe that all models try to use all slots for each image which causes oversegmentation. Order learning and generation quality is less stable across seeds and worse overall than SRI, which we attribute to the simplistic modified Slot Attention encoder.}}
    \label{fig:fixedorder}
\end{figure}

\revision{
The quantitative results are shown in Table~\ref{tab:fixedorder} and visualizations are in Figure~\ref{fig:fixedorder}.
Across random seeds, SRI outperforms SRI-SA* on all metrics and qualitatively.
SRI-SA* fails to consistently learn a good segregation strategy for ShapeStacks across seeds.
However, the best SRI-SA* model nearly matches the average performance of SRI.
This suggests that a more advanced fixed order inference approach could be a competitive alternative to our three-stage inference approach.
}

\subsection{Quantitative Evaluation}
\label{sec:quant}
\input{tables/NEURCOMP-D-23-00020R1-quant}
\input{tables/NEURCOMP-D-23-00020R1-ablations}
\input{tables/NEURCOMP-D-23-00020R1-scene_prior_ablations}
We use the Fr\'echet Inception Distance (\textbf{FID})~\citep{heusel2017gans} to quantify the visual similarity of generated scenes to training scenes. 
Inspired by the evaluation in GNM~\citep{jiang2020generative}, we measure the ability to learn higher-order correlations by manually labeling 100 ShapeStacks scenes generated by each model and computing the percent of scenes that contain a physically plausible stack of blocks, i.e., all blocks in the stack are touching each other (structural accuracy, or \textbf{S-Acc}). 
We use officially released model weights to compute S-Acc for GEN and GENv2-MoG.
Following standard practice for evaluating segregation quality we use the adjusted rand index~\citep{rand1971objective,hubert1985comparing} for foreground objects only (\textbf{ARI-FG}).

Results are in Table~\ref{tab:quant}.
Out of the slot VAEs, SRI achieves the lowest FID scores on all environments and the best S-Acc score.
SRI-G improves GENv2-G's FID by an average of 24\% and SRI-MoG improves GENv2-MoG's FID by an average of 19\%.
On CLEVR6, we observe that SRI achieves the smallest improvement in FID over GENv2, which we believe is caused by \emph{a)} CLEVR6 only having simple correlations to learn (e.g., occlusion) and \emph{b)} small artifacts introduced by GENv2's encoder which seems to negatively impact the FID score on this dataset. 
GENv2-G segregates the walls, floor, and sky into different slots on Objects Room which makes correlation learning difficult, leading to a poor FID score and which SRI-G improves by 36\%.
In general, we find that the segregation quality of SRI closely matches that of the slot VAE used for scene segregation, and that better ARI-FG correlates with better sample quality for SRI.

To demonstrate the flexibility of SRI, we also train SRI using EMORL to extract randomly ordered slots on CLEVR6.
We see a significant improvement in FID from EMORL to SRI-EMORL (244{\tiny{$\pm 19$}} to 48.9{\tiny{$\pm 7$}}, also see Figure~\ref{fig:app:SRI_EMORL}). 

\subsection{Ablation Studies}
\label{sec:ablations}
We now examine each contribution individually with ShapeStacks's validation set. 

\textbf{Scene-level variable: }First, we isolate the scene-level variable by adding a scene-level variable to GEN and GENv2, leaving all else with these models the same.
For scene-level inference with GEN, we encode the ordered sequence of slots obtained by sampling from its ordered slot posterior.
Scene-level inference for GENv2 is the same as with SRI.
Results are in Table~\ref{tab:scene_ablations}.
Augmenting GEN with a scene variable does little to improve its performance since it still fails to properly segregate ShapeStacks scenes.
GENv2 enjoys a moderate improvement in FID score and S-Acc but does not match the performance of SRI---the latter which has the combined benefit of both the scene variable and of using a consistent object order.
\revision{The higher negative ELBO values for SRI are due to the additional KL terms.}
These results suggest the scene variable alone does not account for all of SRI's demonstrated performance improvements.

\textbf{SRI Inference: }Next, we ablate the core aspects of SRI inference using the SRI-G model (Table~\ref{tab:ablations}).
\textit{a)} We remove the scene-level variable. The prior for SRI's first slot becomes $p(\rvz_1) := \mathcal{N}(0,1)$.
Without a scene-level variable, sample quality is significantly worse.
SRI inference no longer works properly; the imagination rollout is now \emph{unconditional} and imagined scenes are not similar to the input $\rvx$. 
The model fails to learn a consistent generation order to align the slot segregation posterior to.
\textit{b)} We ablate the order alignment step of SRI inference. 
That is, during inference we never transform the first stage slot posterior with random slot order $o_{1:K}$ into an autoregressive posterior with imagined order $\hat{\pi}_{1:K}$ (we also therefore remove $\mathcal{L}_{\text{crossOrderKL}}$ from the loss).
SRI with a randomly ordered slot posterior has difficulty learning object correlations and the sample quality is poor.
\textit{c)} If we only remove $\mathcal{L}_{\text{crossOrderKL}}$, SRI's prior fails to learn a consistent order and correlation estimation suffers. 
This results in a significant drop in sample quality (see Figure~\ref{fig:slot_stability_ablation}).
Overall, we find that the scene-level variable, order alignment, and auxiliary loss all play a critical role for SRI inference.

\section{Conclusion}
In this work we demonstrated strategies for improving the ability of autoregressive slot VAEs to learn multi-object correlations, which resulted in gains in scene generation quality.
To achieve this, we proposed augmenting slots with a scene-level latent variable and learning a consistent object order.
These improvements are agnostic to the choice of model that initially extracts the slots from images.
\revision{Recently, it was shown that frozen image encoders pretrained with self-supervised learning greatly improve the robustness of slot-based models, particularly on real world scenes~\citep{seitzer2022bridging}. 
We believe exploring the use of frozen encoders with SRI is a promising direction.}

One limitation of SRI inference is that it does not learn causal relations between objects but rather only correlations, which only weakly generalize outside the training distribution~\citep{scholkopf2021toward}.
Extending SRI to dynamic settings would enable exploring causal inference of dependencies.
Another limitation of SRI is it requires solving a computationally demanding matching problem during inference.
We suggest a linear-time approximate matching algorithm in this work, but this may scale poorly to more complex environments.
Finally, there appears to be only limited benefits to learning an object ordering for environments such as CLEVR6 where scenes have little structure and objects are only weakly correlated. 
However, learning an order does not seem to hurt performance in this situation.

There are not immediate negative societal implications of this work.
However, future work on deep generative modeling that builds on our ideas could plausibly cause harm. 
Research on mitigation strategies for these scenarios is ongoing~\citep{mishkin2022risks}.

\subsection*{Acknowledgments}
We thank Dave Biagioni and Peter Graf for providing valuable comments and suggested revisions. 
Patrick Emami was supported in parts by the Florida DOT under grant BDV31 977-116, an FEF McKnight fellowship, and the National Renewable Energy Laboratory, operated by Alliance for Sustainable Energy, LLC, for the U.S. Department of Energy (DOE) under Contract No. DE-AC36-08GO28308. 
This work was supported by the Laboratory Directed Research and Development (LDRD) Program at NREL. The views expressed in the article do not necessarily represent the views of the DOE or the U.S. Government. 
The U.S. Government retains and the publisher, by accepting the article for publication, acknowledges that the U.S. Government retains a nonexclusive, paid-up, irrevocable, worldwide license to publish or reproduce the published form of this work, or allow others to do so, for U.S. Government purposes.

\bibliography{NEURCOMP-D-23-00020R1}

\newpage
\appendix
\section{Appendix}

\subsection{Matching Pseudocode}
\input{algorithms/NEURCOMP-D-23-00020R1-matching}
Alternative matching algorithms such as the Hungarian algorithm~\citep{munkres1957algorithms} can be used instead at a higher computational cost; however, we leave an empirical comparison of these algorithms for future work.

\subsection{SRI Inference Pseudocode}
\input{algorithms/NEURCOMP-D-23-00020R1-inference}
Our proposed inference algorithm involves \textit{aligning} the object order of the segregation slot posterior to a learned object order (Lines 6-7) and then transforming this posterior into an autoregressive distribution by predicting a sequence of correlated variances (Lines 8-10).

\subsection{ELBO Derivation}

We derive a lower bound on the log marginal scene likelihood as follows:
\begin{align}
    \log p(\rvx) &= \log \int p_\theta(\rvx, \rvs, \rvz_{\pi_{1:K}}) d\rvs, d\rvz_{\pi_{1:K}}\\
    &= \log \int \frac{q_\phi(\rvs, \rvz_{\pi_{1:K}} \mid  \rvx) }{q_\phi(\rvs, \rvz_{\pi_{1:K}} \mid \rvx)} p_\theta(\rvx, \rvs, \rvz_{\pi_{1:K}}) d\rvs, d\rvz_{\pi_{1:K}}\\
    &= \log \mathbb{E}_{q_\phi(\rvs, \rvz_{\pi_{1:K}} \mid \rvx)}\left[ \frac{p_\theta(\rvx, \rvs, \rvz_{\pi_{1:K}})}{q_\phi(\rvs, \rvz_{\pi_{1:K}} \mid \rvx)} \right]. 
\end{align}
Applying Jensen's inequality and then factorizing gives:
\begin{align}
    \log p(\rvx) &\geq \mathbb{E}_{q_\phi(\rvs, \rvz_{\pi_{1:K}} \mid \rvx)}\left[ \log \frac{p_\theta(\rvx, \rvs, \rvz_{\pi_{1:K}})}{q_\phi(\rvs, \rvz_{\pi_{1:K}} \mid \rvx)} \right] \\
    &= \mathbb{E}_{q_\phi(\rvs \mid \rvx)} \left[  \mathbb{E}_{q_\phi(\rvz_{\pi_{1:K}} \mid \rvs, \rvx)} \left[ \log \frac{ p(\rvs) p_\theta(\rvz_{\pi_{1:K}} \mid \rvs) p_\theta(\rvx \mid \rvz_{\pi_{1:K}}) }{q_\phi(\rvs \mid \rvx)  q_\phi(\rvz_{\pi_{1:K}} \mid \rvs, \rvx)} \right]  \right]. \label{eq:app:elbo}
\end{align}
We use the fact that $\log(\frac{A}{B} \frac{C}{D}) = \log(\frac{A}{B}) + \log(\frac{C}{D})$ to factor Eq.~\ref{eq:app:elbo} into a sum of three losses.
The autoregressive slot posterior order obtained for inference {$\color{blue}{\hat{\pi}_{1:K}}$}  is shown in blue.

The first term is a negative log-likelihood loss:
\begin{align}
    \mathcal{L}_{\text{NLL}} = -\mathbb{E}_{q_\phi(\rvs \mid \rvx)} \left[ \mathbb{E}_{q_\phi(\rvz_{\color{blue}{\hat{\pi}_{1:K}}} \mid \rvs, \rvx)} \left[ \log p_\theta(\rvx \mid \rvz_{\color{blue}{\hat{\pi}_{1:K}}}) \right] \right].
\end{align}
The second is a slot-level reverse KL loss with the approximate order:
\begin{align}
    &\mathcal{L}_{\text{slotKL}} = -\mathbb{E}_{q_\phi(\rvs \mid\rvx)}\left[ \mathbb{E}_{q_\phi(\rvz_{\color{blue}{\hat{\pi}_{1:K}}} \mid \rvs, \rvx)} \left[ \log \frac{p_\theta(\rvz_{\color{blue}{\hat{\pi}_{1:K}}} \mid \rvs)}{q_\phi(\rvz_{\color{blue}{\hat{\pi}_{1:K}}} \mid \rvs, \rvx)} \right] \right] \\
    &= \mathbb{E}_{q_\phi(\rvs \mid\rvx)}\left[ \mathbb{E}_{q_\phi(\rvz_{\color{blue}{\hat{\pi}_{1:K}}} \mid \rvs, \rvx)} \left[ \log \frac{q_\phi(\rvz_{\color{blue}{\hat{\pi}_{1:K}}} \mid \rvs, \rvx)}{p_\theta(\rvz_{\color{blue}{\hat{\pi}_{1:K}}} \mid \rvs)} \right] \right] \\
    &= \mathbb{E}_{q_\phi(\rvs \mid\rvx)}\left[ \sum_{k=1}^K \mathbb{E}_{q_\phi(\rvz_{\color{blue}{\hat{\pi}_{1:k}}} \mid \rvs, \rvx)} \left[ \log \frac{q_\phi(\rvz_{\color{blue}{\hat{\pi}_k}} \mid \rvz_{\color{blue}{\hat{\pi}_{1:k-1}}}, \rvs, \rvx)}{p_\theta(\rvz_{\color{blue}{\hat{\pi}_k}} \mid \rvz_{\color{blue}{\hat{\pi}_{1:k-1}}}, \rvs)} \right] \right]\\
    &= \mathbb{E}_{q_\phi(\rvs \mid \rvx)}\left[ \infdiv{q_\phi(\rvz_{\color{blue}{\hat{\pi}_1}} \mid \rvs, \rvx)}{p_\theta(\rvz_{\color{blue}{\hat{\pi}_1}} \mid \rvs)} \right]\\
    &\phantom{= }+ \mathbb{E}_{q_\phi(\rvs \mid \rvx)}\left[ \sum_{k=2}^K \mathbb{E}_{q_\phi(\rvz_{\color{blue}{\hat{\pi}_{1:k-1}}} \mid \rvs, \rvx)} \left[ \infdiv{q_\phi(\rvz_{\color{blue}{\hat{\pi}_{k}}} \mid \rvz_{\color{blue}{\hat{\pi}_{1:k-1}}}, \rvs, \rvx)}{p_\theta(\rvz_{\color{blue}{\hat{\pi}_{k}}} \mid \rvz_{\color{blue}{\hat{\pi}_{1:k-1}}}, \rvs)} \right] \right].\label{eq:app:kl-1}
\end{align}
The third term is a scene-level reverse KL loss:
\begin{align}
    \mathcal{L}_{\text{sceneKL}} &= -\mathbb{E}_{q_\phi(\rvs \mid \rvx )}\left[ \mathbb{E}_{q_\phi(\rvz_{\pi_{1:K}} \mid \rvs, \rvx)} \left[ \log \frac{p(\rvs)}{q_\phi(\rvs \mid \rvx )} \right] \right] \\ 
    &= -\mathbb{E}_{q_\phi(\rvs \mid \rvx )}\left[ \log \frac{p(\rvs)}{q_\phi(\rvs \mid \rvx )} \right]  \\
    &= \mathbb{E}_{q_\phi(\rvs \mid \rvx )}\left[ \log \frac{q_\phi(\rvs \mid \rvx) }{p(\rvs)} \right]  \\
    &= \infdiv{q_\phi(\rvs \mid \rvx )}{p(\rvs)}.
\end{align}
We also define an auxiliary slot-level reverse KL loss similar to Eq.~\ref{eq:app:kl-1} except that the inner expectation is now taken with respect to the slot posterior and the $K$-step rollout with order $\pi_{1:K}$:
\begin{align}
    &\mathcal{L}_{\text{crossOrderKL}} = \mathbb{E}_{q_\phi(\rvs \mid \rvx)}\left[ \infdiv{q_\phi(\rvz_{\color{blue}{\hat{\pi}_1}} \mid \rvs, \rvx)}{p_\theta(\rvz_{\pi_1} \mid \rvs)} \right]\\
    &\phantom{= }+ \mathbb{E}_{q_\phi(\rvs \mid \rvx)}\left[ \sum_{k=2}^K \mathbb{E}_{q_\phi(\rvz_{\color{blue}{\hat{\pi}_{1:k-1}}} \mid \rvs, \rvx)} \left[ \mathbb{E}_{p_\theta(\rvz_{\pi_{1:k-1}} \mid \rvs )} \left[ \infdiv{q_\phi(\rvz_{\color{blue}{\hat{\pi}_{k}}} \mid \star)}{p_\theta(\rvz_{\pi_k} \mid \star)} \right] \right] \right],
\end{align}
where $q_\phi(\rvz_{\color{blue}{\hat{\pi}_{k}}} \mid \star) := q_\phi(\rvz_{\color{blue}{\hat{\pi}_{k}}} \mid \rvz_{\color{blue}{\hat{\pi}_{1:k-1}}}, \rvs, \rvx)$ and $p_\theta(\rvz_{\pi_k} \mid \star) := p_\theta(\rvz_{\pi_{k}} \mid \rvz_{\pi_{1:k-1}}, \rvs)$.
The auxiliary KL term is always non-negative by definition, so adding it to the negative ELBO only increases the upper bound on $-\log p(\rvx)$ (equivalently, loosens the lower bound on $\log p(\rvx)$).
Summed together, the four terms form the SRI loss: $\mathcal{L}_{\text{SRI}} = {\mathcal{L}}_{\text{NLL}} + \mathcal{L}_{\text{slotKL}} + \mathcal{L}_{\text{crossOrderKL}} + \mathcal{L}_{\text{sceneKL}}$.
In practice, we approximate the expectations with a single sample and average the loss over a minibatch of size $B = 32$ of dataset samples. We leave a formal characterization of how the accuracy of the matching between $\hat{\pi}$ and $\pi$ affects the lower bound on the marginal log likelihood for future work.
Intuitively, as long as the two slot orders are ``close'', we can expect these lower bounds to be similar as well.  

\subsection{Experiment details}

\subsubsection{Hyperparameters}

\textbf{Architecture: }We use  $|\rvz_k| = 64$ for all environments following GENv2 and EMORL. We set $|\rvs| = 128$ to be twice the slot dimension.
The number of attention layers for scene posterior estimation is fixed at $L = 3$; we found that varying $L$ incurred little change in performance.
We use the same hyperparameters for the image encoder and decoder architectures as used by GENv2 and EMORL for SRI-G, SRI-MoG, and SRI-EMORL, respectively.
Both of the GENv2 and EMORL implementations are based on open source released by the respective authors.\footnote{\url{https://github.com/pemami4911/EfficientMORL}}\footnote{\url{https://github.com/applied-ai-lab/genesis}}
For SRI-EMORL we fix the number of iterative refinement steps to 2.
Following GEN and GENv2, SRI's autoregressive prior and posterior LSTM uses 256 hidden units. 
When training GENv2 jointly with SRI we replace its autoregressive prior with a slot-wise independent prior.
SRI uses GELU non-linear activations~\citep{hendrycks2016gaussian} after each linear projection of the scene variable and in the attention layers. 
All layers use Xavier weight initialization~\citep{glorot2010understanding}.

In the scene encoder, each set embedding layer is defined as follows:
\begin{align}
    \label{eq:value_computation}
     &\alpha = \texttt{softmax}\Bigl(\frac{(\rvz_{o_{1:K}} \mW_k)(\rvz_{o_{1:K}} \mW_q)^{\intercal}}{\sqrt{z}} \Bigr),
     \hspace{1em}  \hat{\rvz}_{o_{i}} = \sum_{j=1}^K \alpha_{ij} \bigl( \mW_v^{\intercal} [\rvz_{o_j}; \rvz_{o_i} - \rvz_{o_j}] \bigr),\\
     \label{eq:residual}
    &\rvz'_{o_i} =  \rvz_{o_i} + g_\phi(\rvz_{o_i}, \rvz_{o_{1:K}}), \hspace{1em}  g_\phi(\rvz_{o_i}, \rvz_{o_{1:K}}) := \texttt{MLP}(\texttt{LayerNorm}(\rvz_{o_i} + \hat{\rvz}_{o_{1:K}}).
\end{align}
Here, $\mW_q,\mW_k \in \mathbb{R}^{z \times z}, \mW_v \in \mathbb{R}^{2z \times z}$, $[;]$ is concatenation, and $\alpha$ is the $K \times K$ scaled dot-product self-attention computed with keys and queries given by linear projections of $\rvz_{o_{1:K}}$.
To potentially create a stronger inductive bias for encoding directed dependencies, we used the difference between slots $\rvz_{o_i} - \rvz_{o_j}$ as an additional feature in Eq.~\ref{eq:value_computation}~\citep{santoro2017simple}.
However, empirically we did not observe any improvements from this modification.

\textbf{Optimization: }We mostly adopt the same hyperparameters here as used by GENv2 and EMORL.
SRI based on GENv2 uses the Adam optimizer~\citep{kingma2014adam} with default hyperparameters and a learning rate of 1e-4 but without any learning rate schedule.
SRI-EMORL uses an initial learning rate of 4e-4, grad norm clipping to 5, and learning rate schedule consisting of a linear warmup for 10K steps then multiplicative decay with by a rate of 0.5 every 100K steps.
A batch size of 32 is used for all models.
Both GENv2 and EMORL use GECO to balance reconstruction and KL during optimization; we discuss the GECO hyperparameters in Section~\ref{sec:app:geco}.

\textbf{Datasets: }
Both CLEVR6 and Objects Room can be accessed freely online under the Apache License 2.0.\footnote{\url{https://github.com/deepmind/multi_object_datasets}}
The ShapeStacks dataset is also freely available for download online under the GNU License 3.0.\footnote{\url{https://ogroth.github.io/shapestacks/}}.
We use the same preprocessing protocol for CLEVR6 as~\citet{pmlr-v139-emami21a}, which is to center crop the images to 192x192 and the resize them to size 96x96. Objects Room and ShapeStacks have 64x64 RGB images.

\subsubsection{Balancing reconstruction and KL with GECO}
\label{sec:app:geco}
Both GENv2 and EMORL use GECO to balance  the reconstruction and KL losses~\citep{rezende2018taming}. 
Each code base has its own GECO implementation and GECO hyperparameter schedule which we use to train SRI-G, SRI-MoG, and SRI-EMORL.

SRI-G (unnormalized Gaussian image likelihood and global $\sigma$ = 0.7) uses a per-pixel and per-channel GECO target of -0.353 on Objects Room and ShapeStacks and -0.356 on CLEVR6.
We decreased the GECO learning rate from 1e-5 to 1e-6.
Once the GECO target is reached the learning rate is multiplied by a speedup factor of 10 to accelerate the decay of the GECO Lagrange parameter.

SRI-MoG uses the same GECO hyperparameters as GENv2-MoG on ShapeStacks and Objects Room (target is 0.5655---note that the MoG likelihood here is normalized) except we also use the decreased GECO learning rate of 1e-6.
However, on Objects Room we notice that this increasesed training instability, likely because this warms up the KL less aggressively than the faster GECO learning rate does.
We use restarting from model checkpoints to mitigate this here.

For SRI-EMORL on CLEVR6 we used a target of -2.265 for unnormalized Gaussian image likelihood with global $\sigma$ = 0.1.
The rule of thumb used to tune the GECO target is that the target should be reached after about 20\% of the training steps.
This gives ample time for the GECO Lagrange parameter to automatically decay back to 1 so that a valid ELBO is eventually maximized.
We use a GECO learning rate of 1e-6 and speedup of 10 as well.

\subsubsection{Baselines and compute}
\textbf{GENESIS and GENESIS-v2: }The authors of GENv2 have released pre-trained weights for GEN trained on ShapeStacks and GENv2-MoG trained on Objects Room and ShapeStacks.
We use these weights for model visualizations and to compute the structure accuracy metric.
To train GENv2-MoG on CLEVR6 we had to lower the standard deviation to 0.1.
We adjusted the GENv2-MoG GECO target accordingly by tuning it to -2.265.
To compute the FID score, we follow the same protocol as GENv2~\citep{engelcke2021genesis} and use 10K real and generated samples.


\textbf{Compute: }In general, the majority of the compute is taken up by the segregation VAE (GENv2 and EMORL in this work). SRI adds a few lightweight neural networks and therefore only marginally increases training time.

\begin{itemize}
\item On Objects Room and ShapeStacks, SRI-X takes about 20 hours to reach 500K steps using 2 NVIDIA A100 GPUs. On CLEVR6, it about 24 hours to reach 425K steps, at which point we stopped training since the model showed signs of convergence and so as to keep training times at about one day or faster on equivalent hardware to ease reproducibility.
\item The GENv2 and GEN baselines are negligibly faster to train. We train these baselines using the same setup as SRI.
\item SRI-EMORL uses 8 NVIDIA A100 GPUs to reach 225K CLEVR6 which takes about 27 hours, at which point we cut off training to keep a similar compute budget to SRI with GENv2. The memory footprint of the two steps of iterative refinement used to estimate the segregation posterior is large, hence the need for 8 GPUs.
\end{itemize}

\subsubsection{Open source software}

This project was conducting using the following open source Python packages:
PyTorch~\citep{torch}, numpy~\citep{numpy}, jupyter~\citep{jupyter}, matplotlib~\citep{matplotlib}, scikit-learn~\citep{sklearn}, and sacred~\citep{greff2017sacred}.

\subsection{Additional Qualitative Results}
\label{sec:app:additional_results}

\begin{figure}[h]
    \centering
    \includegraphics[scale=0.6]{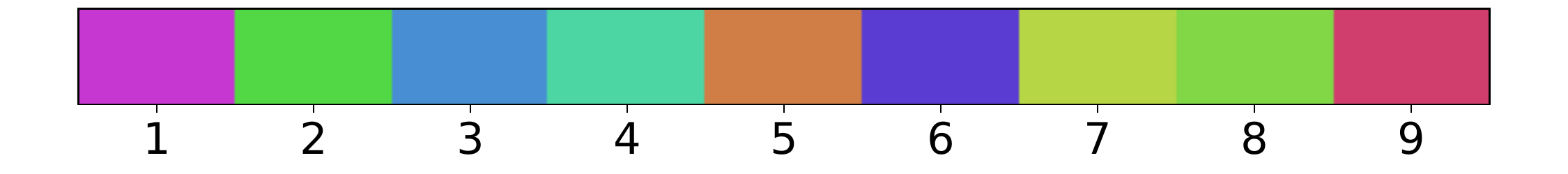}
    \caption{Mask color to slot number legend}
    \label{fig:app:slot_color_palette}
\end{figure}

\begin{figure}[t]
    \centering
    \includegraphics[scale=1.4]{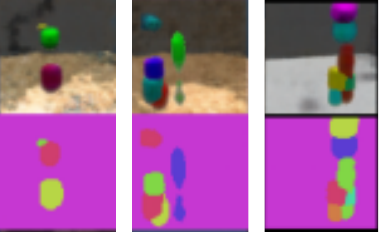}
    \caption{\revision{\textbf{Failure cases.} Examples of  ShapeStacks scenes randomly sampled from SRI-MoG that are of poor quality. Floating blocks and blocks occupying the same space are hallmarks of failures.  \label{fig:app:failures}}}
\end{figure}

\revision{\textbf{Failure cases: } We show failed scene generations from SRI-MoG in Figure~\ref{fig:app:failures}. Future directions that could help improve scene generation quality include making the model end-to-end differentiable and using a pretrained image encoder.}

\begin{figure}[t]
    \centering
    \begin{subfigure}{0.48\textwidth}
        \centering
        \includegraphics[trim=2.5cm 0 2.5cm 0,clip=true,scale=0.5]{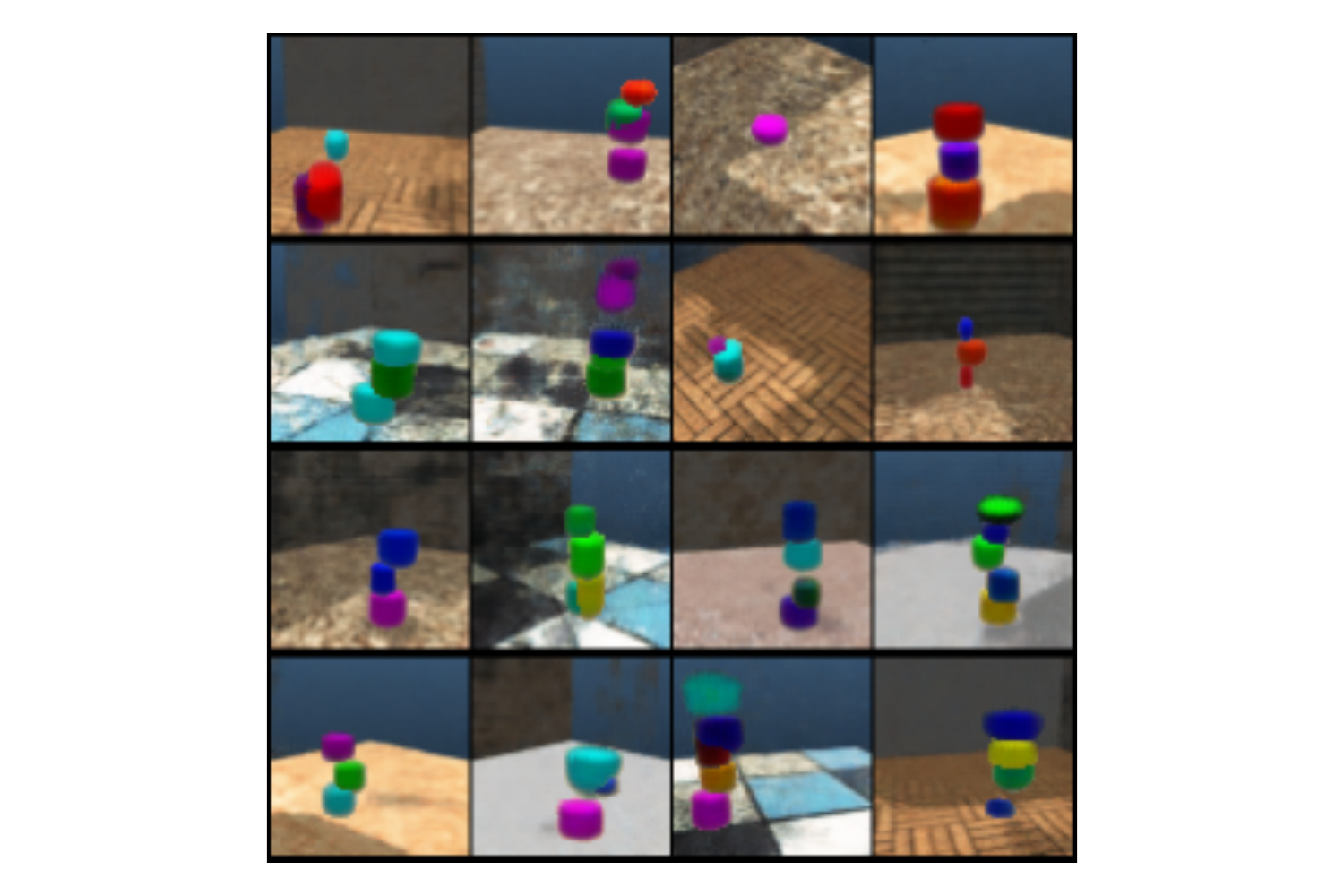}
        \caption{ShapeStacks images\label{fig:app:shapestacks_4x4}}
    \end{subfigure}
    \begin{subfigure}{0.48\textwidth}
        \centering
        \includegraphics[trim=2.5cm 0 2.5cm 0,clip=true,scale=0.5]{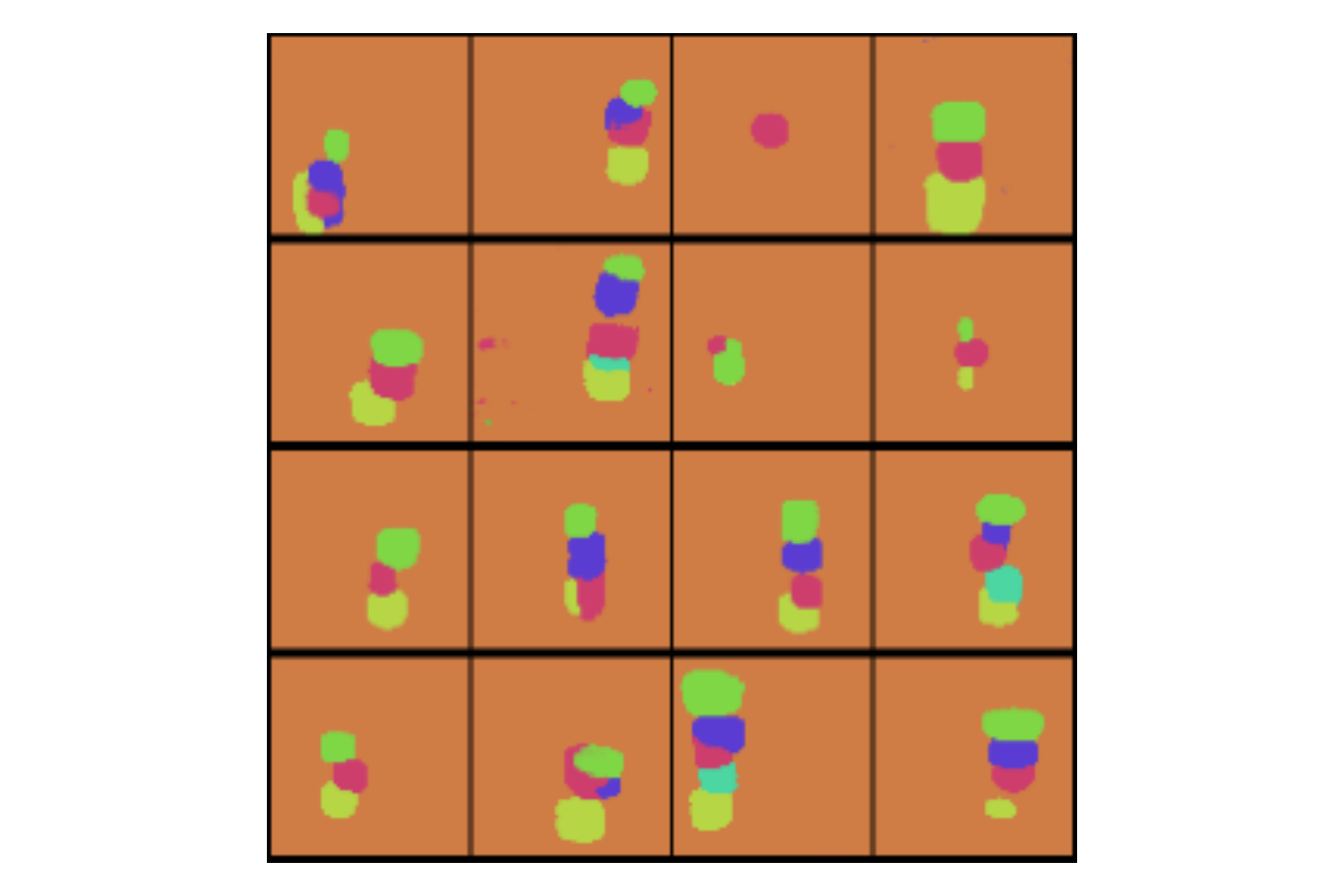}
        \caption{ShapeStacks masks\label{fig:app:shapestacks_4x4_masks}}
    \end{subfigure}
    \qquad
    \begin{subfigure}{0.48\textwidth}
        \centering
        \includegraphics[trim=2.5cm 0 2.5cm 0,clip=true,scale=0.5]{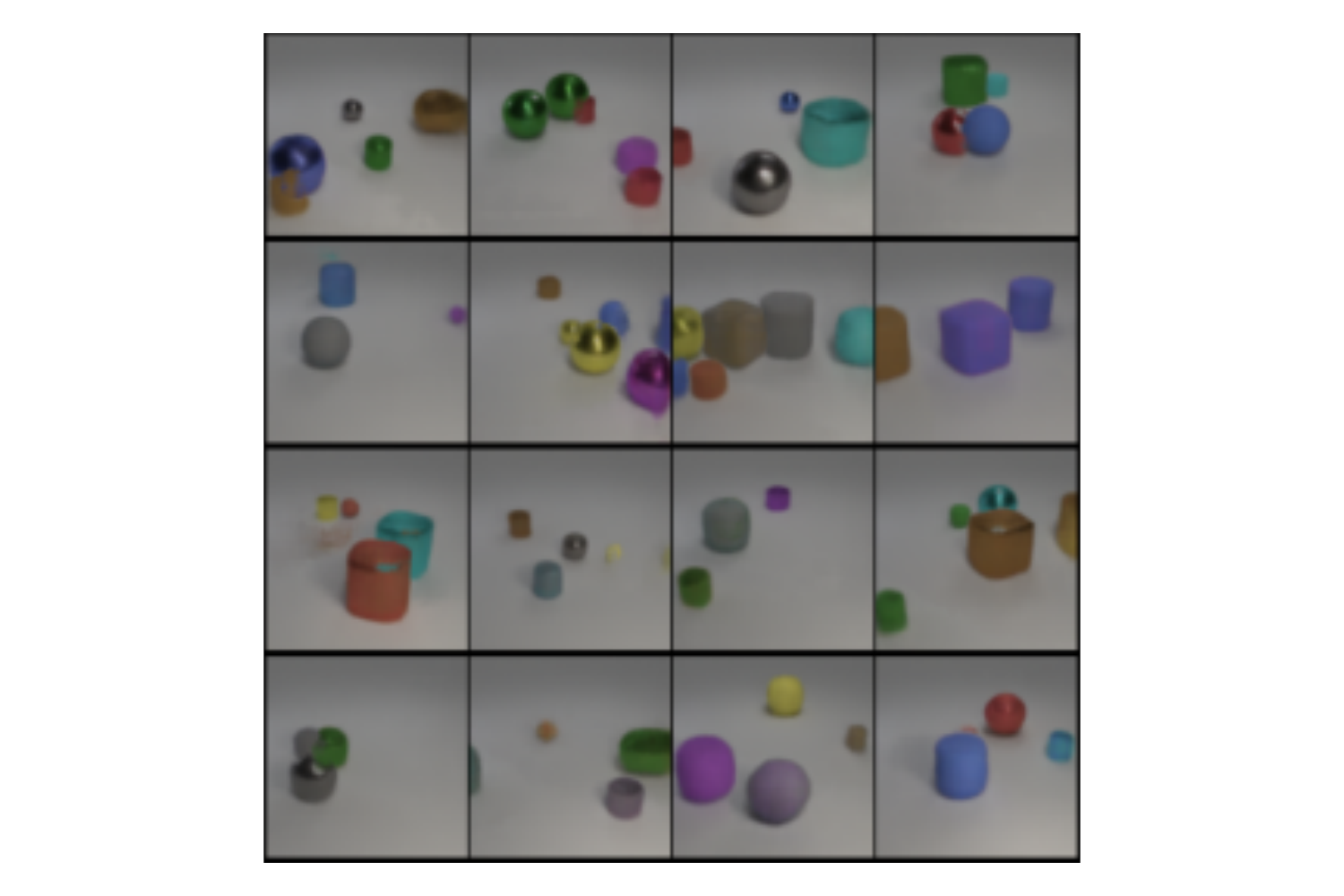}
        \caption{CLEVR6 images\label{fig:app:clevr6_4x4}}
    \end{subfigure}
    \begin{subfigure}{0.48\textwidth}
        \centering
        \includegraphics[trim=2.5cm 0 2.5cm 0,clip=true,scale=0.5]{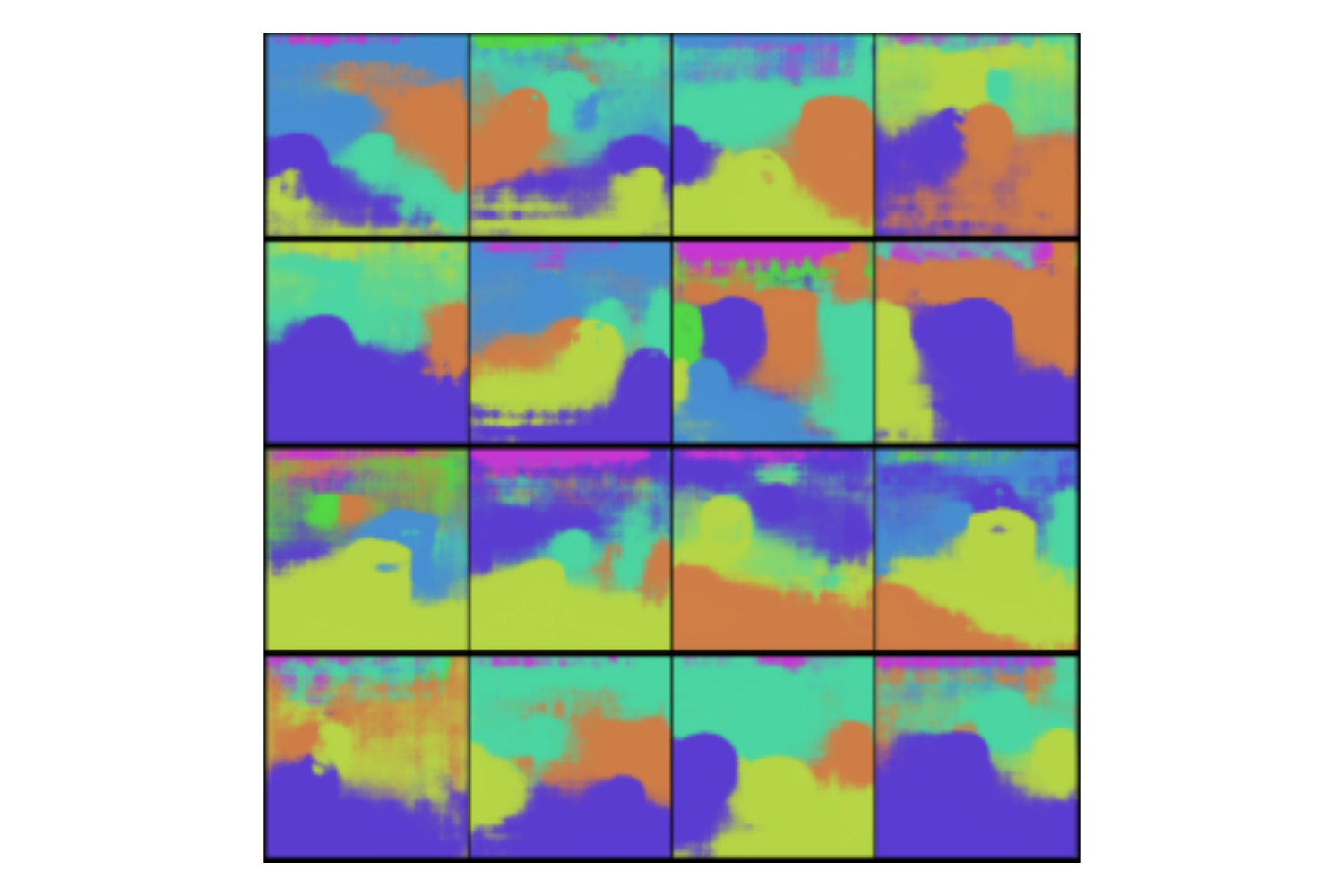}
        \caption{CLEVR6 masks\label{fig:app:clevr6_4x4_masks}}
    \end{subfigure}
    \qquad
    \begin{subfigure}{0.48\textwidth}
        \centering
        \includegraphics[trim=2.5cm 0 2.5cm 0,clip=true,scale=0.5]{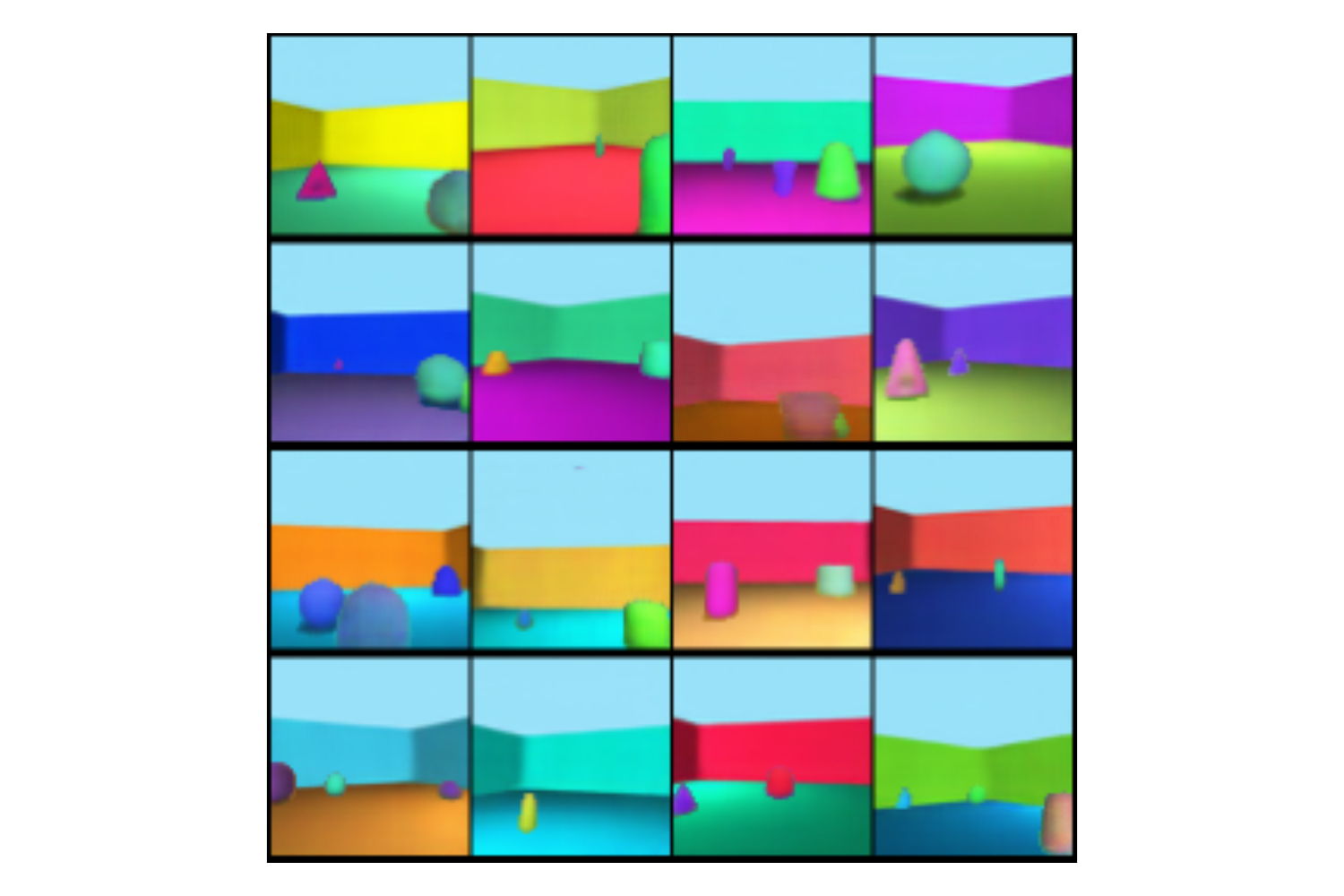}
        \caption{Objects Room images\label{fig:app:objectsroom_4x4}}
    \end{subfigure}
    \begin{subfigure}{0.48\textwidth}
        \centering
        \includegraphics[trim=2.5cm 0 2.5cm 0,clip=true,scale=0.5]{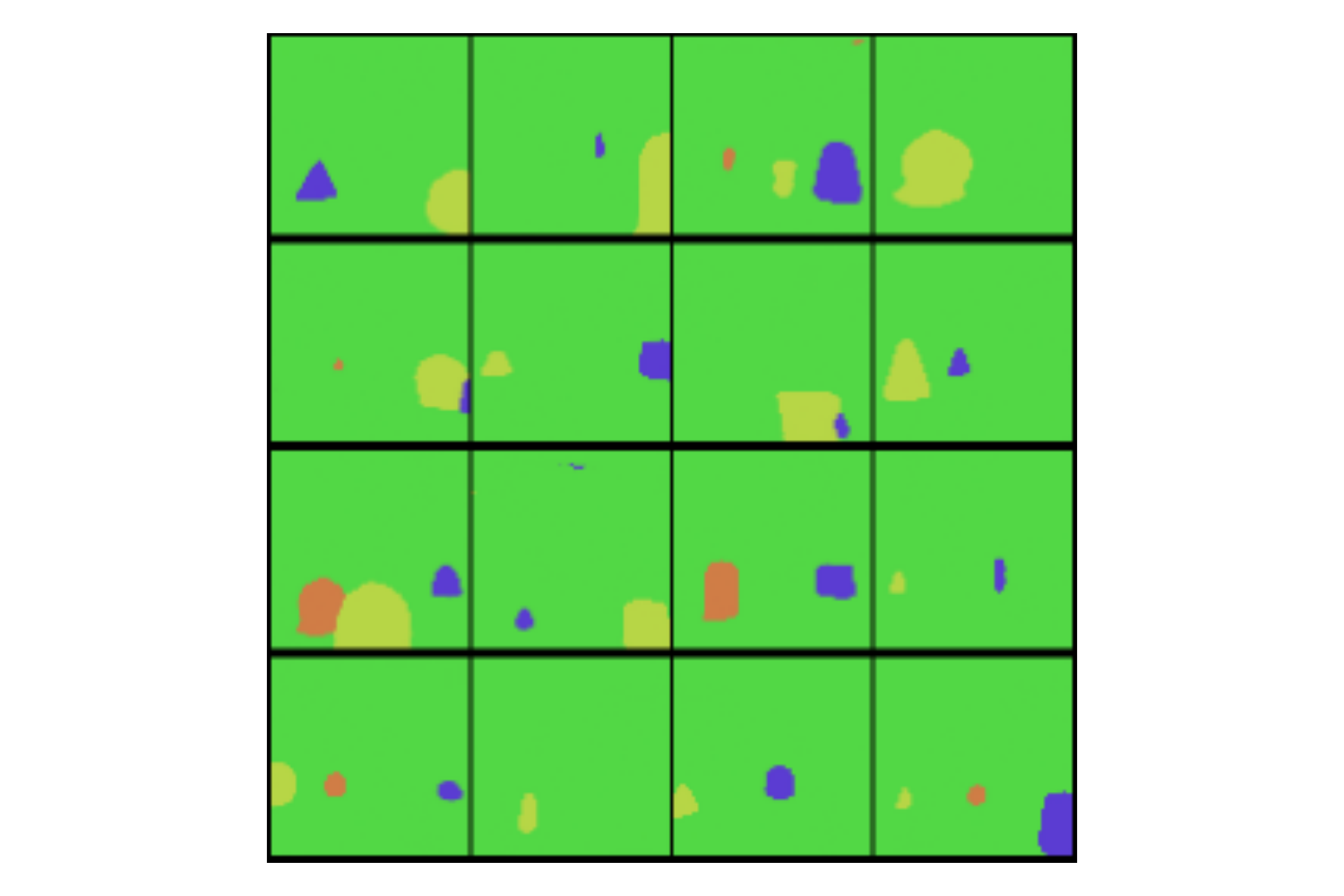}
        \caption{Objects Room masks\label{fig:app:objectsroom_4x4_masks}}
    \end{subfigure}
    \caption{Additional random samples generated by SRI-MoG.\label{fig:app:SRI_MoG_qual}}
\end{figure}

\begin{figure}[t]
    \centering
    \begin{subfigure}{0.48\textwidth}
        \centering
        \includegraphics[trim=2.5cm 0 2.5cm 0,clip=true,scale=0.5]{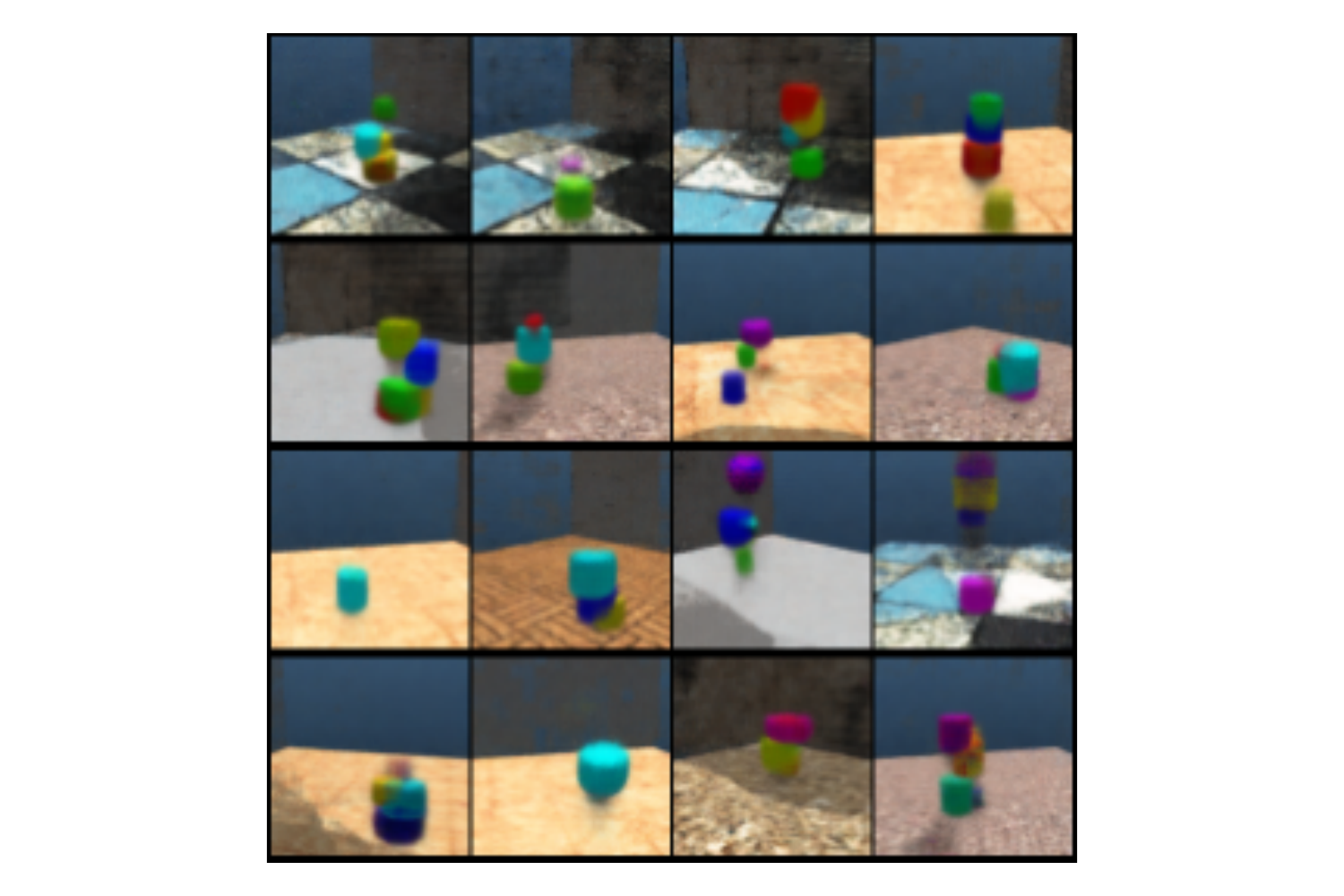}
        \caption{ShapeStacks images\label{fig:app:shapestacks_SRI_G_4x4}}
    \end{subfigure}
    \begin{subfigure}{0.48\textwidth}
        \centering
        \includegraphics[trim=2.5cm 0 2.5cm 0,clip=true,scale=0.5]{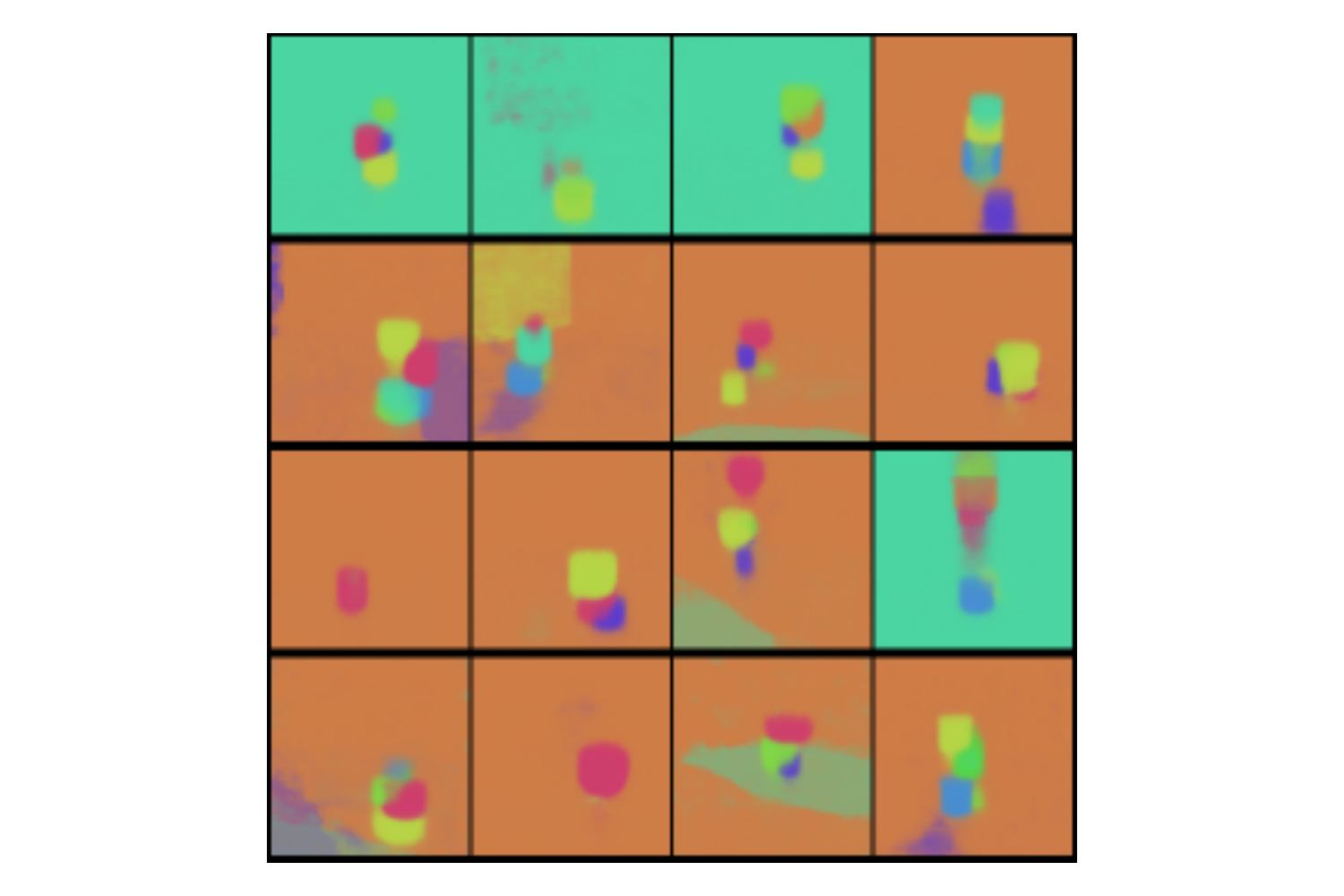}
        \caption{ShapeStacks masks\label{fig:app:shapestacks_SRI_G_4x4_masks}}
    \end{subfigure}
    \qquad
    \begin{subfigure}{0.48\textwidth}
        \centering
        \includegraphics[trim=2.5cm 0 2.5cm 0,clip=true,scale=0.5]{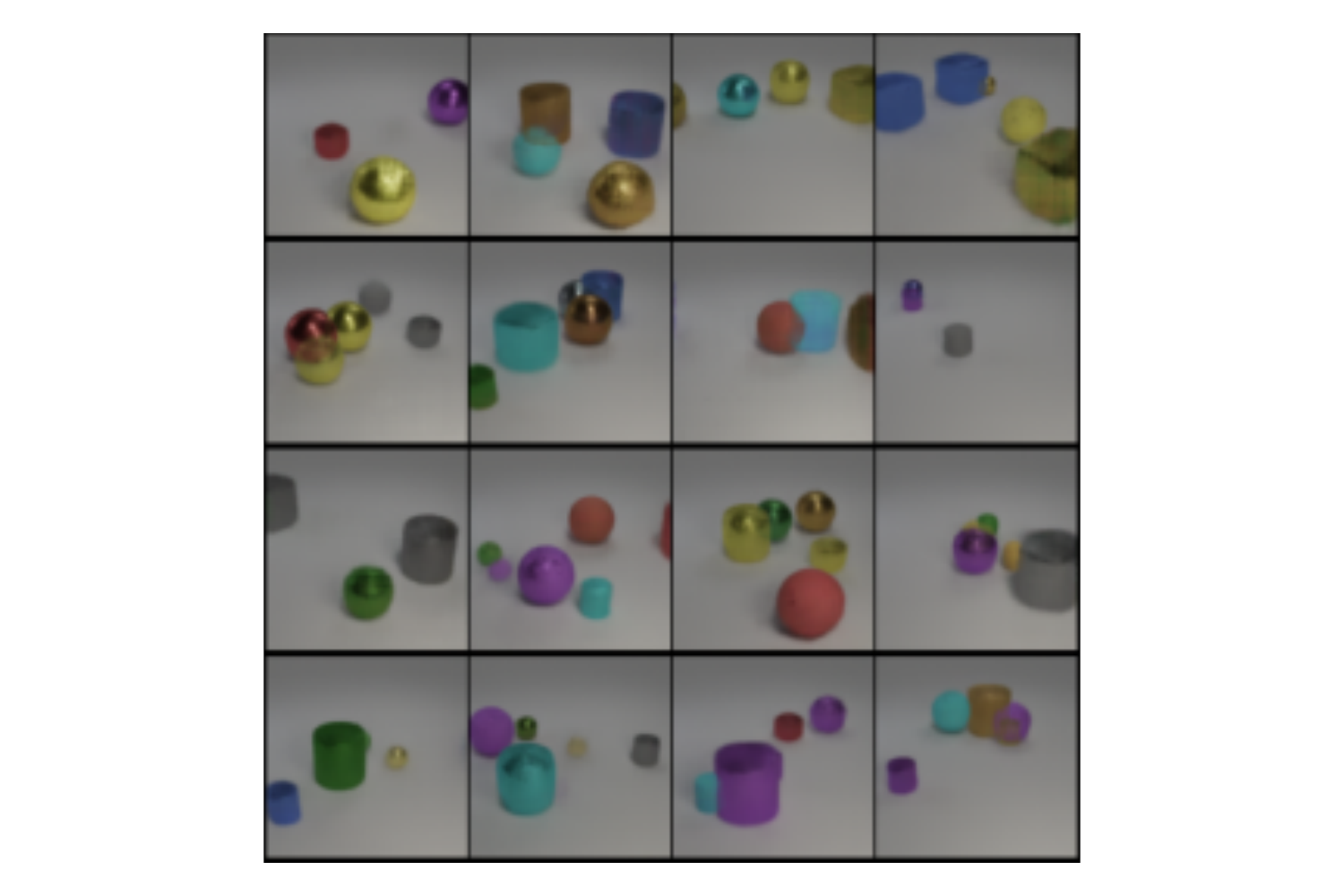}
        \caption{CLEVR6 images\label{fig:app:clevr6_SRI_G_4x4}}
    \end{subfigure}
    \begin{subfigure}{0.48\textwidth}
        \centering
        \includegraphics[trim=2.5cm 0 2.5cm 0,clip=true,scale=0.5]{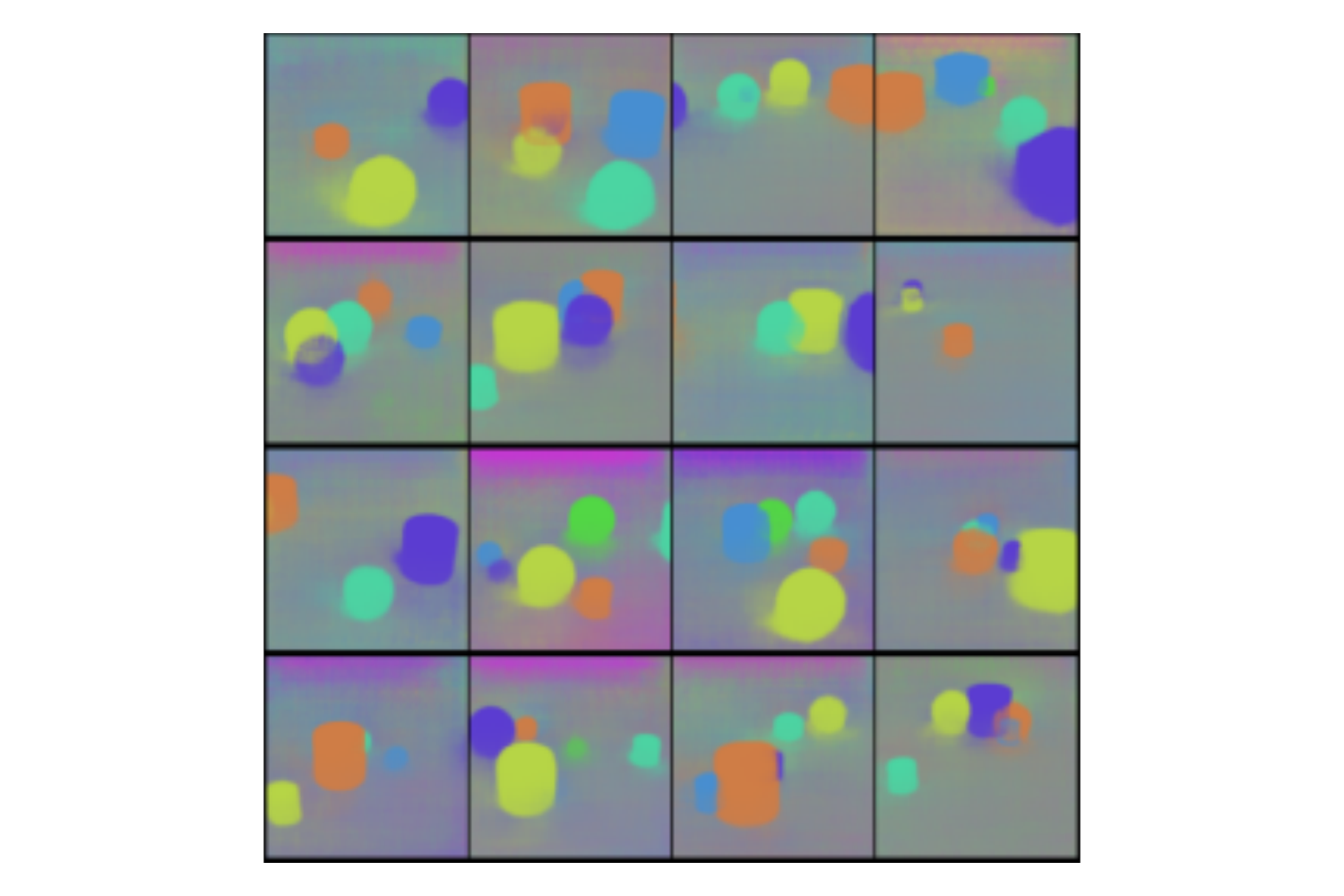}
        \caption{CLEVR6 masks\label{fig:app:clevr6_SRI_G_4x4_masks}}
    \end{subfigure}
    \qquad
    \begin{subfigure}{0.48\textwidth}
        \centering
        \includegraphics[trim=2.5cm 0 2.5cm 0,clip=true,scale=0.5]{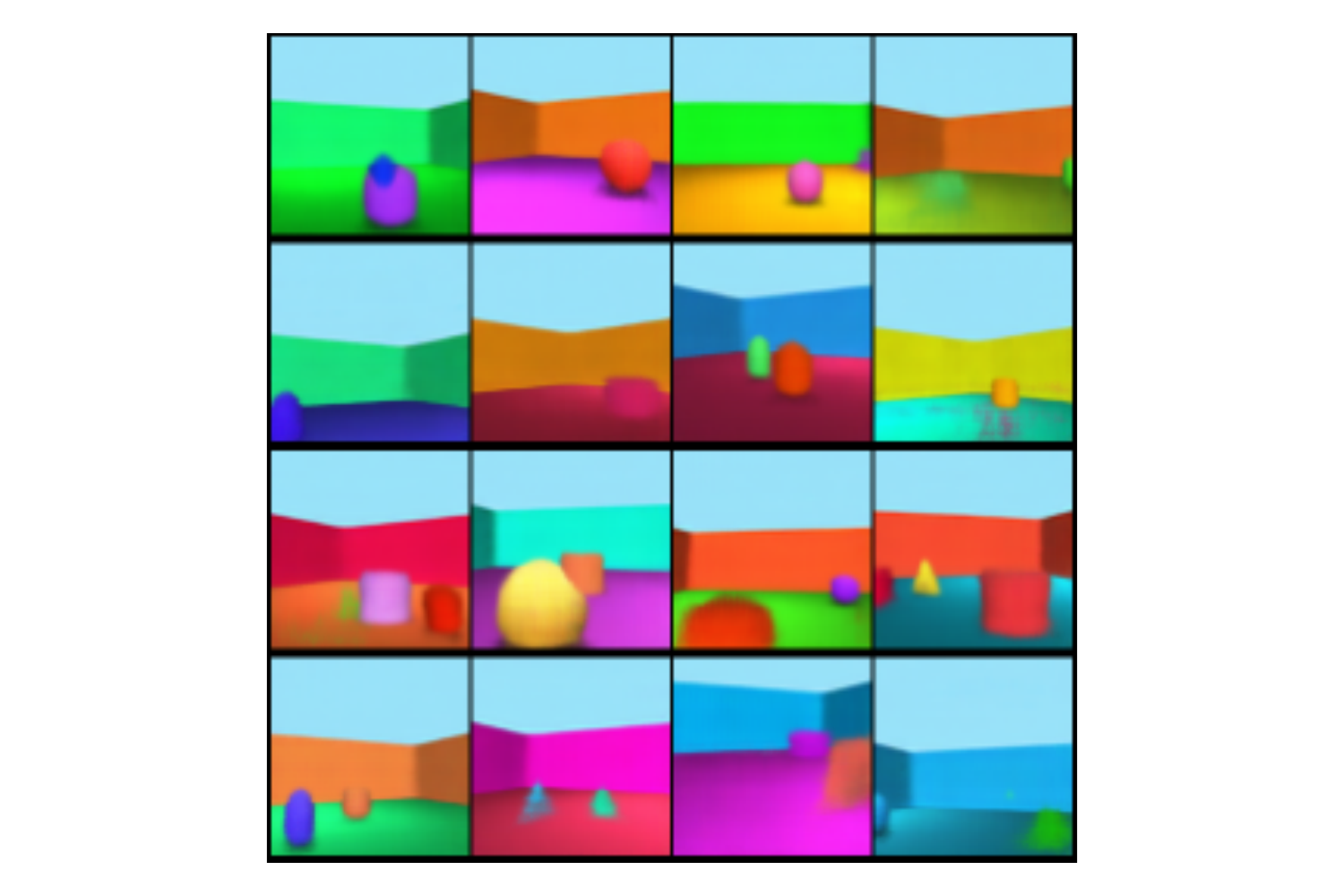}
        \caption{Objects Room images\label{fig:app:objectsroom_SRI_G_4x4}}
    \end{subfigure}
    \begin{subfigure}{0.48\textwidth}
        \centering
        \includegraphics[trim=2.5cm 0 2.5cm 0,clip=true,scale=0.5]{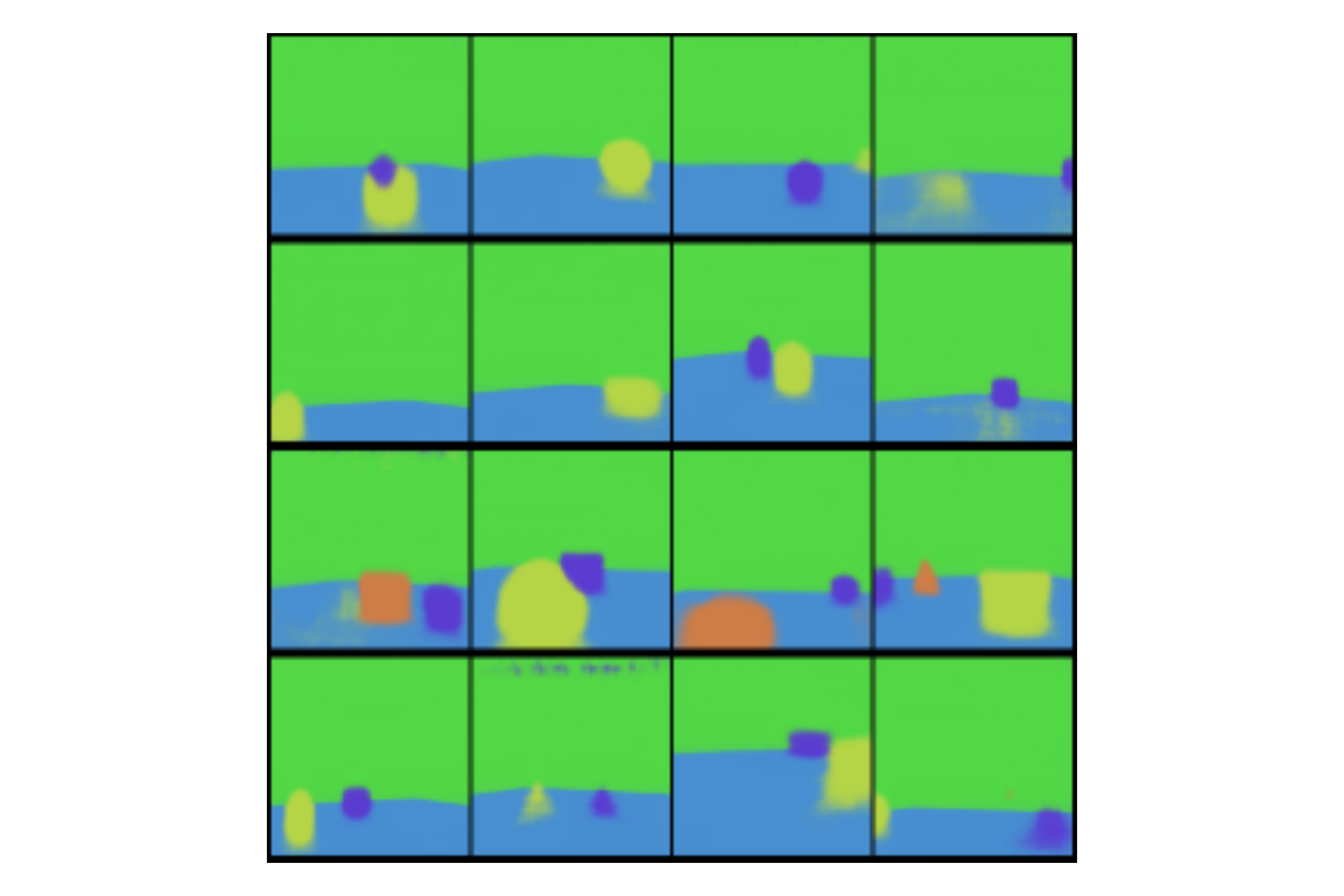}
        \caption{Objects Room masks\label{fig:app:objectsroom_SRI_G_4x4_masks}}
    \end{subfigure}
    \caption{Additional random samples generated by SRI-G.\label{fig:app:SRI_G_qual}}
\end{figure}

\begin{figure}[t]
    \centering
    \begin{subfigure}{0.48\textwidth}
        \centering
        \includegraphics[trim=2.5cm 0 2.5cm 0,clip=true,scale=0.5]{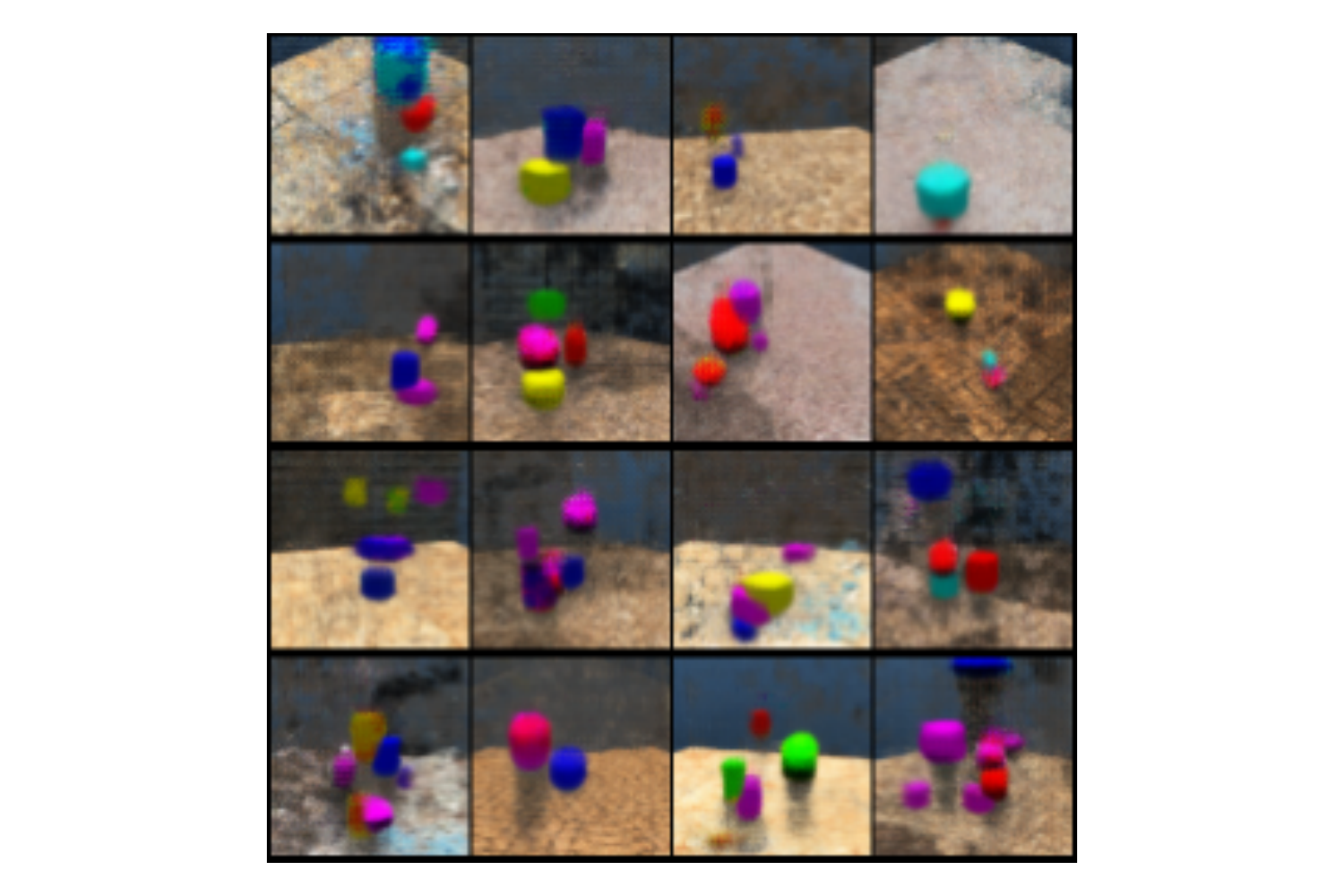}
        \caption{ShapeStacks images\label{fig:app:shapestacks_GENv2_4x4}}
    \end{subfigure}
    \begin{subfigure}{0.48\textwidth}
        \centering
        \includegraphics[trim=2.5cm 0 2.5cm 0,clip=true,scale=0.5]{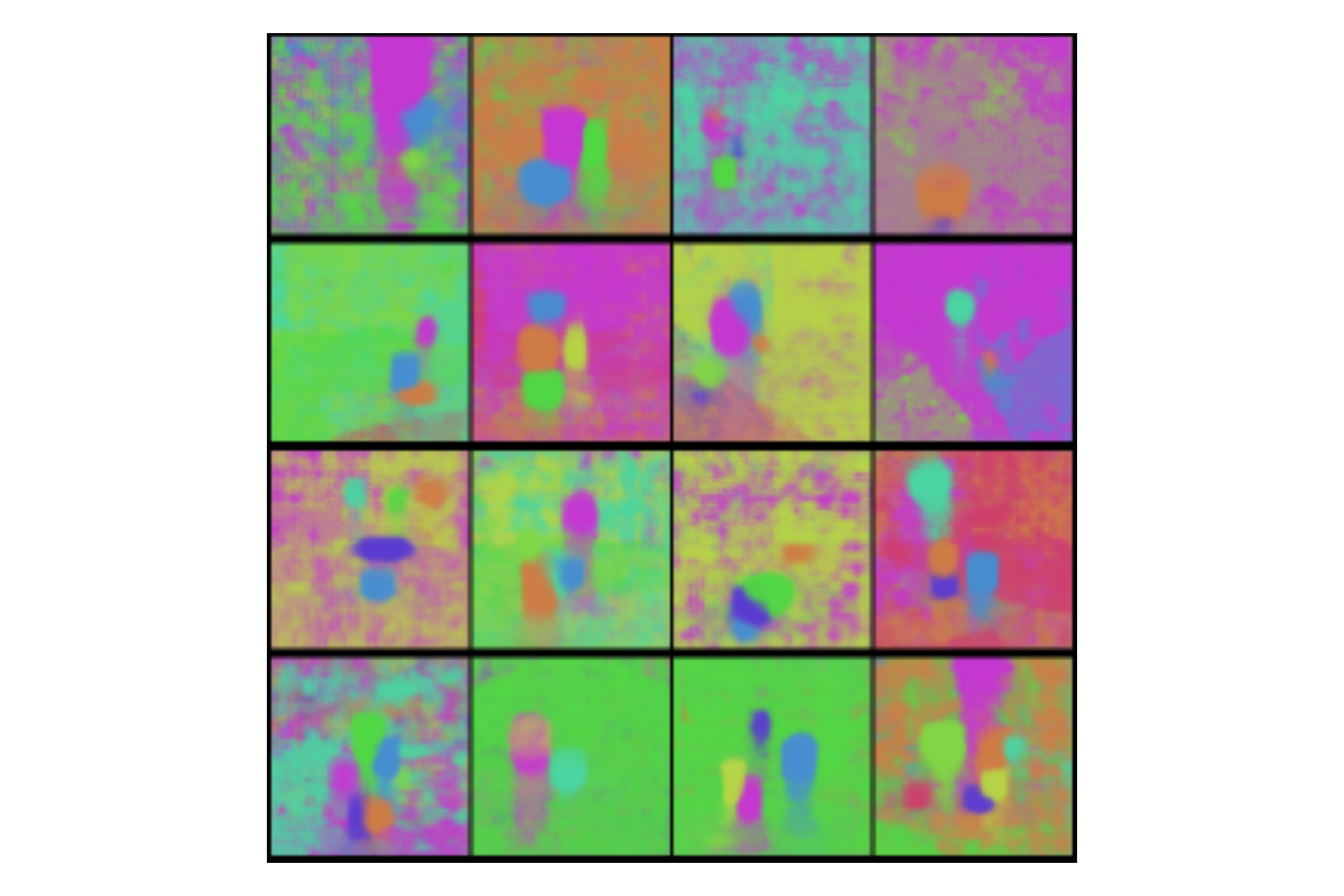}
        \caption{ShapeStacks masks\label{fig:app:shapestacks_GENv2_4x4_masks}}
    \end{subfigure}
    \qquad
    \begin{subfigure}{0.48\textwidth}
        \centering
        \includegraphics[trim=2.5cm 0 2.5cm 0,clip=true,scale=0.5]{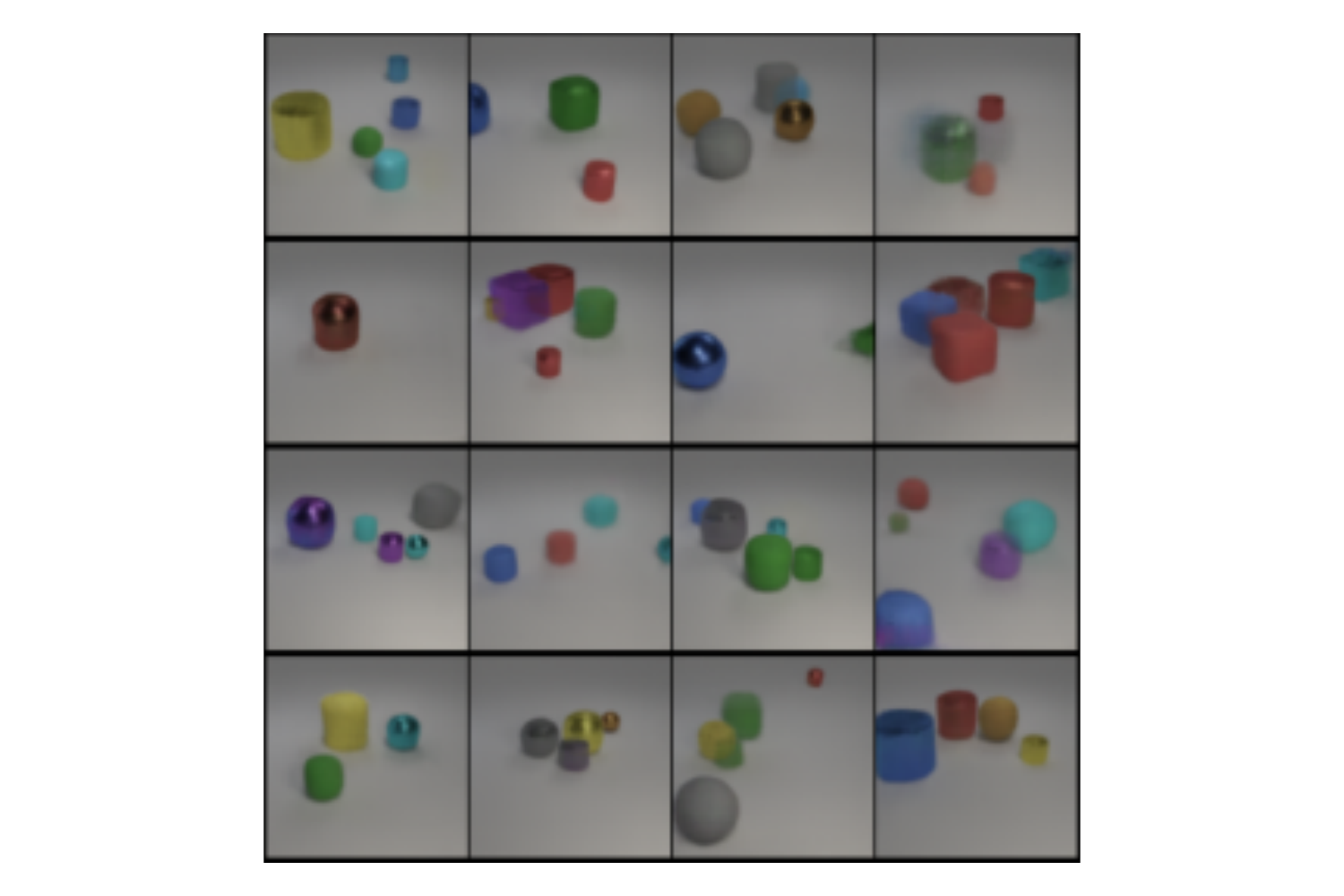}
        \caption{CLEVR6 images\label{fig:app:clevr6_GENv2_4x4}}
    \end{subfigure}
    \begin{subfigure}{0.48\textwidth}
        \centering
        \includegraphics[trim=2.5cm 0 2.5cm 0,clip=true,scale=0.5]{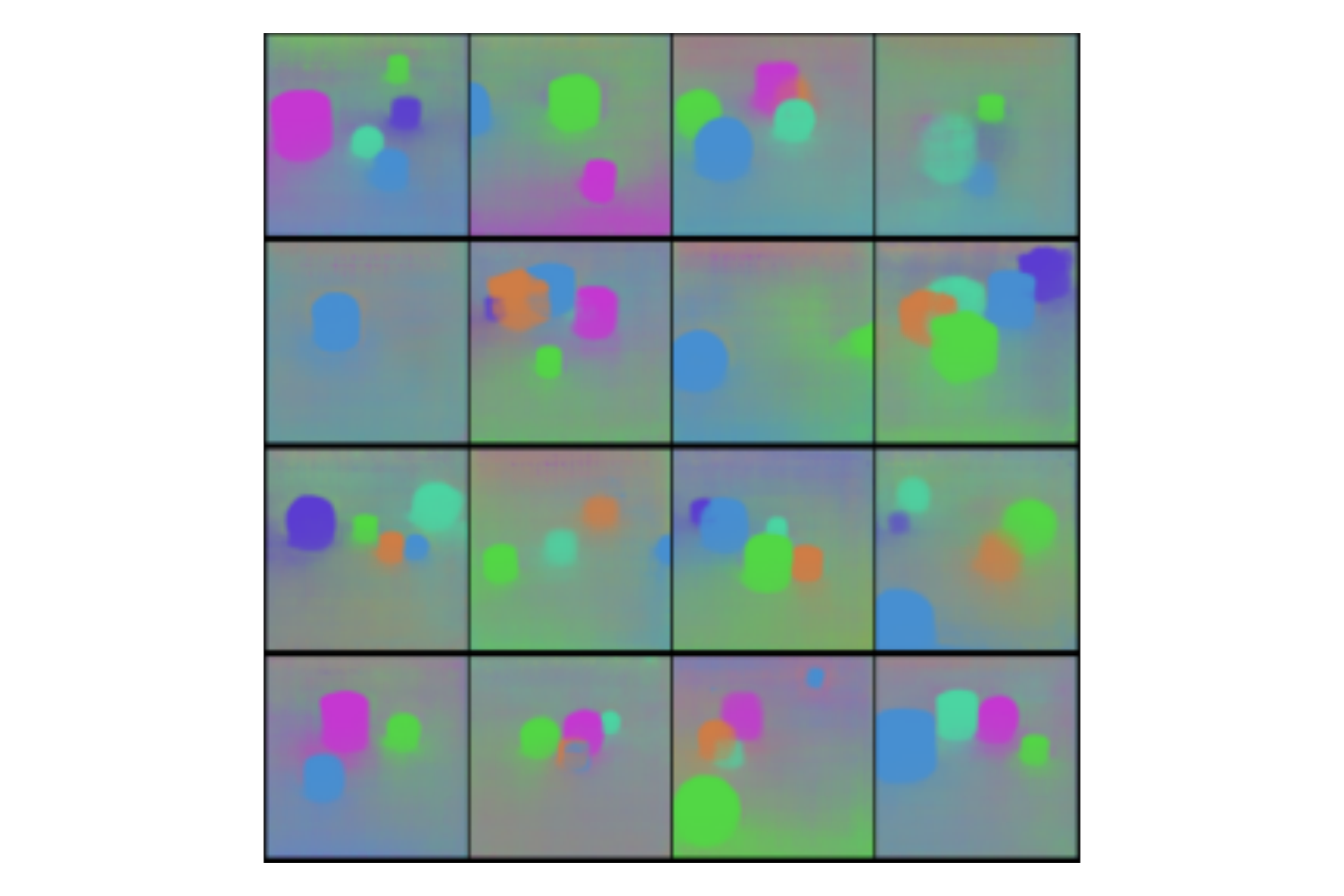}
        \caption{CLEVR6 masks\label{fig:app:clevr6_GENv2_4x4_masks}}
    \end{subfigure}
    \qquad
    \begin{subfigure}{0.48\textwidth}
        \centering
        \includegraphics[trim=2.5cm 0 2.5cm 0,clip=true,scale=0.5]{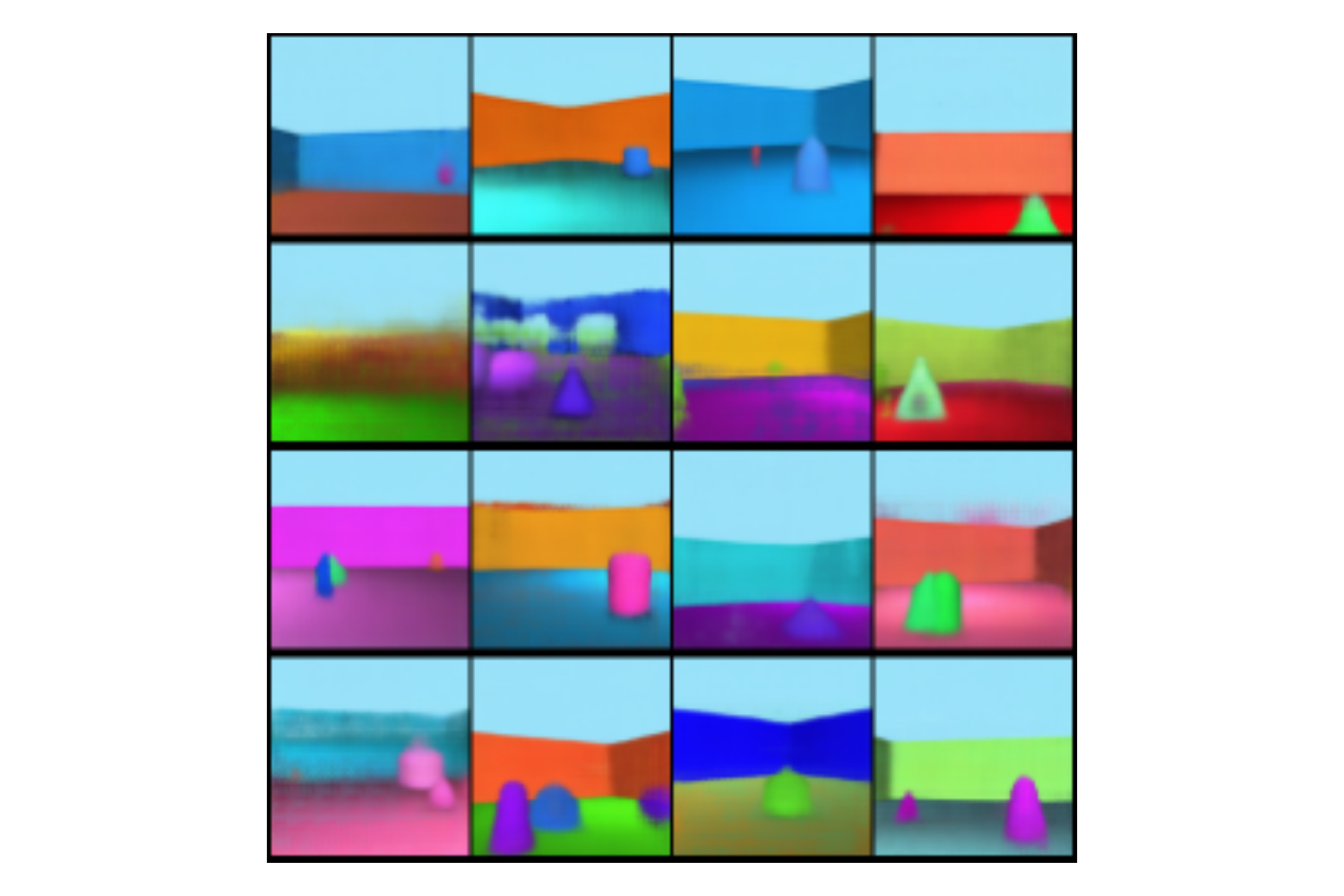}
        \caption{Objects Room images\label{fig:app:objectsroom_GENv2_4x4}}
    \end{subfigure}
    \begin{subfigure}{0.48\textwidth}
        \centering
        \includegraphics[trim=2.5cm 0 2.5cm 0,clip=true,scale=0.5]{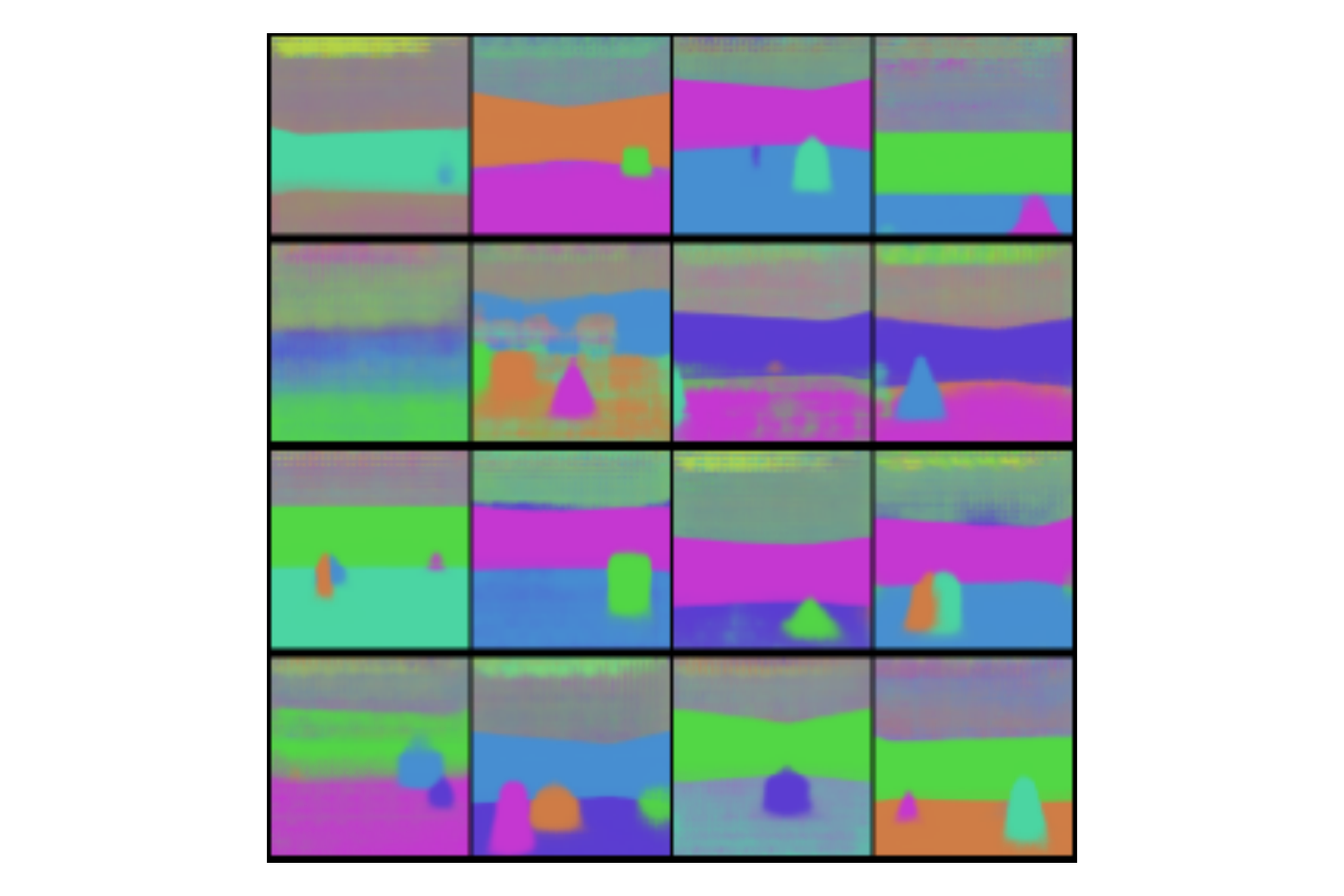}
        \caption{Objects Room masks\label{fig:app:objectsroom_GENv2_4x4_masks}}
    \end{subfigure}
    \caption{Additional random samples generated by GENESISv2-G.\label{fig:app:GENv2_qual}}
\end{figure}

\textbf{Additional random samples: } We visualize extra randomly sampled scenes from SRI-MoG (Figure~\ref{fig:app:SRI_MoG_qual}), SRI-G (Figure~\ref{fig:app:SRI_G_qual}), and GENv2-G (Figure~\ref{fig:app:GENv2_qual}).

\begin{figure}
    \centering
    \includegraphics[scale=0.75]{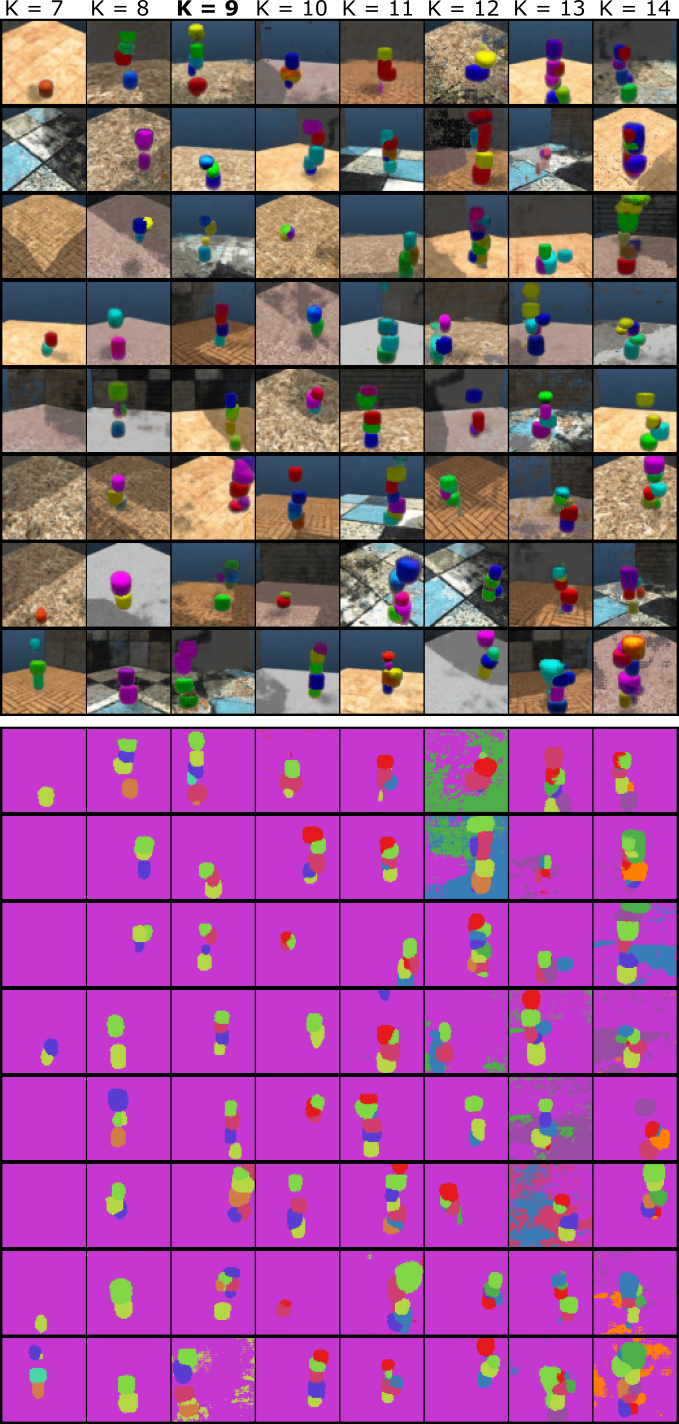}
    \caption{\textbf{Generalization to different numbers of slots $K$}. Random ShapeStacks samples.  SRI-MoG was trained with $K=9$.}
    \label{fig:app:generalization_K}
\end{figure}

\textbf{Generalizing to different numbers of slots: }We demonstrate that the number of slots can be changed at test time to generate scenes with fewer or more objects than seen during training in Figure~\ref{fig:app:generalization_K}.

\begin{figure}
    \centering
    \includegraphics[scale=0.8]{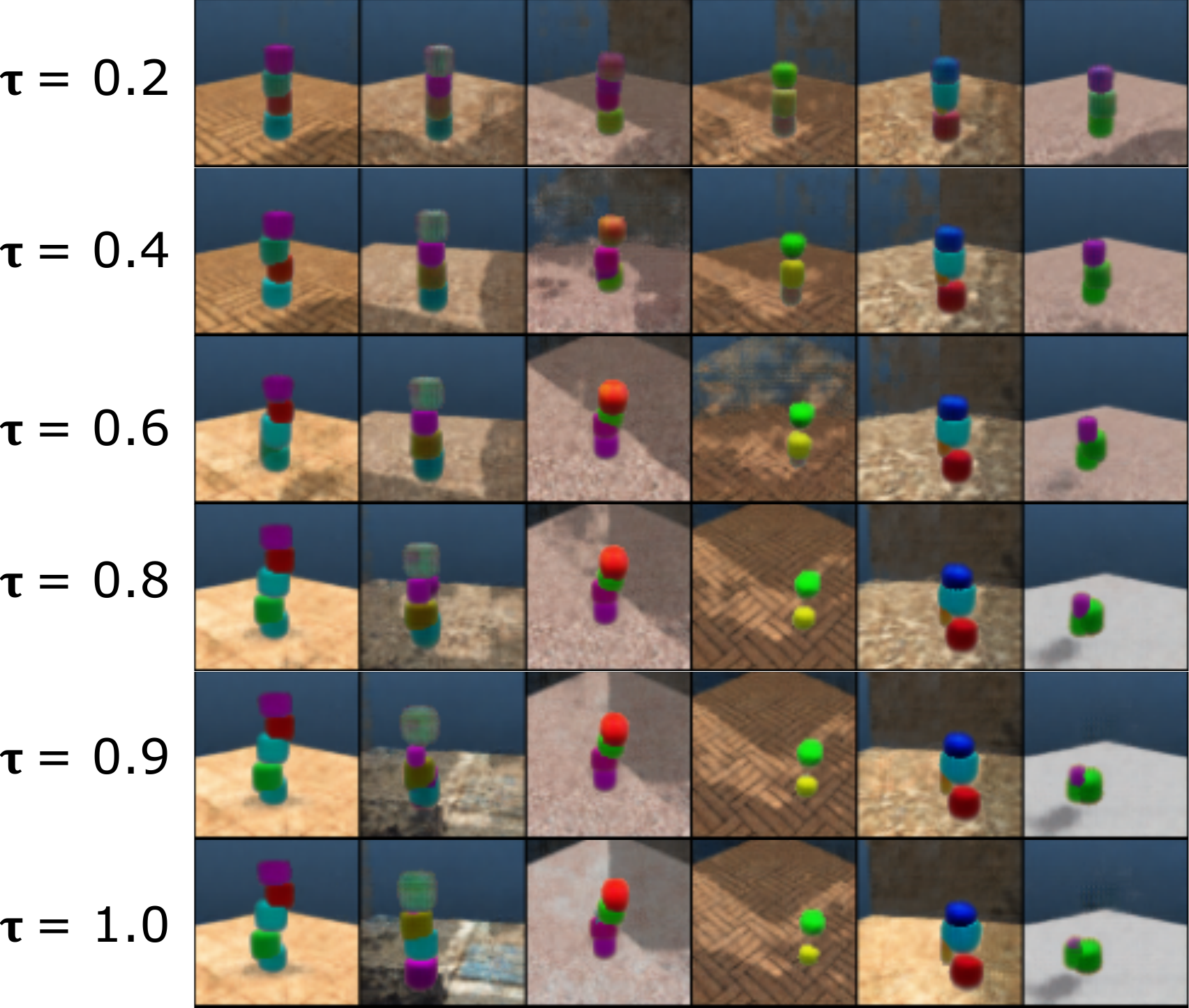}
    \caption{Randomly sampled ShapeStacks images from SRI-MoG with different temperature scaling parameters $\tau$ in the scene-level and slot prior. Scaling down the variances in a hierarchical Gaussian prior is known to help samples stay in regions of high probability~\citep{NEURIPS2020_e3b21256}. We confirm here that values of $\tau$ near 1.0 slightly improve sample quality on challenging datasets with minimal impact on sample diversity. Note that we use $\tau = 1.0$ in this work for all evaluations unless stated otherwise.}
    \label{fig:app:temperature}
\end{figure}

\textbf{Effect of temperature scaling: }We investigate whether temperature scaling provides any further qualitative improvement in sample quality (Figure~\ref{fig:app:temperature}).

\begin{figure}
    \centering
    \begin{subfigure}{0.48\textwidth}
        \includegraphics[trim=2.5cm 0 2.5cm 0,clip=true,scale=0.68]{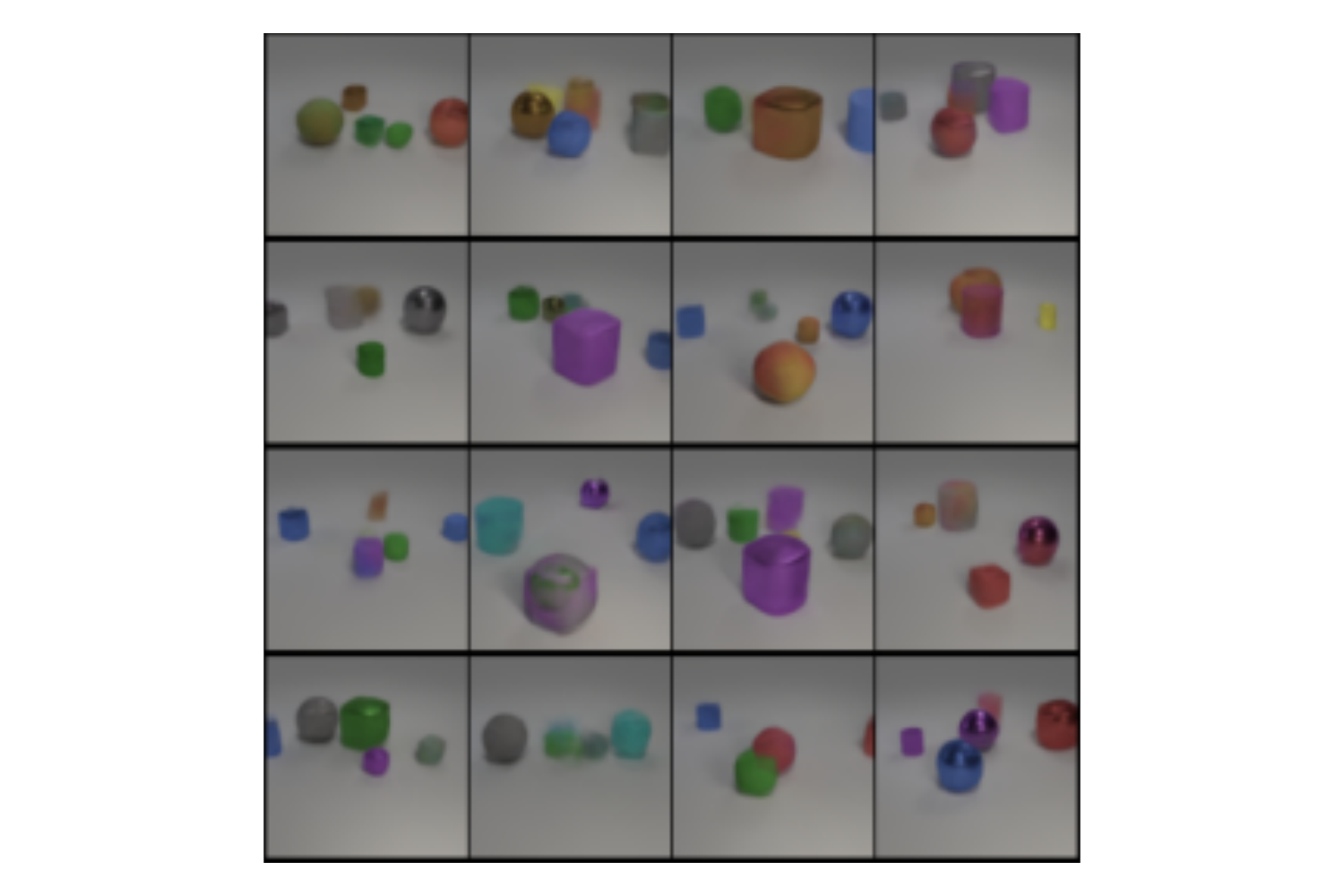}
        \caption{CLEVR6 images}
        \label{fig:app:clevr6_SRI_EMORL_4x4}
    \end{subfigure}
    \begin{subfigure}{0.48\textwidth}
        \includegraphics[trim=2.5cm 0 2.5cm 0,clip=true,scale=0.68]{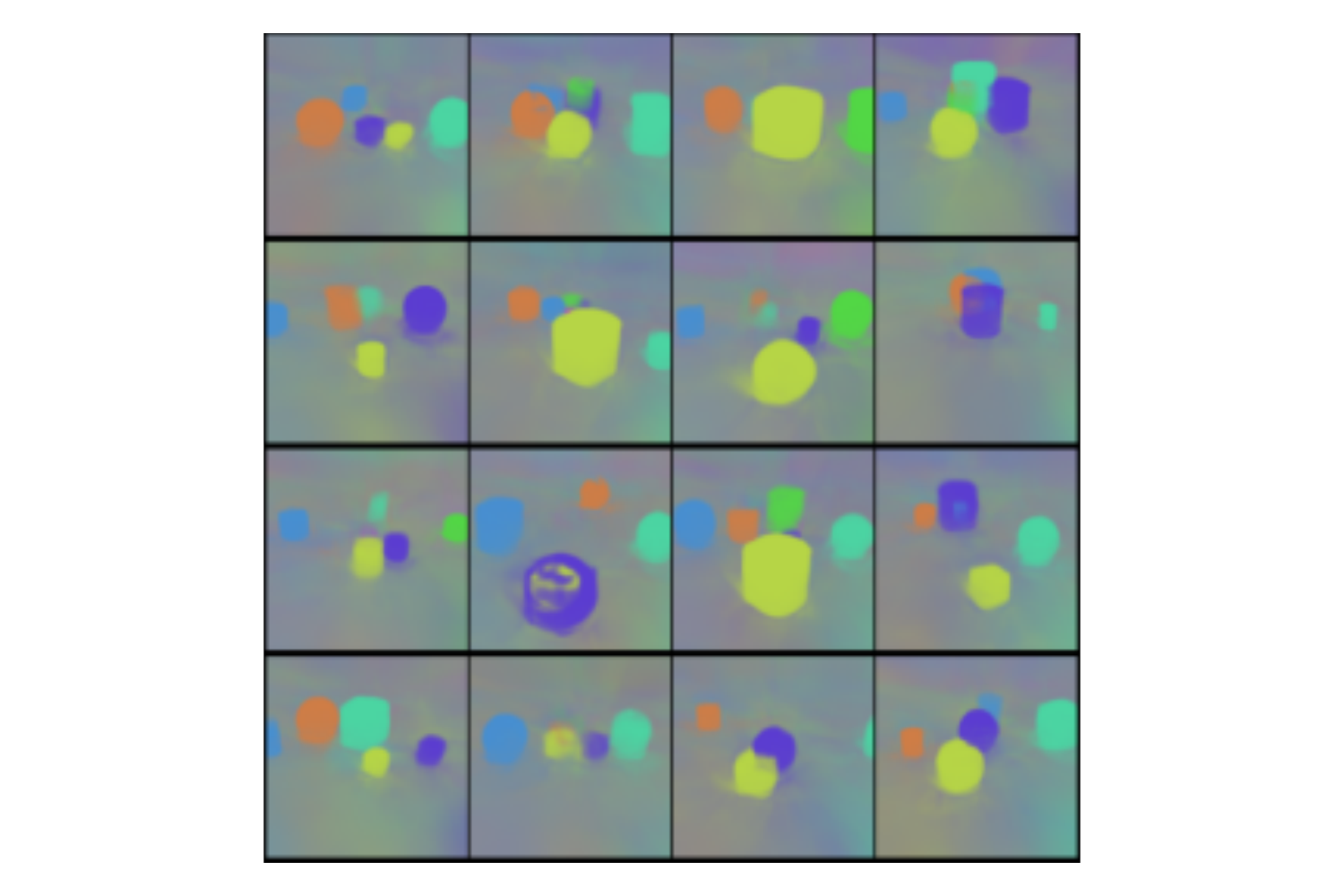}
        \caption{CLEVR6 masks}
        \label{fig:app:clevr6_SRI_EMORL_4x4_masks}
    \end{subfigure}
    \caption{Random samples generated by SRI-EMORL. These images are generated with temperature $\tau = 0.8$. \label{fig:app:SRI_EMORL}}
\end{figure}

\textbf{Random samples from SRI-EMORL: }Random samples from SRI-EMORL on CLEVR6 (visualized with temperature scaling) (Figure~\ref{fig:app:SRI_EMORL}).

\begin{figure}
    \centering
    \includegraphics[scale=0.75]{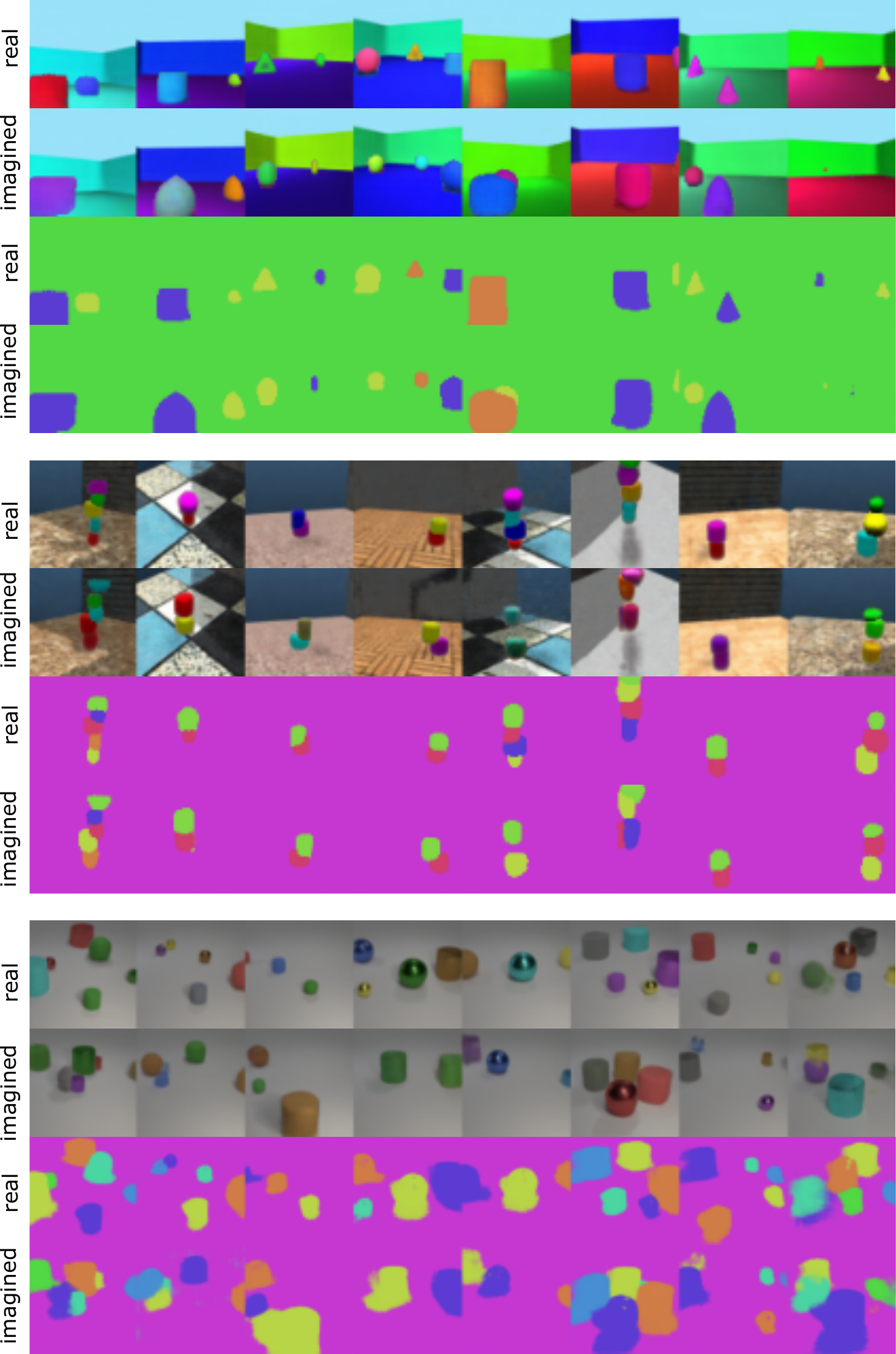}
    \caption{\textbf{Imagination examples.} Side-by-side comparisons of reconstructed RGB images and masks from SRI-MoG's aligned autoregressive posterior with imagined RGB images and masks.}
    \label{fig:app:imagination_real}
\end{figure}

\textbf{Imagination rollouts: }We provide more examples of pairs of reconstructed scenes with the corresponding imagination rollout used by SRI for slot order estimation (Figure~\ref{fig:app:imagination_real}).

\begin{figure}
    \centering
    \includegraphics[scale=0.7]{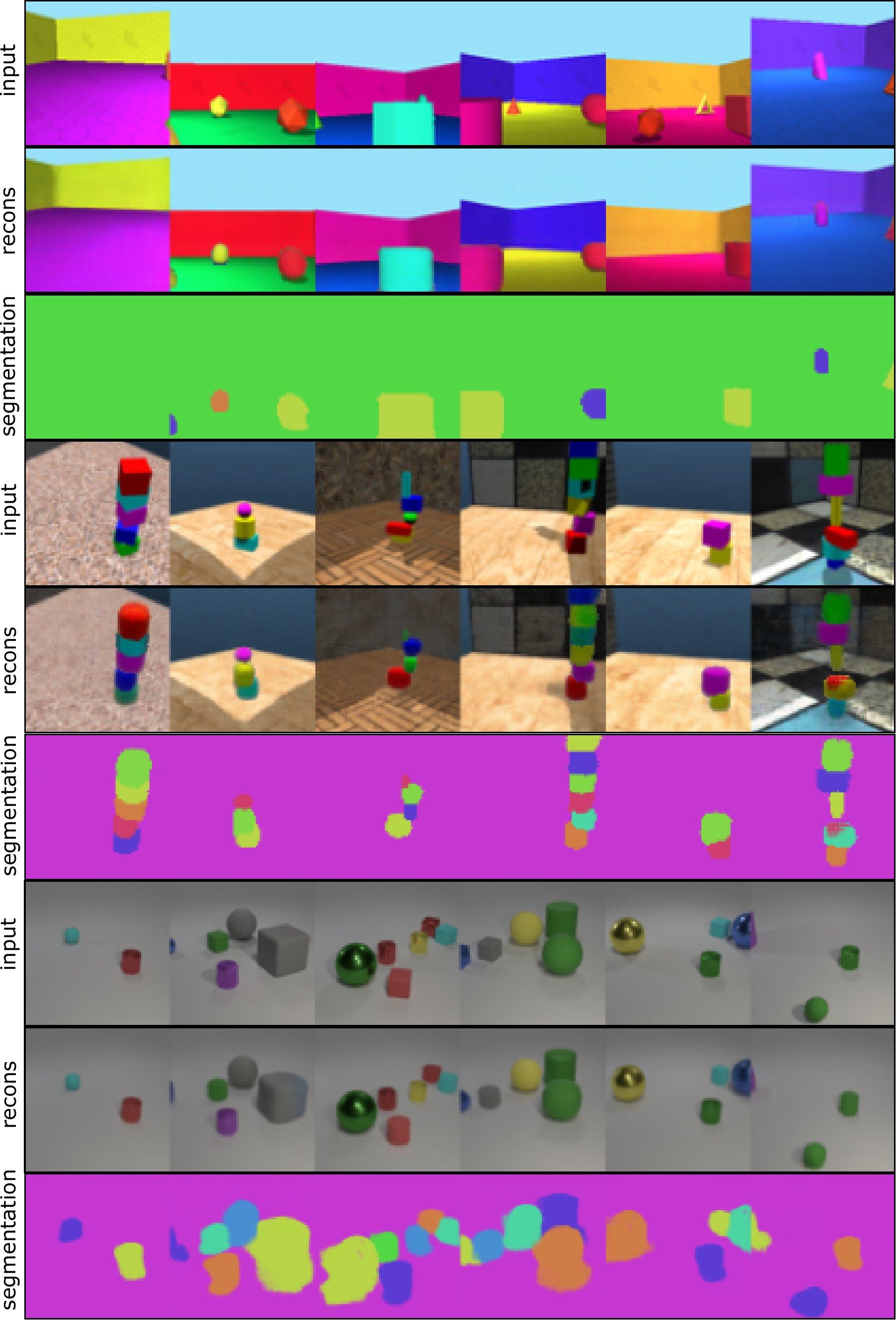}
    \caption{SRI-MoG reconstruction and segmentation.}
    \label{fig:app:sri_mog_recons}
\end{figure}

\begin{figure}
    \centering
    \includegraphics[scale=0.7]{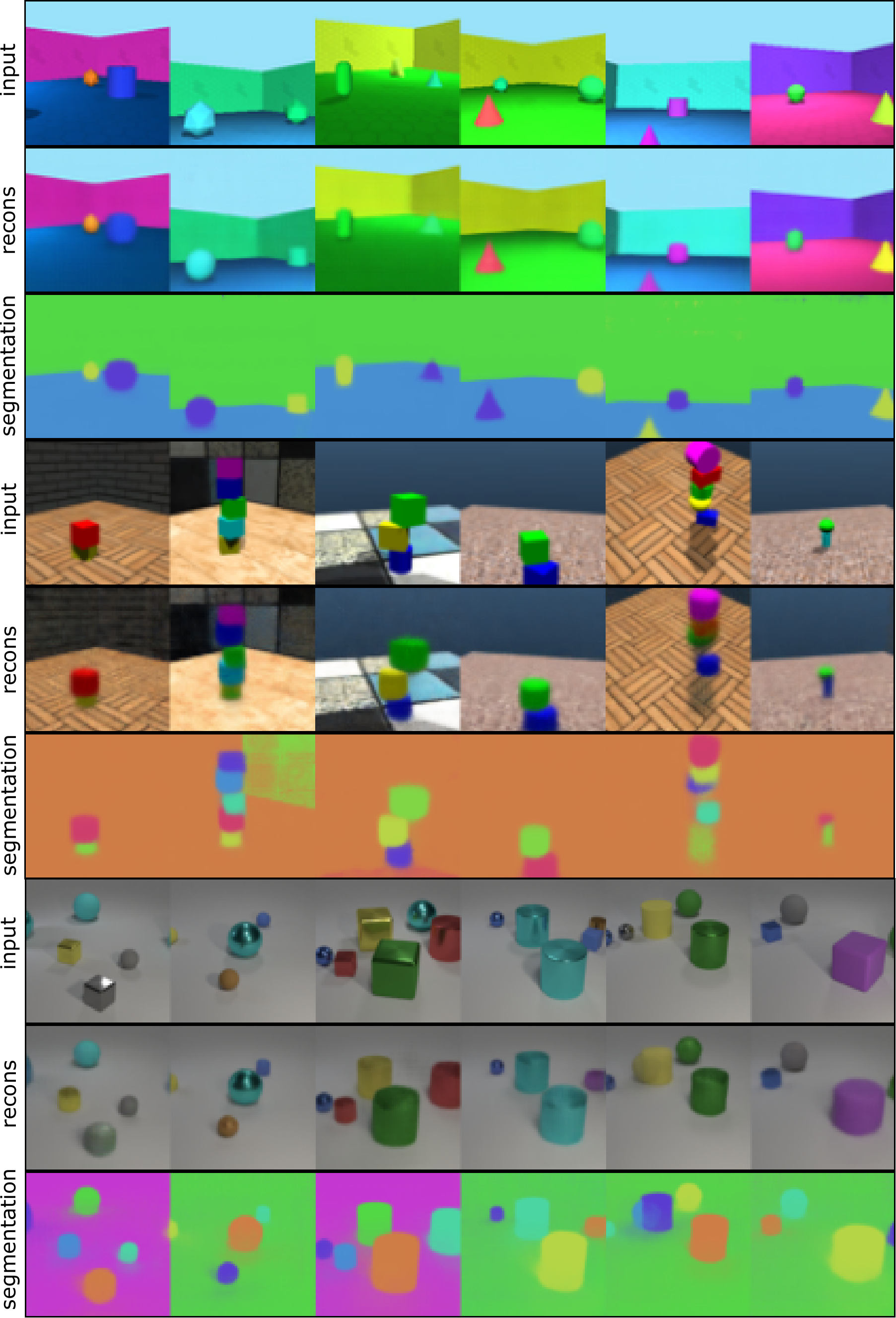}
    \caption{SRI-G reconstruction and segmentation.}
    \label{fig:app:sri_recons}
\end{figure}

\begin{figure}
    \centering
    \begin{subfigure}[t]{\textwidth}
        \centering
        \includegraphics[scale=0.5]{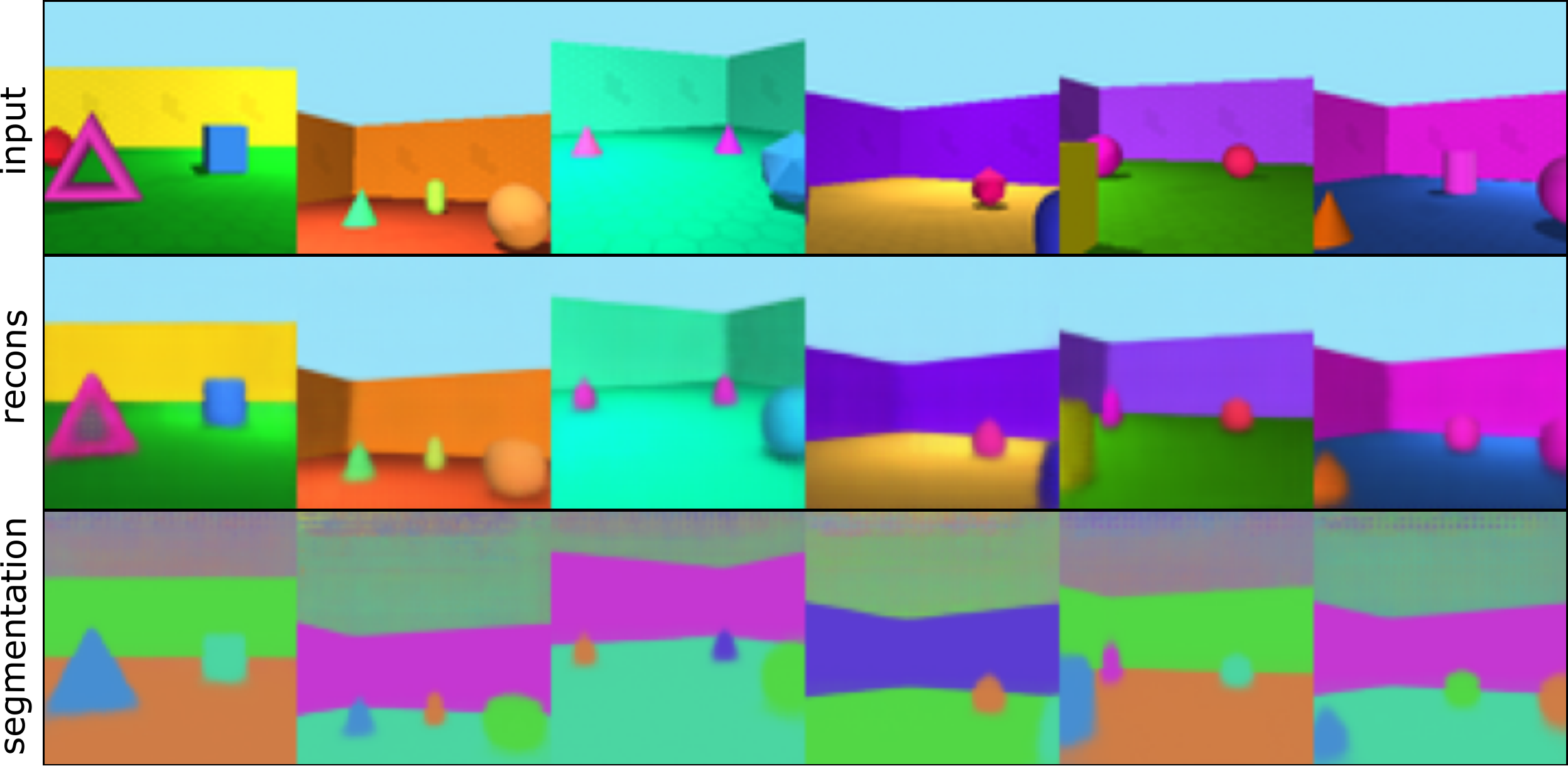}
    \end{subfigure}
    \begin{subfigure}[t]{\textwidth}
        \centering
        \includegraphics[scale=0.5]{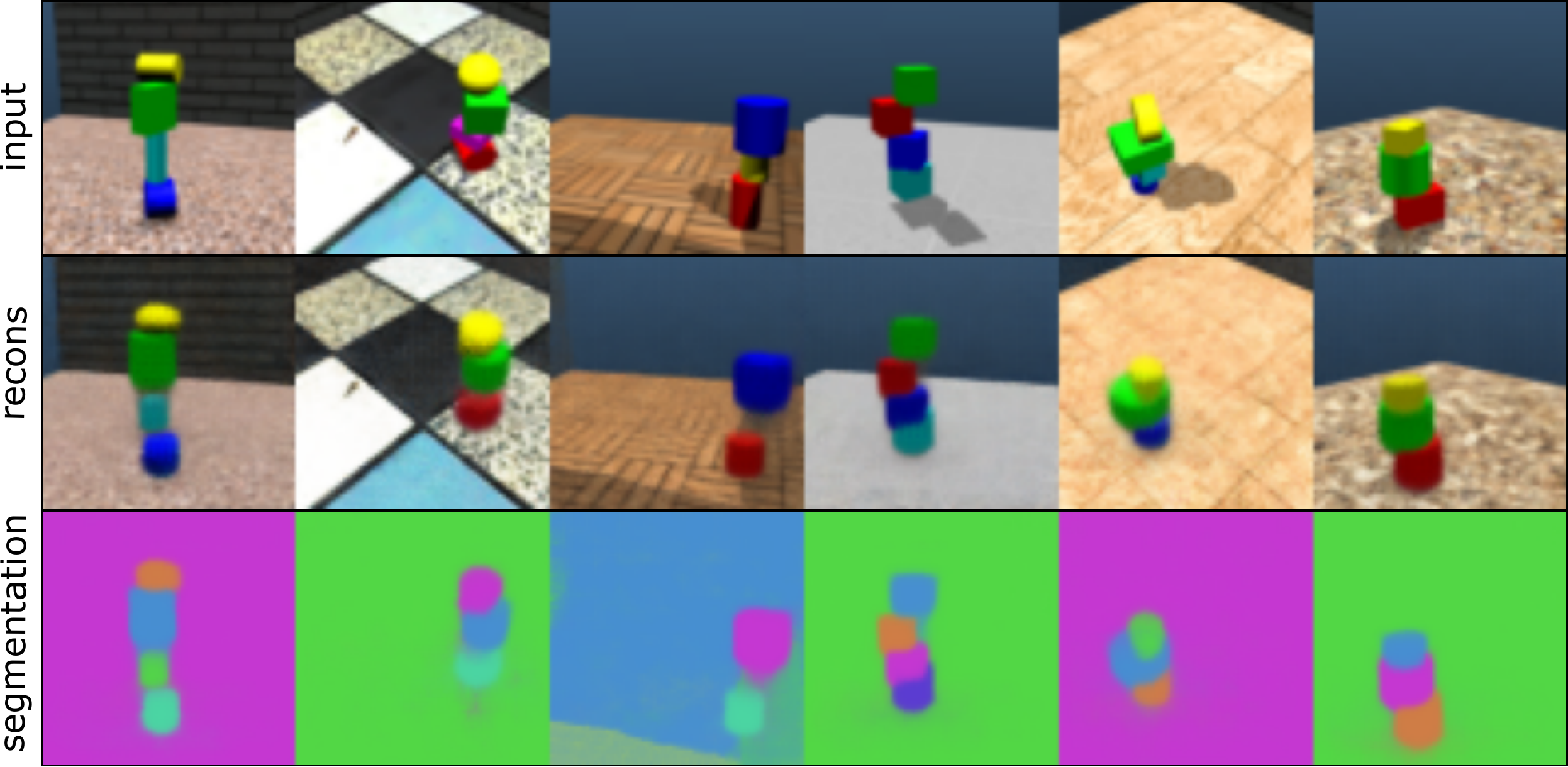}
    \end{subfigure}
    \begin{subfigure}[t]{\textwidth}
        \centering
        \includegraphics[scale=0.5]{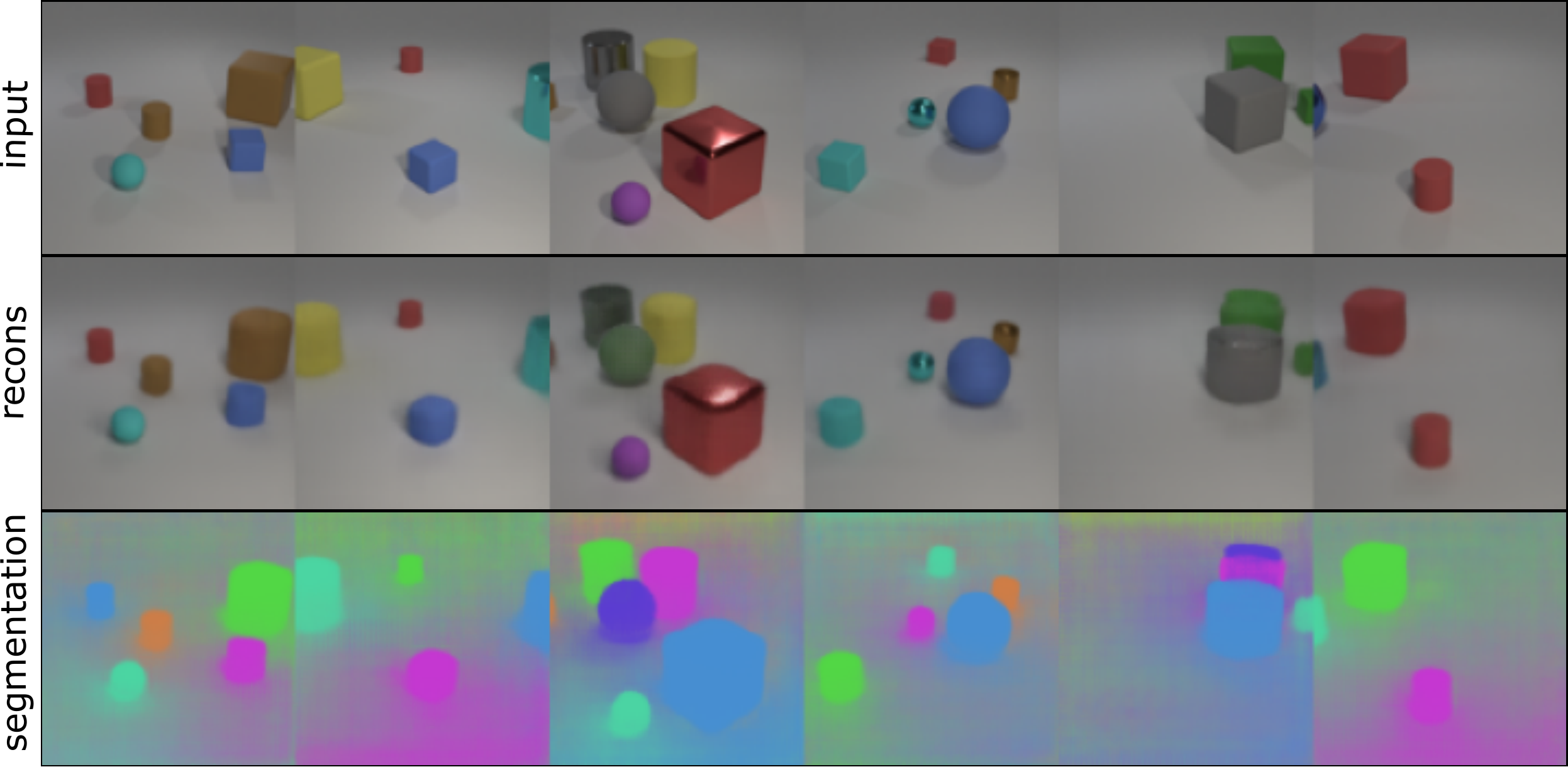}
    \end{subfigure}
    \caption{GENv2-G reconstruction and segmentation.\label{fig:app:GenV2G_recons}}
\end{figure}

\textbf{Reconstruction and segmentation examples: }Examples of reconstructed and segmented scenes from each dataset for SRI-MoG (Figure~\ref{fig:app:sri_mog_recons}), SRI-G (Figure~\ref{fig:app:sri_recons}), and GENv2-G (Figure~\ref{fig:app:GenV2G_recons}).

\end{document}

%% file: NEURCOMP-D-23-00020R1-macros.tex
\usepackage{amsmath,amsfonts,bm,amssymb}
\usepackage[utf8]{inputenc} 
\usepackage[T1]{fontenc}    
\usepackage{booktabs}       
\usepackage{nicefrac}       
\usepackage{microtype}      
\usepackage[table]{xcolor}         
\usepackage{mathtools}
\usepackage{multicol}
\usepackage{multirow}
\usepackage{algorithm}
\usepackage{algorithmic}
\usepackage{pifont}
 \usepackage{wrapfig}
\usepackage{xspace}
\usepackage{xpatch}
\usepackage{xstring}
\usepackage{xparse}
\usepackage{subcaption}
\usepackage{lipsum}
\usepackage{hyperref}
\usepackage{url}
\usepackage[export]{adjustbox}
\usepackage{sidecap}

\sidecaptionvpos{figure}{b}

\definecolor{mydarkblue}{rgb}{0,0.08,0.45}
\definecolor{myblue}{RGB}{103,169,207}
\definecolor{myred}{RGB}{239,138,98}

\hypersetup{%
colorlinks=true,
linkcolor=mydarkblue,
citecolor=mydarkblue,
filecolor=mydarkblue,
urlcolor=mydarkblue}

\DeclareRobustCommand{\note}[1]{}

\DeclarePairedDelimiterX{\SquareBrackets}[1]{[}{]}{#1}
\DeclarePairedDelimiterX{\RoundBrackets}[1]{(}{)}{#1}
\DeclarePairedDelimiterX{\DivergenceBrackets}[2]{[}{]}{#1\;\delimsize\|\;#2}

\NewDocumentCommand{\pr}{ O{p} r() }{
  \def\prArg{#2}\patchcmd{\prArg}{|}{\mid}{}{}#1\RoundBrackets{\prArg}}
\NewDocumentCommand{\p}{ r() }{\pr[p](#1)}
\NewDocumentCommand{\q}{ r() }{\pr[q](#1)}
\NewDocumentCommand{\prm}{ r() }{\pr[\mathrm{p}](#1)}
\NewDocumentCommand{\Normal}{ r() }{\pr[\operatorname{Normal}](#1)}
\NewDocumentCommand{\Cat}{ r() }{\pr[\operatorname{Cat}](#1)}
\NewDocumentCommand{\Beta}{ r() }{\pr[\operatorname{Beta}](#1)}
\NewDocumentCommand{\Bernoulli}{ r() }{\pr[\operatorname{Bernoulli}](#1)}
\NewDocumentCommand{\Dir}{ r() }{\pr[\operatorname{Dir}](#1)}
\NewDocumentCommand{\Poisson}{ r() }{\pr[\operatorname{Poisson}](#1)}

\newlength\widthE

\DeclarePairedDelimiterX{\infdivx}[2]{ \big( }{ \big) }{%
  #1\;\delimsize\|\;#2%
}
\newcommand{\infdiv}{D_{KL}\infdivx}

\newcommand{\revision}[1]{\textcolor{black}{#1}}

\newlength\myindent
\def\eqref#1{equation~\ref{#1}}

\def\1{\bm{1}}

\def\rvs{{\mathbf{s}}}

\def\rvx{{\mathbf{x}}}

\def\rvz{{\mathbf{z}}}

\def\vs{{\bm{s}}}

\def\mC{{\bm{C}}}

\def\mW{{\bm{W}}}

\DeclareMathAlphabet{\mathsfit}{\encodingdefault}{\sfdefault}{m}{sl}
\SetMathAlphabet{\mathsfit}{bold}{\encodingdefault}{\sfdefault}{bx}{n}

\DeclareMathOperator*{\argmin}{arg\,min}

%% file: tables/NEURCOMP-D-23-00020R1-fixedorder.tex
{
\begin{table}[t]
\centering
\caption{\revision{\textbf{SRI-SA* Results.} Mean $\pm$ std. dev. over 3 Shapestacks runs. All results for SRI-SA* are shown due to the high variance across random seeds.}}
\label{tab:fixedorder}
\begin{adjustbox}{max width=\columnwidth}
\begin{tabular}{@{}llll@{}}
\toprule
 Model &
  FID\tiny{$\downarrow$} & ARI-FG\tiny{$\uparrow$} &
  -ELBO (bpd)\tiny{$\downarrow$} \\
  \midrule 
  SRI-SA* & $146.3_{\tiny{\pm 83}}^{[79/97/263]}$ & $0.412_{\tiny{\pm 0.316}}^{[0.82/0.36/0.05]}$ & $0.940_{\tiny{\pm 0.162}}^{[0.823/0.826/1.17]}$ \\
  SRI & \textbf{71.0}\tiny{$\pm 3$} & \textbf{0.78}\tiny{$\pm0.02$} &  \textbf{0.829}\tiny{$\pm 0.001$} \\
 \bottomrule
\end{tabular}
\end{adjustbox}
\end{table}
}

%% file: tables/NEURCOMP-D-23-00020R1-quant.tex
\begin{table}[t]
\caption[Objects Room and ShapeStacks quantitative results]{\textbf{Main quantitative results}. Mean $\pm$ std. dev. over 3 training runs. Results* are from Table~3 in~\citet{engelcke2021genesis}.}
\label{tab:quant}
\centering
\begin{adjustbox}{max width=\textwidth}
\begin{tabular}{lccccccc}
\toprule
\multicolumn{1}{c}{} & 
\multicolumn{2}{c}{\textbf{CLEVR6}}  &
  \multicolumn{2}{c}{\textbf{Objects Room}}  &
  \multicolumn{3}{c}{\textbf{ShapeStacks}}  \\
   \cmidrule(lr){2-3}\cmidrule(lr){4-5}\cmidrule(lr){6-8}
\multicolumn{1}{l}{Model} &
 
  \multicolumn{1}{c}{FID\tiny{$\downarrow$}}  &
   \multicolumn{1}{c}{ARI-FG\tiny{$\uparrow$}}  &
   \multicolumn{1}{c}{FID\tiny{$\downarrow$}}  &
   \multicolumn{1}{c}{ARI-FG\tiny{$\uparrow$}}  &
  \multicolumn{1}{c}{FID\tiny{$\downarrow$}} &
   \multicolumn{1}{c}{ARI-FG\tiny{$\uparrow$}}  &
  \multicolumn{1}{c}{S-Acc (\%)\tiny{$\uparrow$}}\\ \midrule
EMORL              &  244.0\tiny{$\pm 19$}  & 0.96\tiny{$\pm 0.02$}         & 178.3\tiny{$\pm 27$} & 0.47\tiny{$\pm 0.22$} & 258.4\tiny{$\pm 57$} & 0.60\tiny{$\pm 0.04$} &  - \\
GEN                            & 116.9\tiny{$\pm4$}  & 0.82\tiny{$\pm0.07$} & 62.8*\tiny{$\pm 3$} & 0.63*\tiny{$\pm 0.03$} & 186.8*\tiny{$\pm 18$}   & 0.70*\tiny{$\pm 0.05$} & -  \\
\midrule
  GENv2-G       &  61.0\tiny{$\pm 3$} & 0.98\tiny{$\pm 0.00$} & 87.6\tiny{$\pm 4$}  & 0.75\tiny{$\pm 0.02$} & 115.3\tiny{$\pm 6$} & 0.68\tiny{$\pm 0.02$} & 50.7\tiny{$\pm 3.1$}  \\
  SRI-G  & \textbf{60.6}\tiny{$\pm 8$} & 0.96\tiny{$\pm 0.02$} & \textbf{55.7}\tiny{$\pm 4$} & 0.74\tiny{$\pm 0.01$} & \textbf{74.7}\tiny{$\pm 5$} & 0.70\tiny{$\pm 0.01$} &  \textbf{72.0}\tiny{$\pm 3.5$} \\
 \midrule
GENv2-MoG  &  61.0\tiny{$\pm 3$} & 0.98\tiny{$\pm 0.00$}& 52.6*\tiny{$\pm 3$}  & 0.85*\tiny{$\pm 0.01$} & 112.7*\tiny{$\pm 3$} & 0.81*\tiny{$\pm 0.00$} & 59  \\
  SRI-MoG  & \textbf{54.4}\tiny{$\pm 2$}  & 0.97\tiny{$\pm0.01$}                           & \textbf{48.4}\tiny{$\pm 4$} & 0.83\tiny{$\pm0.02$} & \textbf{70.4}\tiny{$\pm 3$} & 0.78\tiny{$\pm0.02$} & \textbf{80.7}\tiny{$\pm 5.0$}\\
\bottomrule
\end{tabular}
\end{adjustbox}
\end{table}

%% file: tables/NEURCOMP-D-23-00020R1-ablations.tex
{\rowcolors{2}{gray!25}{white}
\begin{table}[h!]
\centering
\caption{\textbf{SRI ablations.} SRI-G on ShapeStacks, mean $\pm$ std. dev. over 3 runs. The negative ELBO in bits-per-dim (bpd) is a single sample Monte Carlo estimate averaged over 5k validation scenes.}
\label{tab:ablations}
\begin{adjustbox}{max width=\textwidth}
\begin{tabular}{@{}ccccclcc@{}}
\toprule
 & Learned order & Scene variable &
    $\mathcal{L}_{\text{crossOrderKL}}$ &
  FID\tiny{$\downarrow$} & & 
  $\mathcal{L}_{\text{SRI}}$(bpd)\tiny{$\downarrow$} &
  S-Acc (\%)\tiny{$\uparrow$} \\ \midrule
  a) & \ding{51}& & \ding{51} & 132.9{\tiny{$\pm 2$}} &  ({\textcolor{red}{-57.9}}) &  0.836\tiny{$\pm 0.001$}      & 31.7\tiny{$\pm 6.4$} \\
  b) &  & \ding{51} & - &  123.2\tiny{$\pm 3$}& ({\textcolor{red}{-48.2}}) & 0.845\tiny{$\pm 0.001$} & 43.7\tiny{$\pm 10.7$}\\
  c) & \ding{51} & \ding{51} &           &      106.3\tiny{$\pm 3$} & ({\textcolor{red}{-31.3}}) & 0.833\tiny{$\pm 0.001$} & 61.3\tiny{$\pm 1.5$} \\
 d) & \ding{51} & \ding{51} & \ding{51} & \textbf{75.0}{\tiny{$\pm 6$}}& (+0.0)  & \textbf{0.829}\tiny{$\pm 0.001$} &  \textbf{72.0}\tiny{$\pm 3.5$} \\
 \bottomrule
\end{tabular}
\end{adjustbox}
\end{table}
}

%% file: tables/NEURCOMP-D-23-00020R1-scene_prior_ablations.tex
{
\begin{table}[t]
\centering
\caption{\textbf{Scene variable ablation.} Mean $\pm$ std. dev. over 3 Shapestacks runs except for results$^*$ computed with model weights released by the authors. The negative ELBO in bits-per-dim (bpd) is a single sample Monte Carlo estimate averaged over 5k validation scenes.}
\label{tab:scene_ablations}
\begin{adjustbox}{max width=\columnwidth}
\begin{tabular}{@{}llll@{}}
\toprule
 Model &
  FID\tiny{$\downarrow$} & 
  -ELBO (bpd)\tiny{$\downarrow$} &
  S-Acc (\%)\tiny{$\uparrow$} \\ \midrule
  GEN$^*$ & 233.3 &  0.832 & - \\
  GEN + scene variable & 217.5\tiny{$\pm22$} & 1.051\tiny{$\pm0.120$} & -\\
  GENv2$^*$ & 107.4 & 0.821 & 59\\
  GENv2 + scene variable & 84.5\tiny{$\pm 4$} & 0.823\tiny{$\pm 0.000$} & 66.3\tiny{$\pm10.7$}\\
  \midrule
  SRI & 71.0\tiny{$\pm 3$} & 0.829\tiny{$\pm 0.001$}& 80.7\tiny{$\pm 5.0$} \\
 \bottomrule
\end{tabular}
\end{adjustbox}
\end{table}
}

%% file: algorithms/NEURCOMP-D-23-00020R1-matching.tex
\begin{algorithm}[th]
\renewcommand{\baselinestretch}{1.0}
    \caption{\textbf{Greedy matching}. Uses $O(K)$ time and $O(K^2)$ space.}
    \label{alg:matching}
    \begin{algorithmic}[1]
    
    \STATE \textbf{Input:} Randomly ordered segregation posterior means $\mu_{o_{1:K}}$, imagination rollout means $\mu_{\pi_{1:K}}$
    \STATE $\mC[i,j] = \lVert \mu_{\pi_i} - \mu_{o_j}\rVert_2,$ \hspace{1pt} $ \forall i,j = 1,\dots,K$
    \STATE $\sigma := \texttt{zeros}(K, K)$
    \FOR{index $i = 0\dots K-1$}
        \STATE $j^* = \argmin \mC[i,:]$
        \STATE $\mC[:,j^*] = +\inf$
        \STATE $\sigma[j^*,i] = 1$
    \ENDFOR
    \RETURN Permutation $\sigma$
    \end{algorithmic}
\end{algorithm}

%% file: algorithms/NEURCOMP-D-23-00020R1-inference.tex
\begin{algorithm}[h]
    \caption{\textbf{SRI Inference}. All sampling uses the Gaussian reparameterization trick~\citep{kingma2013auto}.}
    \label{alg:inference}
    \begin{algorithmic}[1]
    \STATE \textbf{Input:} Scene observation $\rvx$, segregation VAE encoder $e_\psi(\rvx)$, scene-level encoder $f_\phi(\rvz)$, autoregressive prior $\texttt{LSTM}_\theta$, autoregressive posterior $\texttt{LSTM}_\phi$
    \STATE $\rvz_{o_{1:K}} \sim q_{\psi}(\rvz_{o_{1:K}} \mid \rvx) \leftarrow e_\psi(\rvx)$ \textcolor{gray}{ /* Segregate */ }
    \STATE $\rvs \sim q_\phi(\rvs \mid \rvx) \leftarrow f_\phi(\rvz_{o_{1:K}})$ \textcolor{gray}{ /* Relate */ }
    \STATE $\rvz_{\pi_1} \sim p_\theta(\rvz_{\pi_1} \mid \rvs) \leftarrow \texttt{LSTM}_\theta(\{\rvs\})$  \textcolor{gray}{ /* Imagine */ }
    \STATE $\rvz_{\pi_k} \sim p_\theta( \rvz_{\pi_{k}} \mid \rvz_{\pi_{1:k-1}}, \rvs ) \leftarrow \texttt{LSTM}_\theta(\{\rvs, \rvz_{\pi_{1:k-1}}\})$ for all $k = 2,\dots,K-1$
    \STATE $\sigma \leftarrow $  Matching $(\rvz_{o_{1:K}}, \rvz_{\pi_{1:K}})$  \textcolor{gray}{ /* we pass the posterior means to Algorithm~\ref{alg:matching} */}
    \STATE $\mu_{o_{1:K}} \xrightarrow{\sigma} \mu_{\hat{\pi}_{1:K}}$ \textcolor{gray}{ /* apply the permutation to the $K$ means of $q_\phi(\rvz_{o_{1:K}} \mid \rvx)$ */}
    \STATE $\bar{\sigma}^2_{\hat{\pi}_1} \leftarrow \texttt{LSTM}_\phi(\{\rvs\})$  \textcolor{gray}{ /* predict new correlated variances */}
    \STATE $\bar{\sigma}^2_{\hat{\pi}_k} \leftarrow \texttt{LSTM}_\phi(\{\rvs, \mu_{\hat{\pi}_{1:k-1}}\})$ for all $k = 2,\dots,K-1$
    \STATE $q_\phi(\rvz_{\hat{\pi}_{1:K}} \mid \rvs, \rvx) := \prod_{k=1}^K \mathcal{N}(\mu_{\hat{\pi}_k}, \bar{\sigma}^2_{\hat{\pi}_k})$ \textcolor{gray}{ /* autoregressive slot posterior */}
    \RETURN The sufficient statistics (e.g., means and variances) for the segregation posterior $q_{\psi}(\rvz_{o_{1:K}} \mid \rvx)$, scene posterior $q_\phi(\rvs \mid \rvx)$, autoregressive slot posterior $q_\phi(\rvz_{\hat{\pi}_{1:K}} \mid \rvs, \rvx)$
    \end{algorithmic}
\end{algorithm}

%% file: NEURCOMP-D-23-00020R1.bbl
\begin{thebibliography}{79}
\providecommand{\natexlab}[1]{#1}
\providecommand{\url}[1]{\texttt{#1}}
\expandafter\ifx\csname urlstyle\endcsname\relax
  \providecommand{\doi}[1]{doi: #1}\else
  \providecommand{\doi}{doi: \begingroup \urlstyle{rm}\Url}\fi

\bibitem[Adams and Zemel(2011)]{adams2011ranking}
Ryan~Prescott Adams and Richard~S Zemel.
\newblock Ranking via sinkhorn propagation.
\newblock \emph{ArXiv preprint}, abs/1106.1925, 2011.

\bibitem[Anciukevicius et~al.(2020)Anciukevicius, Lampert, and Henderson]{anciukevicius2020object}
Titas Anciukevicius, Christoph~H Lampert, and Paul Henderson.
\newblock Object-centric image generation with factored depths, locations, and appearances.
\newblock \emph{ArXiv preprint}, abs/2004.00642, 2020.

\bibitem[Baillargeon(2004)]{baillargeon2004infants}
Ren{\'e}e Baillargeon.
\newblock Infants' physical world.
\newblock \emph{Current directions in psychological science}, 13\penalty0 (3):\penalty0 89--94, 2004.

\bibitem[Battaglia et~al.(2013)Battaglia, Hamrick, and Tenenbaum]{battaglia2013simulation}
Peter~W Battaglia, Jessica~B Hamrick, and Joshua~B Tenenbaum.
\newblock Simulation as an engine of physical scene understanding.
\newblock \emph{Proceedings of the National Academy of Sciences}, 110\penalty0 (45):\penalty0 18327--18332, 2013.

\bibitem[Burgess et~al.(2019)Burgess, Matthey, Watters, Kabra, Higgins, Botvinick, and Lerchner]{burgess2019monet}
Christopher~P Burgess, Loic Matthey, Nicholas Watters, Rishabh Kabra, Irina Higgins, Matt Botvinick, and Alexander Lerchner.
\newblock {MONet}: Unsupervised scene decomposition and representation.
\newblock \emph{ArXiv preprint}, abs/1901.11390, 2019.

\bibitem[Chen et~al.(2020)Chen, Radford, Child, Wu, Jun, Luan, and Sutskever]{chen2020generative}
Mark Chen, Alec Radford, Rewon Child, Jeffrey Wu, Heewoo Jun, David Luan, and Ilya Sutskever.
\newblock Generative pretraining from pixels.
\newblock In \emph{Proceedings of the 37th International Conference on Machine Learning, {ICML} 2020, 13-18 July 2020, Virtual Event}, volume 119 of \emph{Proceedings of Machine Learning Research}, pages 1691--1703. {PMLR}, 2020.

\bibitem[Child et~al.(2019)Child, Gray, Radford, and Sutskever]{child2019generating}
Rewon Child, Scott Gray, Alec Radford, and Ilya Sutskever.
\newblock Generating long sequences with sparse transformers.
\newblock \emph{ArXiv preprint}, abs/1904.10509, 2019.

\bibitem[Crawford and Pineau(2019)]{crawford2019spatially}
Eric Crawford and Joelle Pineau.
\newblock Spatially invariant unsupervised object detection with convolutional neural networks.
\newblock In \emph{The Thirty-Third {AAAI} Conference on Artificial Intelligence, {AAAI} 2019, The Thirty-First Innovative Applications of Artificial Intelligence Conference, {IAAI} 2019, The Ninth {AAAI} Symposium on Educational Advances in Artificial Intelligence, {EAAI} 2019, Honolulu, Hawaii, USA, January 27 - February 1, 2019}, pages 3412--3420. {AAAI} Press, 2019.
\newblock \doi{10.1609/aaai.v33i01.33013412}.

\bibitem[Creswell et~al.(2021)Creswell, Kabra, Burgess, and Shanahan]{creswell2021unsupervised}
Antonia Creswell, Rishabh Kabra, Chris Burgess, and Murray Shanahan.
\newblock Unsupervised object-based transition models for 3d partially observable environments.
\newblock \emph{Advances in Neural Information Processing Systems}, 34, 2021.

\bibitem[Deng et~al.(2021)Deng, Zhi, Lee, and Ahn]{deng2021generative}
Fei Deng, Zhuo Zhi, Donghun Lee, and Sungjin Ahn.
\newblock Generative scene graph networks.
\newblock In \emph{9th International Conference on Learning Representations, {ICLR} 2021, Virtual Event, Austria, May 3-7, 2021}. OpenReview.net, 2021.

\bibitem[Emami et~al.(2021)Emami, He, Ranka, and Rangarajan]{pmlr-v139-emami21a}
Patrick Emami, Pan He, Sanjay Ranka, and Anand Rangarajan.
\newblock Efficient iterative amortized inference for learning symmetric and disentangled multi-object representations.
\newblock In Marina Meila and Tong Zhang, editors, \emph{Proceedings of the 38th International Conference on Machine Learning, {ICML} 2021, 18-24 July 2021, Virtual Event}, volume 139 of \emph{Proceedings of Machine Learning Research}, pages 2970--2981. {PMLR}, 2021.

\bibitem[Engelcke et~al.(2020)Engelcke, Kosiorek, Jones, and Posner]{engelcke2019genesis}
Martin Engelcke, Adam~R. Kosiorek, Oiwi~Parker Jones, and Ingmar Posner.
\newblock {GENESIS:} generative scene inference and sampling with object-centric latent representations.
\newblock In \emph{8th International Conference on Learning Representations, {ICLR} 2020, Addis Ababa, Ethiopia, April 26-30, 2020}. OpenReview.net, 2020.

\bibitem[Engelcke et~al.(2021)Engelcke, Parker~Jones, and Posner]{engelcke2021genesis}
Martin Engelcke, Oiwi Parker~Jones, and Ingmar Posner.
\newblock {GENESIS-V2}: Inferring unordered object representations without iterative refinement.
\newblock In M.~Ranzato, A.~Beygelzimer, Y.~Dauphin, P.S. Liang, and J.~Wortman Vaughan, editors, \emph{Advances in Neural Information Processing Systems}, volume~34, pages 8085--8094. Curran Associates, Inc., 2021.

\bibitem[Eslami et~al.(2016)Eslami, Heess, Weber, Tassa, Szepesvari, Kavukcuoglu, and Hinton]{eslami2016attend}
S.~M.~Ali Eslami, Nicolas Heess, Theophane Weber, Yuval Tassa, David Szepesvari, Koray Kavukcuoglu, and Geoffrey~E. Hinton.
\newblock Attend, infer, repeat: Fast scene understanding with generative models.
\newblock In Daniel~D. Lee, Masashi Sugiyama, Ulrike von Luxburg, Isabelle Guyon, and Roman Garnett, editors, \emph{Advances in Neural Information Processing Systems 29: Annual Conference on Neural Information Processing Systems 2016, December 5-10, 2016, Barcelona, Spain}, pages 3225--3233, 2016.

\bibitem[Germain et~al.(2015)Germain, Gregor, Murray, and Larochelle]{germain2015made}
Mathieu Germain, Karol Gregor, Iain Murray, and Hugo Larochelle.
\newblock {MADE:} masked autoencoder for distribution estimation.
\newblock In Francis~R. Bach and David~M. Blei, editors, \emph{Proceedings of the 32nd International Conference on Machine Learning, {ICML} 2015, Lille, France, 6-11 July 2015}, volume~37 of \emph{{JMLR} Workshop and Conference Proceedings}, pages 881--889. JMLR.org, 2015.

\bibitem[Glorot and Bengio(2010)]{glorot2010understanding}
Xavier Glorot and Yoshua Bengio.
\newblock Understanding the difficulty of training deep feedforward neural networks.
\newblock In \emph{Proceedings of the thirteenth international conference on artificial intelligence and statistics}, pages 249--256. JMLR Workshop and Conference Proceedings, 2010.

\bibitem[Gopalakrishnan et~al.(2023)Gopalakrishnan, Irie, Schmidhuber, and van Steenkiste]{gopalakrishnan2023unsupervised}
Anand Gopalakrishnan, Kazuki Irie, J{\"u}rgen Schmidhuber, and Sjoerd van Steenkiste.
\newblock Unsupervised learning of temporal abstractions with slot-based transformers.
\newblock \emph{Neural Computation}, 35\penalty0 (4):\penalty0 593--626, 2023.

\bibitem[Greff et~al.(2017)Greff, Klein, Chovanec, Hutter, and Schmidhuber]{greff2017sacred}
Klaus Greff, Aaron Klein, Martin Chovanec, Frank Hutter, and J{\"u}rgen Schmidhuber.
\newblock The sacred infrastructure for computational research.
\newblock In \emph{Proceedings of the 15th python in science conference (SciPy 2017)}, volume~28, pages 49--56, 2017.

\bibitem[Greff et~al.(2019)Greff, Kaufman, Kabra, Watters, Burgess, Zoran, Matthey, Botvinick, and Lerchner]{pmlr-v97-greff19a}
Klaus Greff, Rapha{\"{e}}l~Lopez Kaufman, Rishabh Kabra, Nick Watters, Christopher Burgess, Daniel Zoran, Loic Matthey, Matthew Botvinick, and Alexander Lerchner.
\newblock Multi-object representation learning with iterative variational inference.
\newblock In Kamalika Chaudhuri and Ruslan Salakhutdinov, editors, \emph{Proceedings of the 36th International Conference on Machine Learning, {ICML} 2019, 9-15 June 2019, Long Beach, California, {USA}}, volume~97 of \emph{Proceedings of Machine Learning Research}, pages 2424--2433. {PMLR}, 2019.

\bibitem[Greff et~al.(2020)Greff, van Steenkiste, and Schmidhuber]{greff2020binding}
Klaus Greff, Sjoerd van Steenkiste, and J{\"u}rgen Schmidhuber.
\newblock On the binding problem in artificial neural networks.
\newblock \emph{ArXiv preprint}, abs/2012.05208, 2020.

\bibitem[Groth et~al.(2018)Groth, Fuchs, Posner, and Vedaldi]{groth2018shapestacks}
Oliver Groth, Fabian~B Fuchs, Ingmar Posner, and Andrea Vedaldi.
\newblock Shapestacks: Learning vision-based physical intuition for generalised object stacking.
\newblock In \emph{Proceedings of the European Conference on Computer Vision (ECCV)}, pages 702--717, 2018.

\bibitem[Grover et~al.(2019)Grover, Wang, Zweig, and Ermon]{grover2019stochastic}
Aditya Grover, Eric Wang, Aaron Zweig, and Stefano Ermon.
\newblock Stochastic optimization of sorting networks via continuous relaxations.
\newblock In \emph{7th International Conference on Learning Representations, {ICLR} 2019, New Orleans, LA, USA, May 6-9, 2019}. OpenReview.net, 2019.

\bibitem[Harris et~al.(2020)Harris, Millman, van~der Walt, Gommers, Virtanen, Cournapeau, Wieser, Taylor, Berg, Smith, Kern, Picus, Hoyer, van Kerkwijk, Brett, Haldane, del Río, Wiebe, Peterson, Gérard-Marchant, Sheppard, Reddy, Weckesser, Abbasi, Gohlke, and Oliphant]{numpy}
Charles~R. Harris, K.~Jarrod Millman, Stéfan~J. van~der Walt, Ralf Gommers, Pauli Virtanen, David Cournapeau, Eric Wieser, Julian Taylor, Sebastian Berg, Nathaniel~J. Smith, Robert Kern, Matti Picus, Stephan Hoyer, Marten~H. van Kerkwijk, Matthew Brett, Allan Haldane, Jaime~Fernández del Río, Mark Wiebe, Pearu Peterson, Pierre Gérard-Marchant, Kevin Sheppard, Tyler Reddy, Warren Weckesser, Hameer Abbasi, Christoph Gohlke, and Travis~E. Oliphant.
\newblock Array programming with {NumPy}.
\newblock \emph{Nature}, 585\penalty0 (7825):\penalty0 357--362, 2020.
\newblock ISSN 1476-4687.
\newblock \doi{10.1038/s41586-020-2649-2}.

\bibitem[Hendrycks and Gimpel(2016)]{hendrycks2016gaussian}
Dan Hendrycks and Kevin Gimpel.
\newblock Gaussian error linear units (gelus).
\newblock \emph{ArXiv preprint}, abs/1606.08415, 2016.

\bibitem[Heusel et~al.(2017)Heusel, Ramsauer, Unterthiner, Nessler, and Hochreiter]{heusel2017gans}
Martin Heusel, Hubert Ramsauer, Thomas Unterthiner, Bernhard Nessler, and Sepp Hochreiter.
\newblock {GANs} trained by a two time-scale update rule converge to a local nash equilibrium.
\newblock In Isabelle Guyon, Ulrike von Luxburg, Samy Bengio, Hanna~M. Wallach, Rob Fergus, S.~V.~N. Vishwanathan, and Roman Garnett, editors, \emph{Advances in Neural Information Processing Systems 30: Annual Conference on Neural Information Processing Systems 2017, December 4-9, 2017, Long Beach, CA, {USA}}, pages 6626--6637, 2017.

\bibitem[Hochreiter and Schmidhuber(1997)]{hochreiter1997long}
Sepp Hochreiter and J{\"u}rgen Schmidhuber.
\newblock Long short-term memory.
\newblock \emph{Neural computation}, 9\penalty0 (8):\penalty0 1735--1780, 1997.

\bibitem[Hubert and Arabie(1985)]{hubert1985comparing}
Lawrence Hubert and Phipps Arabie.
\newblock Comparing partitions.
\newblock \emph{Journal of classification}, 2\penalty0 (1):\penalty0 193--218, 1985.

\bibitem[{Hunter}(2007)]{matplotlib}
J.~D. {Hunter}.
\newblock {Matplotlib: A 2D Graphics Environment}.
\newblock \emph{{C}omputing in {S}cience \& {E}ngineering}, 9\penalty0 (3):\penalty0 90--95, 2007.

\bibitem[Jiang and Ahn(2020)]{jiang2020generative}
Jindong Jiang and Sungjin Ahn.
\newblock Generative neurosymbolic machines.
\newblock In Hugo Larochelle, Marc'Aurelio Ranzato, Raia Hadsell, Maria{-}Florina Balcan, and Hsuan{-}Tien Lin, editors, \emph{Advances in Neural Information Processing Systems 33: Annual Conference on Neural Information Processing Systems 2020, NeurIPS 2020, December 6-12, 2020, virtual}, 2020.

\bibitem[Jurewicz and Derczynski(2021)]{jurewicz2021set}
Mateusz Jurewicz and Leon Derczynski.
\newblock Set-to-sequence methods in machine learning: a review.
\newblock \emph{Journal of Artificial Intelligence Research}, 71:\penalty0 885--924, 2021.

\bibitem[Kabra et~al.(2019)Kabra, Burgess, Matthey, Kaufman, Greff, Reynolds, and Lerchner]{multiobjectdatasets19}
Rishabh Kabra, Chris Burgess, Loic Matthey, Raphael~Lopez Kaufman, Klaus Greff, Malcolm Reynolds, and Alexander Lerchner.
\newblock Multi-object datasets, 2019.

\bibitem[Kabra et~al.(2021)Kabra, Zoran, Erdogan, Matthey, Creswell, Botvinick, Lerchner, and Burgess]{kabra2021simone}
Rishabh Kabra, Daniel Zoran, Goker Erdogan, Loic Matthey, Antonia Creswell, Matt Botvinick, Alexander Lerchner, and Chris Burgess.
\newblock {SIMONe}: View-invariant, temporally-abstracted object representations via unsupervised video decomposition.
\newblock In M.~Ranzato, A.~Beygelzimer, Y.~Dauphin, P.S. Liang, and J.~Wortman Vaughan, editors, \emph{Advances in Neural Information Processing Systems}, volume~34, pages 20146--20159. Curran Associates, Inc., 2021.

\bibitem[Kahneman et~al.(1992)Kahneman, Treisman, and Gibbs]{kahneman1992reviewing}
Daniel Kahneman, Anne Treisman, and Brian~J Gibbs.
\newblock The reviewing of object files: Object-specific integration of information.
\newblock \emph{Cognitive psychology}, 24\penalty0 (2):\penalty0 175--219, 1992.

\bibitem[Kingma and Ba(2015)]{kingma2014adam}
Diederik~P. Kingma and Jimmy Ba.
\newblock Adam: {A} method for stochastic optimization.
\newblock In Yoshua Bengio and Yann LeCun, editors, \emph{3rd International Conference on Learning Representations, {ICLR} 2015, San Diego, CA, USA, May 7-9, 2015, Conference Track Proceedings}, 2015.

\bibitem[Kingma and Welling(2014)]{kingma2013auto}
Diederik~P. Kingma and Max Welling.
\newblock Auto-encoding variational bayes.
\newblock In Yoshua Bengio and Yann LeCun, editors, \emph{2nd International Conference on Learning Representations, {ICLR} 2014, Banff, AB, Canada, April 14-16, 2014, Conference Track Proceedings}, 2014.

\bibitem[Kluyver et~al.(2016)Kluyver, Ragan-Kelley, P{\'e}rez, Granger, Bussonnier, Frederic, Kelley, Hamrick, Grout, Corlay, Ivanov, Avila, Abdalla, and Willing]{jupyter}
Thomas Kluyver, Benjamin Ragan-Kelley, Fernando P{\'e}rez, Brian Granger, Matthias Bussonnier, Jonathan Frederic, Kyle Kelley, Jessica Hamrick, Jason Grout, Sylvain Corlay, Paul Ivanov, Dami{\'a}n Avila, Safia Abdalla, and Carol Willing.
\newblock {J}upyter {N}otebooks -- a publishing format for reproducible computational workflows.
\newblock In F.~Loizides and B.~Schmidt, editors, \emph{Positioning and Power in Academic Publishing: Players, Agents and Agendas}, pages 87 -- 90. IOS Press, 2016.

\bibitem[Lake et~al.(2017)Lake, Ullman, Tenenbaum, and Gershman]{lake2017building}
Brenden~M Lake, Tomer~D Ullman, Joshua~B Tenenbaum, and Samuel~J Gershman.
\newblock Building machines that learn and think like people.
\newblock \emph{Behavioral and brain sciences}, 40, 2017.

\bibitem[Li et~al.(2020)Li, Eastwood, and Fisher]{nanbo2020learning}
Nanbo Li, Cian Eastwood, and Robert~B. Fisher.
\newblock Learning object-centric representations of multi-object scenes from multiple views.
\newblock In Hugo Larochelle, Marc'Aurelio Ranzato, Raia Hadsell, Maria{-}Florina Balcan, and Hsuan{-}Tien Lin, editors, \emph{Advances in Neural Information Processing Systems 33: Annual Conference on Neural Information Processing Systems 2020, NeurIPS 2020, December 6-12, 2020, virtual}, 2020.

\bibitem[Li et~al.(2021{\natexlab{a}})Li, Raza, Hu, Sun, and Fisher]{li2021object}
Nanbo Li, Muhammad~Ahmed Raza, Wenbin Hu, Zhaole Sun, and Robert Fisher.
\newblock Object-centric representation learning with generative spatial-temporal factorization.
\newblock \emph{Advances in Neural Information Processing Systems}, 34, 2021{\natexlab{a}}.

\bibitem[Li et~al.(2021{\natexlab{b}})Li, Trabucco, Park, Luo, Shen, Darrell, and Gao]{li2021discovering}
Xuanlin Li, Brandon Trabucco, Dong~Huk Park, Michael Luo, Sheng Shen, Trevor Darrell, and Yang Gao.
\newblock Discovering non-monotonic autoregressive orderings with variational inference.
\newblock In \emph{9th International Conference on Learning Representations, {ICLR} 2021, Virtual Event, Austria, May 3-7, 2021}. OpenReview.net, 2021{\natexlab{b}}.

\bibitem[Lin et~al.(2020{\natexlab{a}})Lin, Wu, Peri, Fu, Jiang, and Ahn]{linimproving}
Zhixuan Lin, Yi{-}Fu Wu, Skand~Vishwanath Peri, Bofeng Fu, Jindong Jiang, and Sungjin Ahn.
\newblock Improving generative imagination in object-centric world models.
\newblock In \emph{Proceedings of the 37th International Conference on Machine Learning, {ICML} 2020, 13-18 July 2020, Virtual Event}, volume 119 of \emph{Proceedings of Machine Learning Research}, pages 6140--6149. {PMLR}, 2020{\natexlab{a}}.

\bibitem[Lin et~al.(2020{\natexlab{b}})Lin, Wu, Peri, Sun, Singh, Deng, Jiang, and Ahn]{lin2020space}
Zhixuan Lin, Yi{-}Fu Wu, Skand~Vishwanath Peri, Weihao Sun, Gautam Singh, Fei Deng, Jindong Jiang, and Sungjin Ahn.
\newblock {SPACE:} unsupervised object-oriented scene representation via spatial attention and decomposition.
\newblock In \emph{8th International Conference on Learning Representations, {ICLR} 2020, Addis Ababa, Ethiopia, April 26-30, 2020}. OpenReview.net, 2020{\natexlab{b}}.

\bibitem[Locatello et~al.(2020)Locatello, Weissenborn, Unterthiner, Mahendran, Heigold, Uszkoreit, Dosovitskiy, and Kipf]{locatello2020object}
Francesco Locatello, Dirk Weissenborn, Thomas Unterthiner, Aravindh Mahendran, Georg Heigold, Jakob Uszkoreit, Alexey Dosovitskiy, and Thomas Kipf.
\newblock Object-centric learning with slot attention.
\newblock In Hugo Larochelle, Marc'Aurelio Ranzato, Raia Hadsell, Maria{-}Florina Balcan, and Hsuan{-}Tien Lin, editors, \emph{Advances in Neural Information Processing Systems 33: Annual Conference on Neural Information Processing Systems 2020, NeurIPS 2020, December 6-12, 2020, virtual}, 2020.

\bibitem[Mena et~al.(2018)Mena, Belanger, Linderman, and Snoek]{mena2018learning}
Gonzalo~E. Mena, David Belanger, Scott~W. Linderman, and Jasper Snoek.
\newblock Learning latent permutations with gumbel-sinkhorn networks.
\newblock In \emph{6th International Conference on Learning Representations, {ICLR} 2018, Vancouver, BC, Canada, April 30 - May 3, 2018, Conference Track Proceedings}. OpenReview.net, 2018.

\bibitem[Mishkin et~al.(2022)Mishkin, Ahmad, Brundage, Krueger, and Sastry]{mishkin2022risks}
Pamela Mishkin, Lama Ahmad, Miles Brundage, Gretchen Krueger, and Girish Sastry.
\newblock Dall·e 2 preview - risks and limitations.
\newblock 2022.

\bibitem[Munkres(1957)]{munkres1957algorithms}
James Munkres.
\newblock Algorithms for the assignment and transportation problems.
\newblock \emph{Journal of the society for industrial and applied mathematics}, 5\penalty0 (1):\penalty0 32--38, 1957.

\bibitem[Parmar et~al.(2018)Parmar, Vaswani, Uszkoreit, Kaiser, Shazeer, Ku, and Tran]{parmar2018image}
Niki Parmar, Ashish Vaswani, Jakob Uszkoreit, Lukasz Kaiser, Noam Shazeer, Alexander Ku, and Dustin Tran.
\newblock Image transformer.
\newblock In Jennifer~G. Dy and Andreas Krause, editors, \emph{Proceedings of the 35th International Conference on Machine Learning, {ICML} 2018, Stockholmsm{\"{a}}ssan, Stockholm, Sweden, July 10-15, 2018}, volume~80 of \emph{Proceedings of Machine Learning Research}, pages 4052--4061. {PMLR}, 2018.

\bibitem[Paszke et~al.(2019)Paszke, Gross, Massa, Lerer, Bradbury, Chanan, Killeen, Lin, Gimelshein, Antiga, Desmaison, Kopf, Yang, DeVito, Raison, Tejani, Chilamkurthy, Steiner, Fang, Bai, and Chintala]{torch}
Adam Paszke, Sam Gross, Francisco Massa, Adam Lerer, James Bradbury, Gregory Chanan, Trevor Killeen, Zeming Lin, Natalia Gimelshein, Luca Antiga, Alban Desmaison, Andreas Kopf, Edward Yang, Zachary DeVito, Martin Raison, Alykhan Tejani, Sasank Chilamkurthy, Benoit Steiner, Lu~Fang, Junjie Bai, and Soumith Chintala.
\newblock {P}y{T}orch: {A}n {I}mperative {S}tyle, {H}igh-{P}erformance {D}eep {L}earning {L}ibrary.
\newblock In \emph{Advances in Neural Information Processing Systems 32}, pages 8024--8035. Curran Associates, Inc., 2019.

\bibitem[Pedregosa et~al.(2011)Pedregosa, Varoquaux, Gramfort, Michel, Thirion, Grisel, Blondel, Prettenhofer, Weiss, Dubourg, Vanderplas, Passos, Cournapeau, Brucher, Perrot, and {{\'E}}douard Duchesnay]{sklearn}
Fabian Pedregosa, Ga{{\"e}}l Varoquaux, Alexandre Gramfort, Vincent Michel, Bertrand Thirion, Olivier Grisel, Mathieu Blondel, Peter Prettenhofer, Ron Weiss, Vincent Dubourg, Jake Vanderplas, Alexandre Passos, David Cournapeau, Matthieu Brucher, Matthieu Perrot, and {{\'E}}douard Duchesnay.
\newblock {Scikit-learn: Machine Learning in Python}.
\newblock \emph{{J}ournal of {M}achine {L}earning {R}esearch}, 12\penalty0 (85):\penalty0 2825--2830, 2011.

\bibitem[Rand(1971)]{rand1971objective}
William~M Rand.
\newblock Objective criteria for the evaluation of clustering methods.
\newblock \emph{Journal of the American Statistical association}, 66\penalty0 (336):\penalty0 846--850, 1971.

\bibitem[Rezende and Viola(2018)]{rezende2018taming}
Danilo~Jimenez Rezende and Fabio Viola.
\newblock Taming {VAEs}.
\newblock \emph{ArXiv preprint}, abs/1810.00597, 2018.

\bibitem[Rezende et~al.(2014)Rezende, Mohamed, and Wierstra]{Rezende2014}
Danilo~Jimenez Rezende, Shakir Mohamed, and Daan Wierstra.
\newblock Stochastic backpropagation and approximate inference in deep generative models.
\newblock In \emph{Proceedings of the 31th International Conference on Machine Learning, {ICML} 2014, Beijing, China, 21-26 June 2014}, volume~32 of \emph{{JMLR} Workshop and Conference Proceedings}, pages 1278--1286. JMLR.org, 2014.

\bibitem[Salimans et~al.(2017)Salimans, Karpathy, Chen, and Kingma]{salimans2017pixelcnn++}
Tim Salimans, Andrej Karpathy, Xi~Chen, and Diederik~P. Kingma.
\newblock Pixelcnn++: Improving the pixelcnn with discretized logistic mixture likelihood and other modifications.
\newblock In \emph{5th International Conference on Learning Representations, {ICLR} 2017, Toulon, France, April 24-26, 2017, Conference Track Proceedings}. OpenReview.net, 2017.

\bibitem[Santoro et~al.(2017)Santoro, Raposo, Barrett, Malinowski, Pascanu, Battaglia, and Lillicrap]{santoro2017simple}
Adam Santoro, David Raposo, David G.~T. Barrett, Mateusz Malinowski, Razvan Pascanu, Peter~W. Battaglia, and Tim Lillicrap.
\newblock A simple neural network module for relational reasoning.
\newblock In Isabelle Guyon, Ulrike von Luxburg, Samy Bengio, Hanna~M. Wallach, Rob Fergus, S.~V.~N. Vishwanathan, and Roman Garnett, editors, \emph{Advances in Neural Information Processing Systems 30: Annual Conference on Neural Information Processing Systems 2017, December 4-9, 2017, Long Beach, CA, {USA}}, pages 4967--4976, 2017.

\bibitem[Sch{\"o}lkopf et~al.(2021)Sch{\"o}lkopf, Locatello, Bauer, Ke, Kalchbrenner, Goyal, and Bengio]{scholkopf2021toward}
Bernhard Sch{\"o}lkopf, Francesco Locatello, Stefan Bauer, Nan~Rosemary Ke, Nal Kalchbrenner, Anirudh Goyal, and Yoshua Bengio.
\newblock Toward causal representation learning.
\newblock \emph{Proceedings of the IEEE}, 109\penalty0 (5):\penalty0 612--634, 2021.

\bibitem[Seitzer et~al.(2023)Seitzer, Horn, Zadaianchuk, Zietlow, Xiao, Simon-Gabriel, He, Zhang, Sch{\"o}lkopf, Brox, et~al.]{seitzer2022bridging}
Maximilian Seitzer, Max Horn, Andrii Zadaianchuk, Dominik Zietlow, Tianjun Xiao, Carl-Johann Simon-Gabriel, Tong He, Zheng Zhang, Bernhard Sch{\"o}lkopf, Thomas Brox, et~al.
\newblock Bridging the gap to real-world object-centric learning.
\newblock In \emph{11th International Conference on Learning Representations, {ICLR} 2023}, 2023.

\bibitem[Singh et~al.(2022)Singh, Deng, and Ahn]{singh2022illiterate}
Gautam Singh, Fei Deng, and Sungjin Ahn.
\newblock Illiterate {DALL-E} learns to compose.
\newblock In \emph{10th International Conference on Learning Representations, {ICLR} 2022}, 2022.

\bibitem[Spelke and Kinzler(2007)]{spelke2007core}
Elizabeth~S Spelke and Katherine~D Kinzler.
\newblock Core knowledge.
\newblock \emph{Developmental science}, 10\penalty0 (1):\penalty0 89--96, 2007.

\bibitem[Stelzner et~al.(2021)Stelzner, Kersting, and Kosiorek]{stelzner2021decomposing}
Karl Stelzner, Kristian Kersting, and Adam~R Kosiorek.
\newblock Decomposing 3d scenes into objects via unsupervised volume segmentation.
\newblock \emph{ArXiv preprint}, abs/2104.01148, 2021.

\bibitem[Uria et~al.(2016)Uria, C{\^o}t{\'e}, Gregor, Murray, and Larochelle]{uria2016neural}
Benigno Uria, Marc-Alexandre C{\^o}t{\'e}, Karol Gregor, Iain Murray, and Hugo Larochelle.
\newblock Neural autoregressive distribution estimation.
\newblock \emph{The Journal of Machine Learning Research}, 17\penalty0 (1):\penalty0 7184--7220, 2016.

\bibitem[Vahdat and Kautz(2020)]{NEURIPS2020_e3b21256}
Arash Vahdat and Jan Kautz.
\newblock {NVAE:} {A} deep hierarchical variational autoencoder.
\newblock In Hugo Larochelle, Marc'Aurelio Ranzato, Raia Hadsell, Maria{-}Florina Balcan, and Hsuan{-}Tien Lin, editors, \emph{Advances in Neural Information Processing Systems 33: Annual Conference on Neural Information Processing Systems 2020, NeurIPS 2020, December 6-12, 2020, virtual}, 2020.

\bibitem[van~den Oord et~al.(2016)van~den Oord, Kalchbrenner, Espeholt, Kavukcuoglu, Vinyals, and Graves]{van2016conditional}
A{\"{a}}ron van~den Oord, Nal Kalchbrenner, Lasse Espeholt, Koray Kavukcuoglu, Oriol Vinyals, and Alex Graves.
\newblock Conditional image generation with pixelcnn decoders.
\newblock In Daniel~D. Lee, Masashi Sugiyama, Ulrike von Luxburg, Isabelle Guyon, and Roman Garnett, editors, \emph{Advances in Neural Information Processing Systems 29: Annual Conference on Neural Information Processing Systems 2016, December 5-10, 2016, Barcelona, Spain}, pages 4790--4798, 2016.

\bibitem[Van~Oord et~al.(2016)Van~Oord, Kalchbrenner, and Kavukcuoglu]{van2016pixel}
Aaron Van~Oord, Nal Kalchbrenner, and Koray Kavukcuoglu.
\newblock Pixel recurrent neural networks.
\newblock In \emph{International conference on machine learning}, pages 1747--1756. PMLR, 2016.

\bibitem[Vaswani et~al.(2017)Vaswani, Shazeer, Parmar, Uszkoreit, Jones, Gomez, Kaiser, and Polosukhin]{vaswani2017attention}
Ashish Vaswani, Noam Shazeer, Niki Parmar, Jakob Uszkoreit, Llion Jones, Aidan~N. Gomez, Lukasz Kaiser, and Illia Polosukhin.
\newblock Attention is all you need.
\newblock In Isabelle Guyon, Ulrike von Luxburg, Samy Bengio, Hanna~M. Wallach, Rob Fergus, S.~V.~N. Vishwanathan, and Roman Garnett, editors, \emph{Advances in Neural Information Processing Systems 30: Annual Conference on Neural Information Processing Systems 2017, December 4-9, 2017, Long Beach, CA, {USA}}, pages 5998--6008, 2017.

\bibitem[Veerapaneni et~al.(2019)Veerapaneni, Co-Reyes, Chang, Janner, Finn, Wu, Tenenbaum, and Levine]{veerapaneni2019entity}
Rishi Veerapaneni, John~D Co-Reyes, Michael Chang, Michael Janner, Chelsea Finn, Jiajun Wu, Joshua~B Tenenbaum, and Sergey Levine.
\newblock Entity abstraction in visual model-based reinforcement learning.
\newblock \emph{ArXiv preprint}, abs/1910.12827, 2019.

\bibitem[Vinyals et~al.(2015)Vinyals, Fortunato, and Jaitly]{vinyals2015pointer}
Oriol Vinyals, Meire Fortunato, and Navdeep Jaitly.
\newblock Pointer networks.
\newblock In Corinna Cortes, Neil~D. Lawrence, Daniel~D. Lee, Masashi Sugiyama, and Roman Garnett, editors, \emph{Advances in Neural Information Processing Systems 28: Annual Conference on Neural Information Processing Systems 2015, December 7-12, 2015, Montreal, Quebec, Canada}, pages 2692--2700, 2015.

\bibitem[Vinyals et~al.(2016)Vinyals, Bengio, and Kudlur]{vinyals2016order}
Oriol Vinyals, Samy Bengio, and Manjunath Kudlur.
\newblock Order matters: Sequence to sequence for sets.
\newblock In Yoshua Bengio and Yann LeCun, editors, \emph{4th International Conference on Learning Representations, {ICLR} 2016, San Juan, Puerto Rico, May 2-4, 2016, Conference Track Proceedings}, 2016.

\bibitem[von K{\"u}gelgen et~al.(2020)von K{\"u}gelgen, Ustyuzhaninov, Gehler, Bethge, and Sch{\"o}lkopf]{von2020towards}
Julius von K{\"u}gelgen, Ivan Ustyuzhaninov, Peter Gehler, Matthias Bethge, and Bernhard Sch{\"o}lkopf.
\newblock Towards causal generative scene models via competition of experts.
\newblock \emph{ArXiv preprint}, abs/2004.12906, 2020.

\bibitem[Watters et~al.(2019{\natexlab{a}})Watters, Matthey, Bosnjak, Burgess, and Lerchner]{watters2019cobra}
Nicholas Watters, Loic Matthey, Matko Bosnjak, Christopher~P Burgess, and Alexander Lerchner.
\newblock {COBRA:} data-efficient model-based rl through unsupervised object discovery and curiosity-driven exploration.
\newblock \emph{ArXiv preprint}, abs/1905.09275, 2019{\natexlab{a}}.

\bibitem[Watters et~al.(2019{\natexlab{b}})Watters, Matthey, Burgess, and Lerchner]{watters2019spatial}
Nicholas Watters, Loic Matthey, Christopher~P Burgess, and Alexander Lerchner.
\newblock Spatial broadcast decoder: A simple architecture for learning disentangled representations in vaes.
\newblock \emph{ArXiv preprint}, abs/1901.07017, 2019{\natexlab{b}}.

\bibitem[Yang et~al.(2019)Yang, Dai, Yang, Carbonell, Salakhutdinov, and Le]{NEURIPS2019_dc6a7e65}
Zhilin Yang, Zihang Dai, Yiming Yang, Jaime~G. Carbonell, Ruslan Salakhutdinov, and Quoc~V. Le.
\newblock {XLNet:} generalized autoregressive pretraining for language understanding.
\newblock In Hanna~M. Wallach, Hugo Larochelle, Alina Beygelzimer, Florence d'Alch{\'{e}}{-}Buc, Emily~B. Fox, and Roman Garnett, editors, \emph{Advances in Neural Information Processing Systems 32: Annual Conference on Neural Information Processing Systems 2019, NeurIPS 2019, December 8-14, 2019, Vancouver, BC, Canada}, pages 5754--5764, 2019.

\bibitem[Yu et~al.(2021)Yu, Guibas, and Wu]{yu2021unsupervised}
Hong-Xing Yu, Leonidas~J Guibas, and Jiajun Wu.
\newblock Unsupervised discovery of object radiance fields.
\newblock \emph{ArXiv preprint}, abs/2107.07905, 2021.

\bibitem[Yuan et~al.(2021)Yuan, Li, and Xue]{yuan2021unsupervised}
Jinyang Yuan, Bin Li, and Xiangyang Xue.
\newblock Unsupervised learning of compositional scene representations from multiple unspecified viewpoints.
\newblock \emph{ArXiv preprint}, abs/2112.03568, 2021.

\bibitem[Yuille and Kersten(2006)]{yuille2006vision}
Alan Yuille and Daniel Kersten.
\newblock Vision as bayesian inference: analysis by synthesis?
\newblock \emph{Trends in cognitive sciences}, 10\penalty0 (7):\penalty0 301--308, 2006.

\bibitem[Zablotskaia et~al.(2021)Zablotskaia, Dominici, Sigal, and Lehrmann]{zablotskaia2021provide}
Polina Zablotskaia, Edoardo~A Dominici, Leonid Sigal, and Andreas~M Lehrmann.
\newblock {PROVIDE}: a probabilistic framework for unsupervised video decomposition.
\newblock In \emph{Uncertainty in Artificial Intelligence}, pages 2019--2028. PMLR, 2021.

\bibitem[Zaheer et~al.(2017)Zaheer, Kottur, Ravanbakhsh, P{\'{o}}czos, Salakhutdinov, and Smola]{zaheer2017deep}
Manzil Zaheer, Satwik Kottur, Siamak Ravanbakhsh, Barnab{\'{a}}s P{\'{o}}czos, Ruslan Salakhutdinov, and Alexander~J. Smola.
\newblock Deep sets.
\newblock In Isabelle Guyon, Ulrike von Luxburg, Samy Bengio, Hanna~M. Wallach, Rob Fergus, S.~V.~N. Vishwanathan, and Roman Garnett, editors, \emph{Advances in Neural Information Processing Systems 30: Annual Conference on Neural Information Processing Systems 2017, December 4-9, 2017, Long Beach, CA, {USA}}, pages 3391--3401, 2017.

\bibitem[Zhang et~al.(2019)Zhang, Hare, and Pr{\"{u}}gel{-}Bennett]{zhang2018learning}
Yan Zhang, Jonathon~S. Hare, and Adam Pr{\"{u}}gel{-}Bennett.
\newblock Learning representations of sets through optimized permutations.
\newblock In \emph{7th International Conference on Learning Representations, {ICLR} 2019, New Orleans, LA, USA, May 6-9, 2019}. OpenReview.net, 2019.

\bibitem[Zhu et~al.(2020)Zhu, Gao, Fan, Huang, Edmonds, Liu, Gao, Zhang, Qi, Wu, et~al.]{zhu2020dark}
Yixin Zhu, Tao Gao, Lifeng Fan, Siyuan Huang, Mark Edmonds, Hangxin Liu, Feng Gao, Chi Zhang, Siyuan Qi, Ying~Nian Wu, et~al.
\newblock Dark, beyond deep: A paradigm shift to cognitive ai with humanlike common sense.
\newblock \emph{Engineering}, 6\penalty0 (3):\penalty0 310--345, 2020.

\bibitem[Zoran et~al.(2021)Zoran, Kabra, Lerchner, and Rezende]{zoran2021parts}
Daniel Zoran, Rishabh Kabra, Alexander Lerchner, and Danilo~J Rezende.
\newblock {PARTS:} unsupervised segmentation with slots, attention and independence maximization.
\newblock In \emph{Proceedings of the IEEE/CVF International Conference on Computer Vision}, pages 10439--10447, 2021.

\end{thebibliography}
